\documentclass[10pt,twocolumn,letterpaper]{article}

\usepackage[pagenumbers]{cvpr} 

\usepackage{graphicx}
\usepackage{amsmath}
\usepackage{amssymb}
\usepackage{booktabs}
\usepackage{enumitem,kantlipsum}
\usepackage{epsfig}
\usepackage{footnote}
\usepackage{csquotes}
\usepackage{multirow}
\usepackage{pict2e}
\usepackage{hhline}
\usepackage{stackengine}
\usepackage{makecell}
\usepackage{color,soul}
\usepackage{algorithm}
\usepackage{bm}

\usepackage{listings}
\usepackage{xcolor}

\definecolor{codeblue}{rgb}{0.0,0.5,0.9}
\definecolor{codegray}{rgb}{0.5,0.5,0.5}
\definecolor{codegreen}{rgb}{0.4,0.7,0.3}
\definecolor{backcolour}{rgb}{1.0,1.0,1.0}
\definecolor{codered}{rgb}{0.9, 0.2, 0.5}
\definecolor{rosegold}{rgb}{0.90, 0.70, 0.60}

\lstdefinestyle{mystyle}{
    backgroundcolor=\color{backcolour},   
    commentstyle=\color{codeblue},
    keywordstyle=\color{codered},
    numberstyle=\tiny\color{codegray},
    stringstyle=\color{codegreen},
    basicstyle=\ttfamily\scriptsize,
    breakatwhitespace=false,         
    breaklines=true,                 
    captionpos=b,                    
    keepspaces=true,                 
    numbers=none,                    
    numbersep=0pt,                  
    showspaces=false,                
    showstringspaces=false,
    showtabs=false,                  
    tabsize=2
}

\usepackage{hyperref}
\hypersetup{colorlinks,linkcolor={red},citecolor={rosegold},urlcolor={rosegold}}

\usepackage[capitalize]{cleveref}
\crefname{section}{Sec.}{Secs.}
\Crefname{section}{Section}{Sections}
\Crefname{table}{Table}{Tables}
\crefname{table}{Tab.}{Tabs.}

\def\cvprPaperID{****} 
\def\confName{CVPR}
\def\confYear{2023}

\begin{document}

\title{N-Gram in Swin Transformers for Efficient Lightweight Image Super-Resolution}

\author{
Haram Choi$^1$~~~
Jeongmin Lee$^2$~~~
Jihoon Yang$^1$\thanks{Corresponding author.}~~~\\
$^1$Department of Computer Science \& Engineering, Sogang University
\space\space\space\space
$^2$LG Innotek
}

\maketitle

\begin{abstract}
While some studies have proven that Swin Transformer (Swin) with window self-attention (WSA) is suitable for single image super-resolution (SR), the plain WSA ignores the broad regions when reconstructing high-resolution images due to a limited receptive field.
In addition, many deep learning SR methods suffer from intensive computations.
To address these problems, we introduce the N-Gram context to the low-level vision with Transformers for the first time.
We define N-Gram as neighboring local windows in Swin, which differs from text analysis that views N-Gram as consecutive characters or words.
N-Grams interact with each other by sliding-WSA, expanding the regions seen to restore degraded pixels.
Using the N-Gram context, we propose NGswin, an efficient SR network with SCDP bottleneck taking multi-scale outputs of the hierarchical encoder.
Experimental results show that NGswin achieves competitive performance while maintaining an efficient structure when compared with previous leading methods.
Moreover, we also improve other Swin-based SR methods with the N-Gram context, thereby building an enhanced model: SwinIR-NG.
Our improved SwinIR-NG outperforms the current best lightweight SR approaches and establishes state-of-the-art results.
Codes are available at \href{https://github.com/rami0205/NGramSwin}{https://github.com/rami0205/NGramSwin}.
\end{abstract}

\section{Introduction}
\label{sec:intro}

The goal of single image super-resolution (SR) is to reconstruct high-resolution (HR) images from low-resolution (LR) images.
Many deep learning-based methods have worked in this field.
In particular, several image restoration studies~\cite{liang2021swinir,wang2022uformer,fang2022hybrid,zhang2022efficient,zheng2022cross,zhang2022accurate} have adapted the window self-attention (WSA) proposed by Swin Transformer (Swin)~\cite{liu2021swin} as it integrates long-range dependency of Vision Transformer~\cite{dosovitskiy2020image} and locality of conventional convolution.
However, two critical problems remain in these works.
First, the receptive field of the plain WSA is limited within a small local window~\cite{yu2021glance,yang2021focal,tu2022maxvit}.
It prevents the models from utilizing the texture or pattern of neighbor windows to recover degraded pixels, producing the distorted images.
Second, recent state-of-the-art SR~\cite{chen2021pre,liang2021swinir,zheng2022cross,zhang2022accurate} and lightweight SR~\cite{zhang2022efficient,behjati2023single,lu2021efficient,du2022fast} networks require intensive computations.
Reducing operations is essential for real-world applications if the parameters are kept around a certain level (\eg, 1M, 4MB sizes), because the primary consumption of semiconductor energy (concerning time) for neural networks is related to Mult-Adds operations~\cite{han2021flash,ruiz2021dopant}.

\begin{figure}
    \centering
    \includegraphics[width=\linewidth]{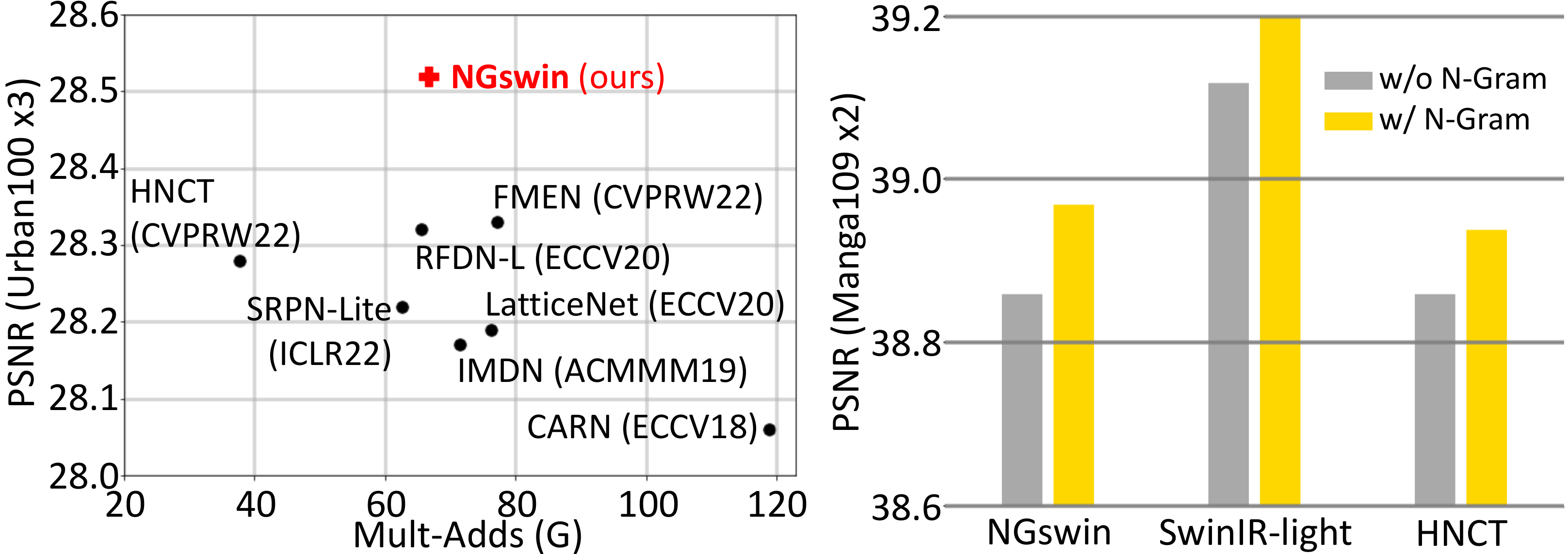}
    \caption{Two tracks of this paper using the N-Gram context. {\bf (Left)} NGswin outperforms previous leading SR methods with an efficient structure. {\bf (Right)} Our proposed N-Gram context improves different Swin Transformer-based SR models.}
    \label{fig_introduction}
    \vspace{-10pt}
\end{figure}

\begin{figure*}[t]
  \centering
  \includegraphics[width=0.95\linewidth]{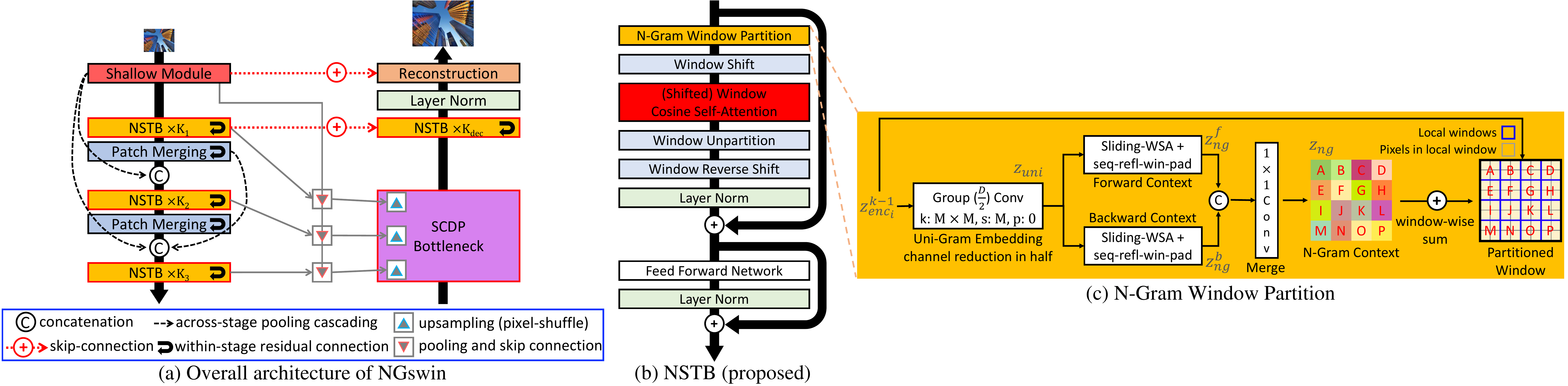}
  \caption{Overall architecture of NGswin and NSTB (N-Gram Swin Transformer Block). {\bf (a)} We adopt an asymmetric U-Net architecture. SCDP Bottleneck (pixel-Shuffle, Concatenation, Depth-wise convolution, and Point-wise projection), a variant of the U-Net bottleneck, takes multi-scale outputs of encoder stages, including the shallow module. {\bf (b)} Our proposed N-Gram method is implemented in NSTB. We also employ scaled-cosine attention and post-normalization. {\bf (c)} $k$, $s$, $p$, and $M$ are kernel size, stride, padding, and local window size, respectively. The dimensionality reduction through uni-Gram embedding makes \textit{sliding-WSA} efficient. The bi-directional contexts share \textit{sliding-WSA} weights. For window-wise sum, a value in $z_{ng}$ is equally added to $M^2$ pixels in one local window at the same position.}
  \label{fig_overall_nstb}
\end{figure*}

To overcome these problems, we define the N-Gram context as the interaction of neighbor local windows.
Neighbor uni-Gram embeddings interact with each other by \textit{sliding-WSA} to produce the N-Gram context features before window partitioning.
The uni-Gram embeddings result from a channel-reducing group convolution~\cite{cohen2016group} to decrease the complexity of N-Gram interaction (see \cref{fig_overall_nstb}\textcolor{red}{c}).
Our N-Gram context efficiently expands the receptive field of WSA for recovery tasks.
This work introduces N-Gram to low-level vision with Transformers for the first time, inspired by the following facts:
N-Gram language models treat the extended context beyond each separate word to understand text statistically~\cite{brown1992class}.
Since images have heavy spatial redundancy, some degraded pixels can be recovered from contextual information of neighbor pixels~\cite{he2022masked}.

As shown in \cref{fig_introduction}, our work progresses in two tracks.
{\bf Mainly}, to solve the problem of the intensive operations in SR, we propose an efficient N-Gram Swin Transformer ({\bf NGswin}).
As illustrated in \cref{fig_overall_nstb}\textcolor{red}{a}, NGswin consists of five components: a shallow module, three hierarchical encoder stages (with \textit{patch-merging}) that contain NSTBs (N-Gram Swin Transformer Blocks), SCDP Bottleneck (pixel-Shuffle, Concatenation, Depth-wise convolution, Point-wise projection), a small decoder stage with NSTBs, and a reconstruction module.
NSTBs employ our N-Gram context and the scaled-cosine attention proposed by Swin V2~\cite{liu2022swin}.
SCDP bottleneck, which takes multi-scale outputs of the encoder, is a variant of bottleneck from U-Net~\cite{ronneberger2015u}.
Experimental results demonstrate that the components above contribute to the efficient and competitive performance of NGswin.
{\bf Secondly}, focusing on improved performances, we apply the N-Gram context to other Swin-based SR models, such as SwinIR-light~\cite{liang2021swinir} and HNCT~\cite{fang2022hybrid}.
Notably, {\bf SwinIR-NG} (improved SwinIR-light with N-Gram) establishes state-of-the-art lightweight SR.

The main contributions of this paper are summarized as:
\begin{enumerate}[label=(\arabic*), left=0.25cm]
    \item We introduce the N-Gram context to the low-level vision with Transformer for the first time. It enables the SR networks to expand the receptive field to recover each degraded pixel by \textit{sliding-WSA}. For efficient calculation of N-Gram WSA, we produce uni-Gram embeddings by a channel-reducing group convolution.
    \vspace{-5pt}
    \item We propose an efficient SR network, NGswin. It exploits the hierarchical encoder (with \textit{patch-merging}), an asymmetrically small decoder, and SCDP bottleneck. These elements are critical for competitive performance in the efficient SR on $\times2$, $\times3$, and $\times4$ tasks.
    \vspace{-15pt}
    \item The N-Gram context improves other Swin Transformer methods. The improved SwinIR-NG achieves state-of-the-art results on lightweight SR.
\end{enumerate}

\section{Related Work}
\label{related}
\noindent
{\bf Efficient SR.}
Many single image super-resolution (SR) studies have increased network efficiency. 
CARN~\cite{ahn2018fast} introduced cascading residual blocks.
IMDN~\cite{hui2019lightweight} used information multi-distillation and selective feature fusion.
LatticeNet~\cite{luo2020latticenet} utilized the lattice filter that varies Fast Fourier Transformation.
ESRT~\cite{lu2021efficient} combined convolutional neural networks (CNN) and channel-reducing Transformers~\cite{vaswani2017attention}.
SwinIR-light~\cite{liang2021swinir} and HNCT~\cite{fang2022hybrid} appended CNN to Swin Transformer~\cite{liu2021swin}.
SRPN-Lite~\cite{zhang2021learning} applied the network pruning technique~\cite{reed1993pruning} on EDSR-baseline~\cite{lim2017enhanced}, a CNN-based lightweight SR model.
Most recently, ELAN-light~\cite{zhang2022efficient} utilized group-wise multi-scale self-attention.

\noindent
{\bf N-Gram.}
In language model (LM), N-Gram is a sequence of consecutive characters or words.
The size $N$ is typically set to 2 or 3~\cite{majumder2002n}.
The N-Gram LM that considers a longer span of context in sentences was operating well statistically in the past.
Even some deep learning LMs still adopted N-Gram.
Sent2Vec~\cite{pagliardini2017unsupervised} used N-Gram embeddings to learn sentence embedding by averaging word-embedding.
To learn the sentence representation better,~\cite{lopez2019word} computed the word N-Gram context by recurrent neural networks (RNN) and passed it to the attention layer.
ZEN~\cite{diao2019zen,song2021zen} trained a BERT-styled~\cite{devlin2018bert} N-Gram encoder for all possible N-Gram pairs from the Chinese or Arabic lexicon to convey salient pairs to the character encoder.
Meanwhile, some high-level vision studies also adopted this concept.
The Pixel N-grams approach~\cite{kulkami2016texture} saw N-Gram in pixel level and a single (horizontal or vertical) direction.
View N-gram Network~\cite{he2019view} regarded consecutive (along time steps) multi-view images of a 3D object as an N-Gram.
In contrast, our N-Gram considers bi-directional 2D information in local window level, given a single image for low-level vision.

\noindent
{\bf Swin Transformer.}
Swin Transformer~\cite{liu2021swin} (SwinV1) proposed window self-attention (WSA) that computes self-attention within non-overlapping local windows, to avoid quadratic time-complexity to the resolution of feature map.
SwinV1 also placed a shifted window scheme in consecutive layers, capturing interaction across windows.
Some studies~\cite{liang2021swinir,fang2022hybrid} utilized effective SwinV1 for SR.
The revised version~\cite{liu2022swin} (SwinV2) modified SwinV1.
For advanced model capacity with milder optimization, SwinV2 introduced residual post-normalization and scaled-cosine attention (in~\cref{eq_wsa}, (\ref{eq_cosine})) instead of pre-normalization configuration and scaled-dot-product attention.

\begin{figure}[b]
    \centering
    \includegraphics[width=0.95\linewidth]{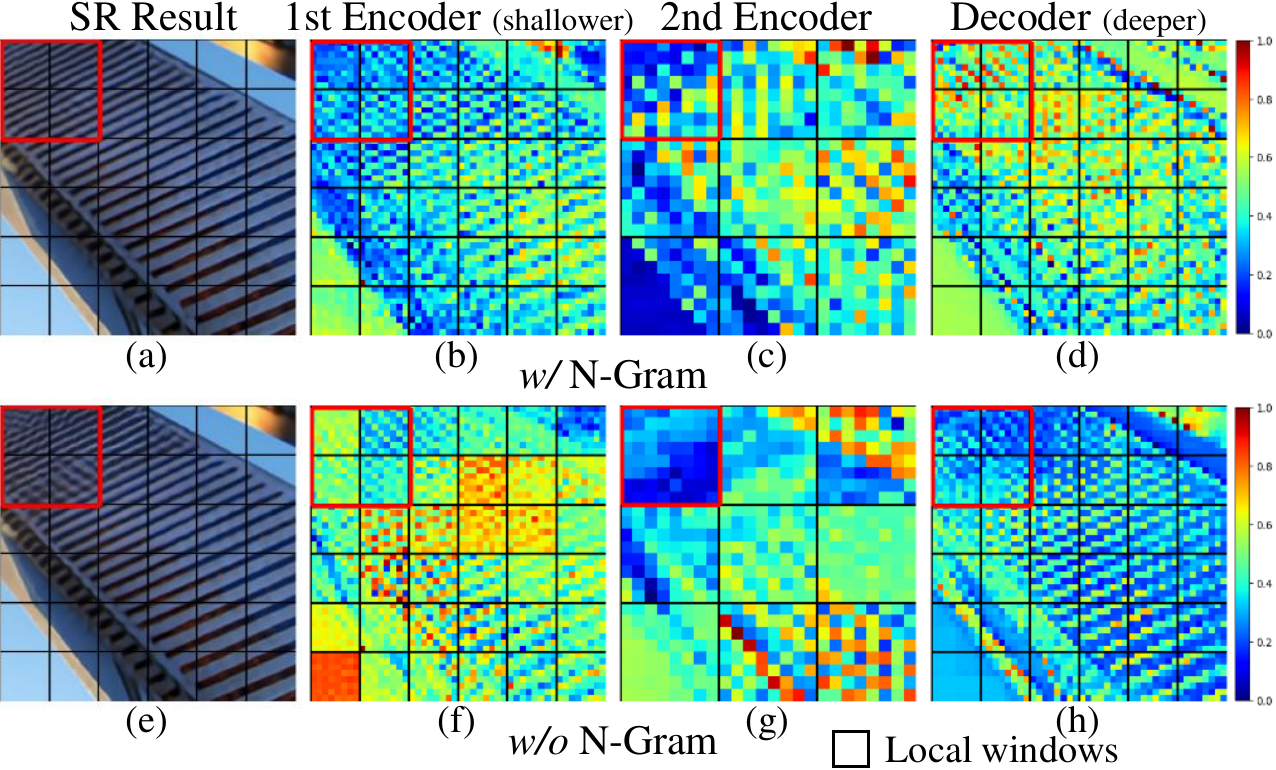}
    \vspace{-5pt}\caption{The feature maps after window self-attention in each intermediary layer of NGswin with and without N-Gram.}
    \label{fig_featuremap}
\end{figure}

\section{Methodology}
\label{Methodology}

\subsection{Problem Verification}
\label{problem}

The plain window self-attention (WSA) suffers from limited receptive field, as criticized in many recent studies~\cite{yu2021glance,yang2021focal,tu2022maxvit,ding2022davit,zhang2022vsa,zheng2022cross,zhang2022accurate}.
We observe this issue by visualizing feature maps after self-attention.
The similar pixel values from self-attention of the deeper layer tend to recover into a homogeneous pattern or texture.
In (h) of \cref{fig_featuremap}, however, the patterns in the red box and its neighbors differ (\textbf{\textit{problem}} $\bm{\alpha}$), causing the distortions in (e).
\textit{Problem} $\alpha$ stems from \textbf{\textit{problem}} $\bm{\beta}$: The plain WSA (\emph{i.e}\onedot, \textit{w/o} N-Gram) of the shallow layer is limited to only a local window.
It cannot utilize surrounding patterns to infer the recovery pattern of each window.
In (f) and (g), the distinctive colors across the adjacent windows reveal \textit{problem} $\beta$.
This seems to be resolved in the deeper layers, but it fails to overcome \textit{problem} $\alpha$.
To address this, we propose the N-Gram context to compensate this vulnerability.
Our N-Gram attention can consider broad regions (\emph{i.e}\onedot, surrounding patterns) beyond a window.
In (a)-(d), the semantically relevant areas yield similar attention results, producing more accurate details.
This crucial advantage solves both \textit{problem} $\alpha$ and $\beta$.

\subsection{Definition of N-Gram in Image}
\label{ngram_in_image}

\begin{figure}[t]
    \centering
    \begin{subfigure}[h]{\linewidth}
        \includegraphics[width=\linewidth]{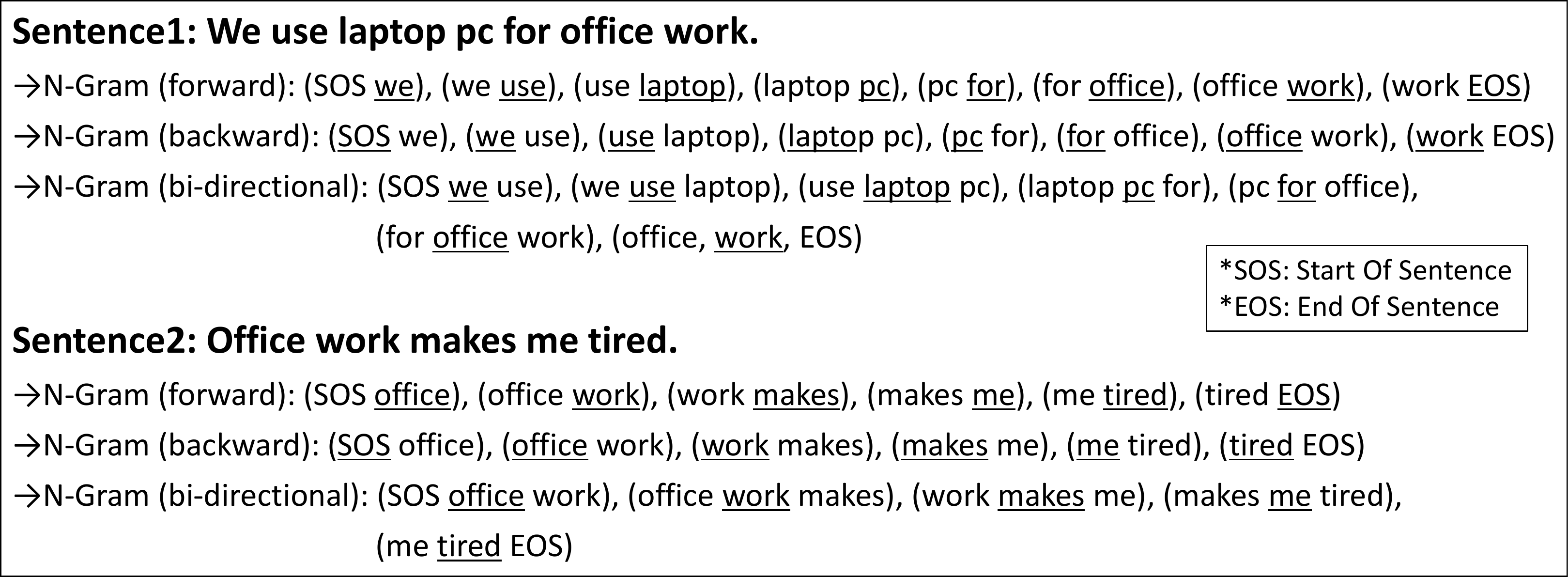}
        \caption{N-Gram in text.}
        \label{fig_ngram_text}
    \vspace{0.2cm}
    \end{subfigure}
    \centering
    \begin{subfigure}[h]{0.9\linewidth}
        \includegraphics[width=\linewidth]{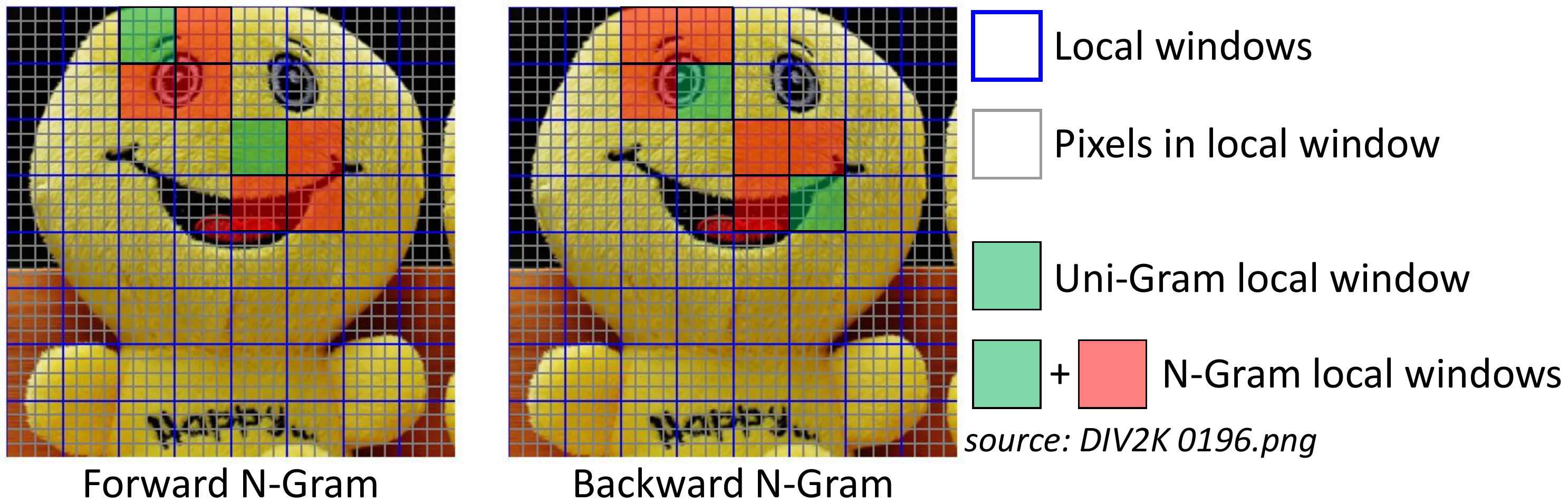}
        \caption{N-Gram in image (proposed).}
        \label{fig_ngram_image}
    \end{subfigure}
\caption{N-Gram in text and image ($N$ = $2$). {\bf (a)} The \underline{underlined words} are the target words and the non-underlined words are neighbors of the target words. {\bf (b)} Each local window is defined as uni-Gram. The lower-right (or upper-left) local windows are defined as forward (or backward) N-Gram neighbors.}
\vspace{-12pt}
\end{figure}

\noindent
{\bf N-Gram in text.}
As shown in~\cref{fig_ngram_text}, the N-Gram language model views the consecutive forward, backward, or bi-directional words as the N-Gram of the target word.
The words are independent of each other for uni-Gram (\emph{i.e}\onedot, word-embedding), but they interact with each other by averaging word-embeddings~\cite{pagliardini2017unsupervised}, RNN~\cite{lopez2019word}, or attention~\cite{diao2019zen} when considering N-Gram.
In contrast, an N-Gram composed of a particular word pair (\eg, “office work”) never interacts with the other N-Gram combinations when producing an N-Gram feature.

\noindent
{\bf N-Gram in image.}
N-Gram in an image should have the properties above.
Accordingly, we define a uni-Gram as a non-overlapping local window in Swin Transformer, within which the pixels interact with each other by self-attention (SA).
N-Gram is defined as the larger window, including neighbors of each uni-Gram.
To sum up, pixels of each uni-Gram and uni-Grams of each N-Gram in image correspond to characters of each word and words of each N-Gram in text, respectively.
As depicted in~\cref{fig_ngram_image}, setting N-Gram size $N$ to 2 indicates a bi-Gram that combines a local window (green area) and its neighbor windows (red areas) at lower-right (forward) or upper-left (backward).
The N-Gram interactions will be explained in~\cref{NSTB}.

\subsection{Overall Architecture of NGswin}
\label{overall_architecture}
As illustrated in \cref{fig_overall_nstb}\textcolor{red}{a},
we adopt U-Net~\cite{ronneberger2015u} architecture: hierarchical encoder stages, a bottleneck layer, a decoder stage, and skip-connection from the encoder to the decoder at the same resolution\footnote{In this paper, \enquote{resolution} indicates height and width of feature maps, excluding network dimension (channel).}.
However, our network's encoder and decoder are asymmetric, which indicates a significantly smaller decoder~\cite{he2022masked,pang2022masked}.

\noindent
{\bf Encoder.}
Given a low-resolution (LR) image $I_{LR} \in \mathbb{R}^{3 \times H \times W}$, a shallow module (a $3 \times 3$ convolution) extracts $z_s\in \mathbb{R}^{HW\times D}$, where $H$, $W$, and $D$ stand for height, width, and network dimension (channels), respectively.
$z_s$ is passed through three encoder stages, each composed of $\mathcal{K}_i$ N-Gram Swin Transformer Blocks (NSTB, \cref{NSTB}) and a $2\times2$ \textit{patch-merging} except the last stage.
We set $\{\mathcal{K}_1,\mathcal{K}_2,\mathcal{K}_3\}$ to $\{6,4,4\}$ by default.
The mapping function $\mathcal{F}^k_{enc_i}$ of $k$-$th$ ($1$$\leq$$k$$\leq$$\mathcal{K}_i$) NSTB in $i$-$th$ ($1$$\leq$$i$$\leq$$3$) encoder stage is formulated as:
\begin{equation}
\label{eq_encoder}
    z^k_{enc_i} = \mathcal{F}^k_{enc_i}(z^{k-1}_{enc_{i}}),\ z^{k}_{enc_{i}} \in \mathbb{R}^{HW/(2^{i-1})^{2} \times D},
\end{equation}
where $z^0_{enc_i}$ equals $z_{enc_{i-1}}$, which results from downsampling $z^{\mathcal{K}_{i-1}}_{enc_{i-1}}$ ($z_{enc_0} = z_s$).
In other words, the first NSTB in the 2nd or 3rd stage takes the output of \textit{patch-merging} in the previous stage as input.
The \textit{patch-merging} follows Swin Transformer~\cite{liu2021swin,liu2022swin}, except that the network dimension is decreased from $4D$ to $D$ instead of $2D$.
Since \textit{patch-merging} halves the resolutions, NGswin consumes much fewer attention computations than state-of-the-art attention-based lightweight SR methods, as revealed in \cref{tab_complexitiy}.

\begin{table}[t]
\caption{Comparison of computational complexity with state-of-the-art networks. Our NGswin is much more efficient. Mult-Adds is evaluated on a $1280\times720$ HR image.}
\label{tab_complexitiy}
    \centering
    \vspace{-2.5pt}
    \resizebox{\linewidth}{!}{
        \begin{tabular}{c|c||cccc}
        \hline
        Scale & {\bf NGswin} & SwinIR-light~\cite{liang2021swinir}\footnotemark & ESRT~\cite{lu2021efficient} & DiVANet~\cite{behjati2023single} & ELAN-light~\cite{zhang2022efficient} \\
        \hline
        x2 & {\bf 140.4G} & 243.7G & 191.4G & 189.0G & 168.4G \\
        x3 & {\bf 66.6G} & 109.5G & 96.4G & 89.0G & 75.7G \\
        x4 & {\bf 36.4G} & 61.7G & 67.7G & 57.0G & 43.2G \\
        \hline
        \end{tabular}
    }
    \label{tab_mult_adds_comp}
    \vspace{4pt}
\end{table}

\footnotetext{We correct Mult-Adds~\cite{liang2021swinir} underestimated on a 1024$\times$720 HR image.}

\noindent
{\bf Pooling Cascading.}
Following the global cascading in CARN~\cite{ahn2018fast}, we employ a cascading mechanism (\textcircled{c} marks and the dashed lines in \cref{fig_overall_nstb}\textcolor{red}{a}) across the stages, including the shallow module.
Unlike CARN, we place 2$\times$2 max-poolings before concatenating the intermediary features because the first and second stages halve the resolutions of features.
This dense connectivity~\cite{huang2017densely} reflects the flow of the information and gradient in the previous layers, which helps the network to learn meaningful representations.

\noindent
{\bf Bottleneck.}
All outputs from the shallow module and the last NSTBs of each encoder stage are taken by SCDP bottleneck.
The bottleneck layer maps them into $z_{scdp} \in \mathbb{R}^{HW \times D}$.
A detailed explanation is in \cref{SCDP}.

\begin{figure}[t]
    \centering
    \includegraphics[width=\linewidth]{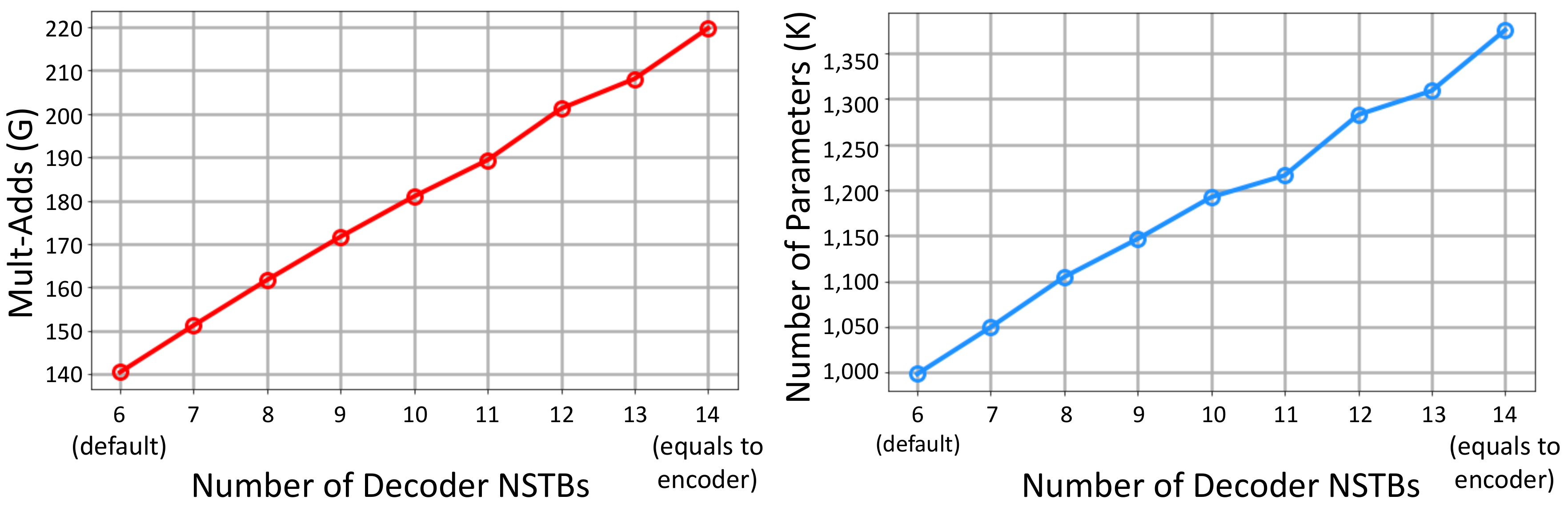}
    \caption{Efficiency of asymmetrically small decoder in NGswin. Mult-Adds is evaluated on $\times 2$ task with a 1280$\times$720 image.}
    \label{fig_asymmetric}
\end{figure}

\noindent
{\bf Decoder.}
$z_{scdp}$ is fed into a single decoder stage, which is asymmetrically smaller than the encoder~\cite{he2022masked,pang2022masked}.
This means fewer stages and NSTBs in our decoder, which highly enhances the efficiency as shown in \cref{fig_asymmetric}.
It contains $\mathcal{K}_{dec}$ (by default, 6) NSTBs and a final layer-norm (LN)~\cite{ba2016layer} that allows stable learning.
The decoder NSTB architecture is the same as the encoder NSTB.
As done in U-Net~\cite{ronneberger2015u}, the input to the decoder is residually connected~\cite{he2016deep} with $z^{\mathcal{K}_1}_{enc_1}$ of the first encoder stage.
$z_s$ and decoder output $z_{dec} \in \mathbb{R}^{HW \times D}$ are added with a global skip-connection~\cite{kim2016accurate,liang2021swinir,ahn2022efficient}.
This boosts optimization and allows the reconstruction module to utilize both locality and long-range dependency.

\noindent
{\bf Reconstruction.}
Following~\cite{kim2016accurate,lim2017enhanced,ahn2018fast,liang2021swinir}, the reconstruction module contains a convolution that adjusts dimension and a pixel-shuffler~\cite{shi2016real}.
Unlike previous works, we additionally place a convolution that produces the SR image $I_{SR}\in \mathbb{R}^{3 \times rH \times rW}$, where $r$ is a scale factor (\eg, $\times4$).
The illustration is in the supplementary~\cref{reconstruction}.

\subsection{N-Gram Swin Transformer Block (NSTB)}
\label{NSTB}

As illustrated in \cref{fig_overall_nstb}\textcolor{red}{b}, our NSTB adopts scaled-cosine attention and post-normalization proposed in SwinV2~\cite{liu2022swin}.
For scaled-cosine window self-attention (WSA), we use the following formula:
\begin{equation}
\label{eq_wsa}
    WSA(Q, K, V) = Softmax(cos(Q,K)/\tau + B)V,
\end{equation}
\noindent
where $Q,K,V \in \mathbb{R}^{M^2 \times D}$ are the query, key, and value matrices; $B \in \mathbb{R}^{M^2 \times M^2}$ is the relative position bias between each pixel within a local window; $\tau$ is a learnable scalar set to larger than 0.01~\cite{liu2022swin}.
$M$ is the window size set to 8 by default, and the corresponding $M^2$ indicates the number of pixels in a local window.
For the given matrices $Q=\begin{bmatrix} q_{ij} \end{bmatrix}$ and $K=\begin{bmatrix} k_{ij} \end{bmatrix}$, the cosine similarity is calculated as:
\vspace{-5pt}
\begin{equation}
\label{eq_cosine}
\vspace{-3pt}
    cos(Q,K) = \begin{bmatrix} q_{ij} / \lVert q_i \rVert_2 \end{bmatrix} \begin{bmatrix} k_{ij} / \lVert k_i \rVert_2 \end{bmatrix}^T
\end{equation}

In window partitioning (the top of \cref{fig_overall_nstb}\textcolor{red}{b}), we implement the N-Gram context algorithm through four steps (see \cref{fig_overall_nstb}\textcolor{red}{c}).
This algorithm is identically applied to other Swin Transformer models (SwinIR-light~\cite{liang2021swinir}, HNCT~\cite{fang2022hybrid}), focusing only on better performances.
As denoted by \cref{eq_encoder}, the input to $k$-$th$ NSTB in $i$-$th$ encoder stage is $z^{k-1}_{enc_{i}} \in \mathbb{R}^{hw \times D}$, where $h=H/2^{i-1}$ and $w=W/2^{i-1}$.

{\bf First},
the input is mapped into the uni-Gram ($N$ $=$ $1$) embedding $z_{uni} \in \mathbb{R}^{\frac{D}{2} \times w_h \times w_w}$ by an $M\times M$ channel-reducing group convolution~\cite{cohen2016group} (stride: $M$, groups: $\frac{D}{2}$).
$w_h$ ($=$$\frac{h}{M}$) and $w_w$ ($=$$\frac{w}{M}$) represent the number of windows in height and width.
It is worth noting that the reduction of channel and resolution by uni-Gram embedding makes N-Gram WSA (in the next step) more efficient.
Considering $\Omega(WSA)=4hwD^2 + 2M^2hwD$~\cite{liu2021swin}, halved $D$ and $M^2$ times reduced $hw$ highly decrease computations.

\begin{figure}[t]
    \centering
    \includegraphics[width=0.6\linewidth]{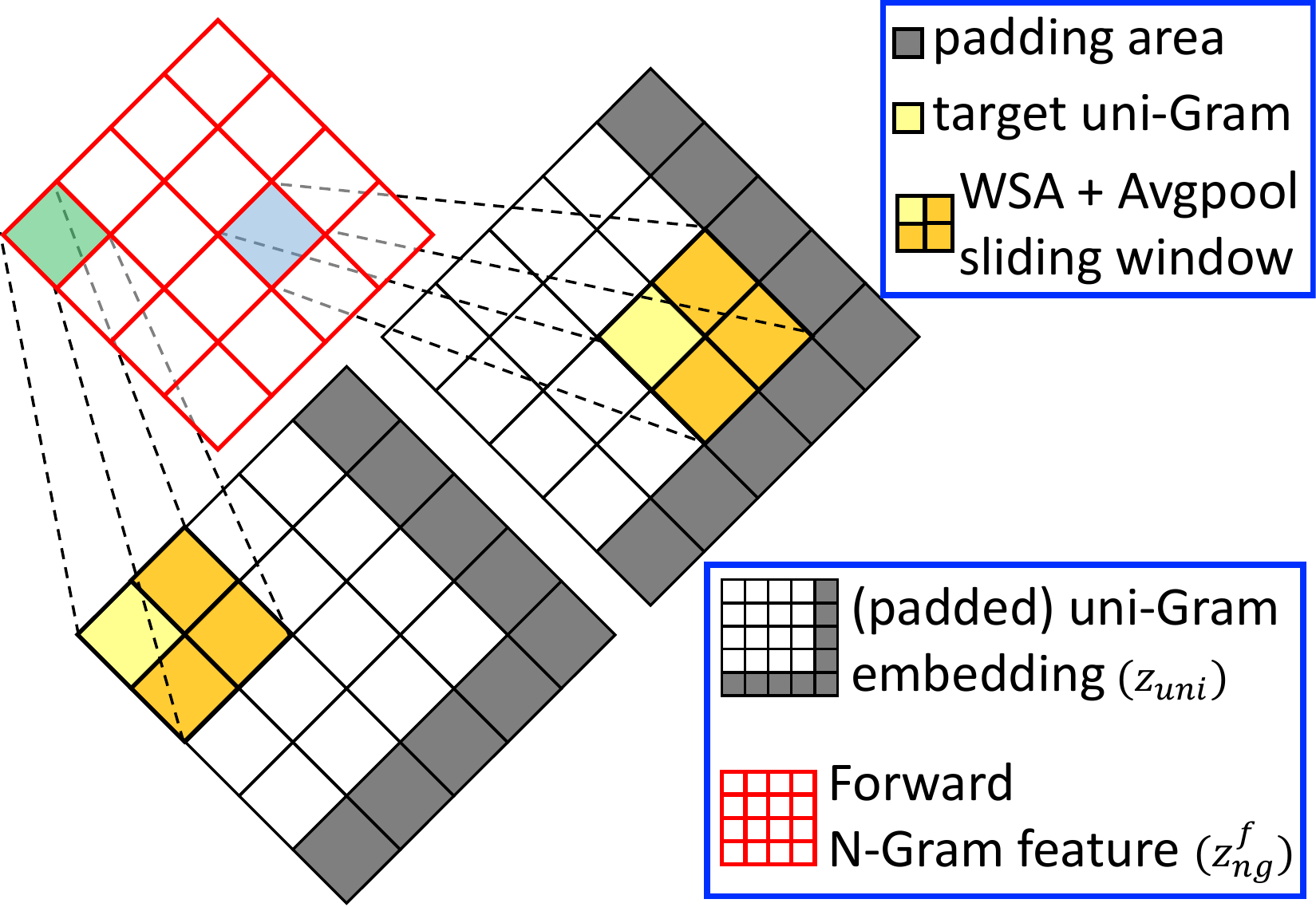}
    \caption{Sliding-WSA. A window operates SA and avg-pool to get the forward N-Gram feature, as the window slides through the uni-Gram embedding. $z^b_{ng}$ can be obtained by upper-left padding.}
    \label{fig_sliding_wsa}
    \vspace{-7pt}
\end{figure}

{\bf Second},
the $N^2$ pixels in each N-Gram ($N$ $>$ $1$) of $z_{uni}$ interact with each other by WSA (by \cref{eq_wsa} with $M$$=$$N$ and $D$$=$$\frac{D}{2}$) to obtain the forward N-Gram feature $z^f_{ng} \in \mathbb{R}^{w_h \times w_w \times \frac{D}{2}}$.
As shown in \cref{fig_sliding_wsa}, we implement \textit{sliding-WSA} as sliding-window convolution operated in CNN.
As an $N$$\times$$N$ window slides through $z_{uni}$, scaled-cosine self-attention and $N$$\times$$N$ average-pooling are computed.
But the scaled-dot-product attention is used for SwinIR-light and HNCT, following their own methods.
We use \textit{seq-refl-win-pad} instead of trivial zero padding for ($N$$-$$1$) size of paddings, as explained in the supplementary~\cref{seq_refl_win_pad}.
Subsequently, we can obtain the backward N-Gram feature $z^b_{ng}$ by reversed \textit{seq-refl-win-pad} (\emph{i.e}\onedot, upper-left side padding).
The computations of bi-directional N-Gram features share the \textit{sliding-WSA} weights.
Since the image is two-dimensional data, our N-Gram can be seen from max quad-directions (lower-right, lower-left, upper-right, and upper-left), unlike text that can be seen from max bi-directions.
However, \cref{tab_ablation_direction_interaction} demonstrates that the trade-off between performance and efficiency is optimized at the bi-directions.

{\bf Third},
after the concatenation of $z^f_{ng}$ and $z^b_{ng}$, a $1\times1$ convolution merges it to produce the N-Gram context $z_{ng}$.

{\bf Finally},
$z_{ng} \in \mathbb{R}^{D \times w_h \times w_w}$ is added window-wise to the partitioned windows (size: $M^2$$\times$$D$$\times$$w_h$$\times$$w_w$) from $z^{k-1}_{enc_{i}}$.
In \cref{fig_overall_nstb}\textcolor{red}{c}, one value in $z_{ng}$ is equally added to $M^2$ pixels in one local window at the same position (marked as the same character) \textemdash\space \emph{i.e}\onedot, the average correlations from self-attention within each N-Gram serve as bias terms of each pixel.
After the four steps, NSTB follows the sequence in \cref{fig_overall_nstb}\textcolor{red}{b}.
The window-shifts are operated in the even numbered blocks, same as in Swin Transformer.

NSTBs and \textit{patch-merging} within a stage are residually connected~\cite{he2016deep} from a previous layer to the next (the rounded arrows of \cref{fig_overall_nstb}\textcolor{red}{a}), rather than dense connections~\cite{huang2017densely}.
The connection to a \textit{patch-merging}, however, is excepted from the third encoder stage and the decoder stage.
For more, see the supplementary~\cref{within_stage}.

\subsection{SCDP Bottlneck}
\label{SCDP}
Many SR models~\cite{zhang2018image,niu2020single,liang2021swinir} commonly never used the hierarchical encoder, which downsamples the resolutions of features after one stage.
As demonstrated in~\cref{tab_ablation_scdp}, the hierarchical networks are inferior to the less (or non) hierarchical architectures.
However, our encoder is constructed hierarchically by \textit{patch-merging}.
Thus, only passing the output $z^{\mathcal{K}_3}_{enc_3}$ of the last NSTB in the final encoder stage to the bottleneck makes the recovery task more challenging.
The use of SCDP bottleneck, though, can convey rich representations of multi-scale features to the decoder and maintain the efficiency of NGswin.
\cref{algorithm_scdp} provides the pseudo-code of SCDP pipeline.
It contains dimensionality rearrangements, non-linear activation functions, and LN omitted in the main text below.

SCDP stands for pixel-\underline{{\bf S}}huffle, \underline{{\bf C}}oncatenation, \underline{{\bf D}}epth-wise convolution, and \underline{{\bf P}}oint-wise projection.
In contrast with the bottleneck of standard U-Net that takes the output of the last encoder layer~\cite{ronneberger2015u,wang2022uformer,Zamir2021Restormer}, SCDP bottleneck takes multi-scale outputs from the encoder.
{\bf First},
as depicted in the blue-edged triangles in~\cref{fig_overall_nstb}\textcolor{red}{a}, for obtaining $z'_{enc_i}$
the pixel-shuffle~\cite{shi2016real} layers upsample the outputs $z^{\mathcal{K}_i}_{enc_i}$ of the last NSTB in each encoder stage into the resolution of $I_{LR}$, $H \times W$.
Before upsizing, $z_s$ is iteratively max-pooled into the resolution of each $z^{\mathcal{K}_i}_{enc_{i}}$, then added to $z^{\mathcal{K}_i}_{enc_{i}}$.
This process gives multi-scale information to the bottleneck.
{\bf Second},
all $z'_{enc_{i}}$ are concatenated in channel (network dimension) space.
{\bf Third}, 
the output passes through a $3$$\times$$3$ depth-wise convolutional layer for learning spatial representations in each channel space.
{\bf Finally}, a point-wise linear projection is applied to match dimension $D$.
As a result, we get $z_{scdp}$ and then add it to $z^{\mathcal{K}_1}_{enc_1}$ to pass it to the decoder.

\begin{algorithm}[t]
    \caption{SCDP Bottleneck Pseudo-code, PyTorch-like}
    \label{algorithm_scdp}
\begin{lstlisting}[language=Python]
# zi: output list of last NSTBs in three encoder stages
# zs: output of shallow module

x = list()
for i in range(3): # pixel-"S"huffle
    x_ = zi[i] + down(zs, i) # before shuffling
    x.append(PixelShuffle(x_, 2**i))
x = torch.cat(x, dim=-1) # "C"oncatenation
x = Rearrange(x, '(h w) d -> d h w') # ignores batch
x = GELU(depth_wise(x)) # "D"epth-wise convolution
x = Rearrange(x, 'd h w -> (h w) d')
x = LayerNorm(point_wise(x)) # "P"oint-wise projection

def down(z, exp): # downsizing zs
    z = Rearrange(z, '(h w) d -> d h w')
    for e in range(exp): # iterative max-poolings
        z = MaxPool2D(z) # 2x2 pool
    z = LeakyReLU(z)
    return Rearrange(z, 'd h w -> (h w) d')
\end{lstlisting}
\end{algorithm}

\section{Experiments}
\label{experiments}

\subsection{Experimental Setup}
\label{setup}

\begin{table*}[t]
    \caption{Comparison of efficient super-resolution results. D2K stands for the DIV2K dataset we used to train NGswin. DF2K indicates a merged dataset of D2K and Flickr2K~\cite{timofte2017ntire} containing 800 + 2,650 HR-LR image pairs. 291 images dataset is from ~\cite{yang2010image,arbelaez2010contour}. Mult-Adds is evaluated on a $1280\times720$ HR image. The best, second best, and third best performances are in \textcolor{red}{red}, \textcolor{blue}{blue}, and \underline{underline}.}
    \label{tab_light_comp}
    \centering
    \resizebox{0.843\linewidth}{!}
    {\begin{tabular}{|c|c|c|c|c|c|c|c|c|c|c|c|c|c|c|}
        \hline
        \multirow{2}{*}{Method} & Training & \multirow{2}{*}{Scale} & \multirow{2}{*}{Mult-Adds} & \multirow{2}{*}{\#Params} & \multicolumn{2}{c|}{Set5} & \multicolumn{2}{c|}{Set14} & \multicolumn{2}{c|}{BSD100} & \multicolumn{2}{c|}{Urban100} & \multicolumn{2}{c|}{Manga109}  \\ \cline{6-15}
        &Dataset&&&&PSNR&SSIM&PSNR&SSIM&PSNR&SSIM&PSNR&SSIM&PSNR&SSIM \\
        \hline
        EDSR-baseline~\cite{lim2017enhanced} & D2K & $\times 2$ & 316.3G & 1,370K & 37.99 & 0.9604 & 33.57 & 0.9175 & 32.16 & 0.8994 & 31.98 & 0.9272 & 38.54 & 0.9769 \\
        MemNet~\cite{tai2017memnet} & 291 & $\times 2$ & 2,662.4G & 677K & 37.78 & 0.9597 & 33.28 & 0.9142 & 32.08 & 0.8978 & 31.31 & 0.9195 & - & - \\
        CARN~\cite{ahn2018fast} & D2K+291 & $\times 2$ & 222.8G & 1,592K & 37.76 & 0.9590 & 33.52 & 0.9166 & 32.09 & 0.8978 & 31.92 & 0.9256 & 38.36 & 0.9765 \\
        IMDN~\cite{hui2019lightweight} & D2K & $\times 2$ & 158.8G & 694K & 38.00 & 0.9605 & 33.63 & 0.9177 & 32.19 & 0.8996 & 32.17 & 0.9283 & 38.88 & \underline{0.9774} \\
        LatticeNet~\cite{luo2020latticenet} & D2K & $\times 2$ & 169.5G & 756K & \underline{38.06} & 0.9607 & \underline{33.70} & 0.9187 & 32.20 & 0.8999 & 32.25 & 0.9288  & \underline{38.94} & \underline{0.9774} \\
        RFDN-L~\cite{liu2020residual} & D2K & $\times 2$ & 145.8G & 626K & \textcolor{blue}{38.08}  & 0.9606  & 33.67  & \underline{0.9190}  & 32.18  & 0.8996  & 32.24  & 0.9290  & \textcolor{blue}{38.95}  & 0.9773 \\
        SRPN-Lite~\cite{zhang2021learning} & DF2K & $\times 2$ & 139.9G & 609K & \textcolor{red}{38.10} & \underline{0.9608} & \underline{33.70} & 0.9189 & \underline{32.25} & \underline{0.9005} & \underline{32.26} & \underline{0.9294} & - & - \\
        HNCT~\cite{fang2022hybrid} & D2K & $\times 2$ & 82.4G & 357K & \textcolor{blue}{38.08}  & \underline{0.9608}  & 33.65  & 0.9182  & 32.22  & 0.9001  & 32.22  & \underline{0.9294}  & 38.87  & \underline{0.9774} \\
        FMEN~\cite{du2022fast} & DF2K & $\times 2$ & 172.0G & 748K & \textcolor{red}{38.10}  & \textcolor{blue}{0.9609}  & \textcolor{blue}{33.75}  & \textcolor{blue}{0.9192}  & \textcolor{blue}{32.26}  & \textcolor{blue}{0.9007}  & \textcolor{blue}{32.41}  & \textcolor{blue}{0.9311}  & \textcolor{blue}{38.95}  & \textcolor{red}{0.9778} \\
        \hhline{|=|=|=|=|=|=|=|=|=|=|=|=|=|=|=|}
        {\bf NGswin (ours)} & {\bf D2K} & {\bf $\times 2$} & {\bf 140.4G} & {\bf 998K} & {\bf 38.05} & {\bf \textcolor{red}{0.9610}} & {\bf \textcolor{red}{33.79}} & {\bf \textcolor{red}{0.9199}} & {\bf \textcolor{red}{32.27}} & {\bf \textcolor{red}{0.9008}} & {\bf \textcolor{red}{32.53}} & {\bf \textcolor{red}{0.9324}} & {\bf \textcolor{red}{38.97}} & {\bf \textcolor{blue}{0.9777}} \\
        \hline\hline
        EDSR-baseline~\cite{lim2017enhanced} & D2K & $\times 3$ & 160.2G & 1,555K & 34.37 & 0.9270 & 30.28 & 0.8417 & 29.09 & 0.8052 & 28.15 & 0.8527 & 33.45 & 0.9439 \\
        MemNet~\cite{tai2017memnet} & 219 & $\times 3$ & 2,662.4G & 677K & 34.09 & 0.9248 & 30.00 & 0.8350 & 28.96 & 0.8001 & 27.56 & 0.8376 & - & - \\
        CARN~\cite{ahn2018fast} & D2K+291 & $\times 3$ & 118.8G & 1,592K & 34.29 & 0.9255 & 30.29 & 0.8407 & 29.06 & 0.8034 & 28.06 & 0.8493 & 33.50 & 0.9440 \\
        IMDN~\cite{hui2019lightweight} & D2K & $\times 3$ & 71.5G & 703K & 34.36 & 0.9270 & 30.32 & 0.8417 & 29.09 & 0.8046 & 28.17 & 0.8519 & 33.61 & 0.9445 \\
        LatticeNet~\cite{luo2020latticenet} & D2K & $\times 3$ & 76.3G & 765K & 34.40 & 0.9272 & 30.32 & 0.8416 & 29.10 & 0.8049 & 28.19 & 0.8513 & 33.63 & 0.9442 \\
        RFDN-L~\cite{liu2020residual} & D2K & $\times 3$ & 65.6G & 633K & \textcolor{blue}{34.47} & \textcolor{blue}{0.9280} & 30.35 & 0.8421 & 29.11 & 0.8053 & \underline{28.32} & 0.8547 & 33.78 & 0.9458 \\
        SRPN-Lite~\cite{zhang2021learning} & DF2K & $\times 3$ & 62.7G & 615K & \textcolor{blue}{34.47} & \underline{0.9276} & 30.38 & 0.8425 & \underline{29.16} & 0.8061 & 28.22 & 0.8534 & - & - \\
        HNCT~\cite{fang2022hybrid} & D2K & $\times 3$ & 37.8G & 363K & \textcolor{blue}{34.47}  & 0.9275  & \textcolor{blue}{30.44}  & \textcolor{blue}{0.8439}  & 29.15  & \textcolor{blue}{0.8067}  & 28.28  & \underline{0.8557}  & \underline{33.81}  & \underline{0.9459} \\
        FMEN~\cite{du2022fast} & DF2K & $\times 3$ & 77.2G & 757K & \underline{34.45}  & 0.9275  & \underline{30.40}  & \underline{0.8435}  & \textcolor{blue}{29.17}  & \underline{0.8063}  & \textcolor{blue}{28.33}  & \textcolor{blue}{0.8562}  & \textcolor{blue}{33.86}  & \textcolor{blue}{0.9462} \\
        \hhline{|=|=|=|=|=|=|=|=|=|=|=|=|=|=|=|}
        {\bf NGswin (ours)} & {\bf D2K} & {\bf $\times 3$} &{\bf  66.6G} & {\bf 1,007K} & {\bf \textcolor{red}{34.52}} & {\bf \textcolor{red}{0.9282}} & {\bf \textcolor{red}{30.53}} & {\bf \textcolor{red}{0.8456}} & {\bf \textcolor{red}{29.19}} & {\bf \textcolor{red}{0.8078}} & {\bf \textcolor{red}{28.52}} & {\bf \textcolor{red}{0.8603}} & {\bf \textcolor{red}{33.89}} & {\bf \textcolor{red}{0.9470}} \\
        \hline\hline
        EDSR-baseline~\cite{lim2017enhanced} & D2K & $\times 4$ & 114.0G & 1,518K & 32.09 & 0.8938 & 28.58 & 0.7813 & 27.57 & 0.7357 & 26.04 & 0.7849 & 30.35 & 0.9067 \\
        MemNet~\cite{tai2017memnet} & 291 & $\times 4$ & 2,662.4G & 677K & 31.74 & 0.8893 & 28.26 & 0.7723 & 27.40 & 0.7281 & 25.50 & 0.7630 & - & - \\
        CARN~\cite{ahn2018fast} & D2K+291 & $\times 4$ & 90.9G & 1,592K & 32.13 & 0.8937 & 28.60 & 0.7806 & \underline{27.58} & 0.7349 & 26.07 & 0.7837 & 30.47 & 0.9084 \\
        IMDN~\cite{hui2019lightweight} & D2K & $\times 4$ & 40.9G & 715K & 32.21 & 0.8948 & 28.58 & 0.7811 & 27.56 & 0.7353 & 26.04 & 0.7838 & 30.45 & 0.9075 \\
        LatticeNet~\cite{luo2020latticenet} & D2K & $\times 4$ & 43.6G & 777K & 32.18 & 0.8943 & 28.61 & 0.7812 & 27.57 & 0.7355 & 26.14 & 0.7844 & 30.54 & 0.9075 \\
        RFDN-L~\cite{liu2020residual} & D2K & $\times 4$ & 37.4G & 643K & \underline{32.28} & \underline{0.8957} & 28.61 & 0.7818 & \underline{27.58} & 0.7363 & \underline{26.20} & 0.7883 & \underline{30.61} & 0.9096 \\
        SRPN-Lite~\cite{zhang2021learning} & DF2K & $\times 4$ & 35.8G & 623K & 32.24 & \textcolor{blue}{0.8958} & 28.69 & \underline{0.7836} & \textcolor{blue}{27.63} & 0.7373 & 26.16 & 0.7875 & - & - \\
        HNCT~\cite{fang2022hybrid} & D2K & $\times 4$ & 22.0G & 373K & \textcolor{blue}{32.31}  & \underline{0.8957}  & \textcolor{blue}{28.71}  & 0.7834  & \textcolor{blue}{27.63}  & \textcolor{blue}{0.7381}  & \underline{26.20}  & \underline{0.7896}  & \textcolor{blue}{30.70}  & \textcolor{blue}{0.9112} \\
        FMEN~\cite{du2022fast} & DF2K & $\times 4$ & 44.2G & 769K & 32.24  & 0.8955  & \underline{28.70}  & \textcolor{blue}{0.7839}  & \textcolor{blue}{27.63}  & \underline{0.7379}  & \textcolor{blue}{26.28}  & \textcolor{blue}{0.7908}  & \textcolor{blue}{30.70}  & \underline{0.9107} \\
        \hhline{|=|=|=|=|=|=|=|=|=|=|=|=|=|=|=|}
        {\bf NGswin (ours)} & {\bf D2K} & {\bf $\times 4$} &{\bf  36.4G} & {\bf 1,019K} & {\bf \textcolor{red}{32.33}} & {\bf \textcolor{red}{0.8963}} & {\bf \textcolor{red}{28.78}} & {\bf \textcolor{red}{0.7859}} & {\bf \textcolor{red}{27.66}} & {\bf \textcolor{red}{0.7396}} & {\bf \textcolor{red}{26.45}} & {\bf \textcolor{red}{0.7963}} & {\bf \textcolor{red}{30.80}} & {\bf \textcolor{red}{0.9128}} \\
        \hline
    \end{tabular}}
\end{table*}
\vspace{-5pt}

\begin{table*}[t]
    \caption{Comparison of state-of-the-art lightweight super-resolution results. SwinIR-NG is SwinIR-light improved with the N-Gram. The mark $\downarrow$ and $^\mathcal{x}$ indicates reduced-channel and DF2K, respectively. `Year' indicates the publication year of each paper. The best and second best results are in \textcolor{red}{red} and \textcolor{blue}{blue}.}
    \label{tab_sota_comp}
    \centering
    \resizebox{0.843\linewidth}{!}
    {\begin{tabular}{|c|c|c|c|c|c|c|c|c|c|c|c|c|c|c|}
        \hline
        \multirow{2}{*}{Method} & \multirow{2}{*}{Year} & \multirow{2}{*}{Scale} & \multirow{2}{*}{Mult-Adds} & \multirow{2}{*}{\#Params} & \multicolumn{2}{c|}{Set5} & \multicolumn{2}{c|}{Set14} & \multicolumn{2}{c|}{BSD100} & \multicolumn{2}{c|}{Urban100} & \multicolumn{2}{c|}{Manga109}  \\ \cline{6-15}
        &&&&&PSNR&SSIM&PSNR&SSIM&PSNR&SSIM&PSNR&SSIM&PSNR&SSIM \\
        \hline
        SwinIR-light~\cite{liang2021swinir} & 2021 & $\times 2$ & 243.7G & 910K & 38.14 & \textcolor{blue}{0.9611} & \textcolor{blue}{33.86} & \textcolor{blue}{0.9206} & \textcolor{red}{32.31} & \textcolor{blue}{0.9012} & \textcolor{blue}{32.76} & \textcolor{red}{0.9340} & \textcolor{blue}{39.12} & \textcolor{red}{0.9783} \\
        ESRT~\cite{lu2021efficient} & 2022 & $\times 2$ & 191.4G & 677K & 38.03 & 0.9600 & 33.75 & 0.9184 & 32.25 & 0.9001 & 32.58 & 0.9318 & \textcolor{blue}{39.12} & 0.9774 \\
        ELAN-light~\cite{zhang2022efficient} & 2022 & $\times 2$ & 168.4G & 582K & \textcolor{red}{38.17} & \textcolor{blue}{0.9611} & \textcolor{red}{33.94} & \textcolor{red}{0.9207} & \textcolor{blue}{32.30} & \textcolor{blue}{0.9012} & \textcolor{blue}{32.76} & \textcolor{red}{0.9340} & \textcolor{blue}{39.12} & \textcolor{red}{0.9783} \\
        DiVANet~\cite{behjati2023single} & 2023 & $\times 2$ & 189.0G & 902K & \textcolor{blue}{38.16} & \textcolor{red}{0.9612} & 33.80 & 0.9195 & 32.29 & \textcolor{blue}{0.9012} & 32.60 & \textcolor{blue}{0.9325} & 39.08 & 0.9775 \\
        \hhline{|=|=|=|=|=|=|=|=|=|=|=|=|=|=|=|}
        {\bf SwinIR-NG (ours)} & {\bf 2023} & {\bf $\times 2$} & {\bf 274.1G} & {\bf 1,181K} & {\bf \textcolor{red}{38.17}} & {\bf \textcolor{red}{0.9612}} & {\bf \textcolor{red}{33.94}} & {\bf 0.9205} & {\bf \textcolor{red}{32.31}} & {\bf \textcolor{red}{0.9013}} & {\bf \textcolor{red}{32.78}} & {\bf \textcolor{red}{0.9340}} & {\bf \textcolor{red}{39.20}} & {\bf \textcolor{blue}{0.9781}} \\
        \hline\hline
        SwinIR-light~\cite{liang2021swinir} & 2021 & $\times 3$ & 109.5G & 918K & \textcolor{blue}{34.62} & \textcolor{blue}{0.9289} & 30.54 & \textcolor{blue}{0.8463} & 29.20 & \textcolor{blue}{0.8082} & 28.66 & \textcolor{blue}{0.8624} & 33.98 & \textcolor{blue}{0.9478} \\
        ESRT~\cite{lu2021efficient} & 2022 & $\times 3$ & 96.4G & 770K & 34.42 & 0.9268 & 30.43 & 0.8433 & 29.15 & 0.8063 & 28.46 & 0.8574 & 33.95 & 0.9455 \\
        ELAN-light~\cite{zhang2022efficient} & 2022 & $\times 3$ & 75.7G & 590K & 34.61 & 0.9288 & \textcolor{blue}{30.55} & \textcolor{blue}{0.8463} & \textcolor{blue}{29.21} & 0.8081 & \textcolor{blue}{28.69} & \textcolor{blue}{0.8624} & \textcolor{blue}{34.00} & \textcolor{blue}{0.9478} \\
        DiVANet~\cite{behjati2023single} & 2023 & $\times 3$ & 89.0G & 949K & 34.60 & 0.9285 & 30.47 & 0.8447 & 29.19 & 0.8073 & 28.58 & 0.8603 & 33.94 & 0.9468 \\
        \hhline{|=|=|=|=|=|=|=|=|=|=|=|=|=|=|=|}
        {\bf SwinIR-NG (ours)} & {\bf 2023} & {\bf $\times 3$} & {\bf 114.1G} & {\bf 1,190K} & {\bf \textcolor{red}{34.64}} & {\bf \textcolor{red}{0.9293}} & {\bf \textcolor{red}{30.58}} & {\bf \textcolor{red}{0.8471}} & {\bf \textcolor{red}{29.24}} & {\bf \textcolor{red}{0.8090}} & {\bf \textcolor{red}{28.75}} & {\bf \textcolor{red}{0.8639}} & {\bf \textcolor{red}{34.22}} & {\bf \textcolor{red}{0.9488}} \\
        \hline\hline
        SwinIR-light~\cite{liang2021swinir} & 2021 & $\times 4$ & 61.7G & 930K & \textcolor{blue}{32.44} & 0.8976 & 28.77 & 0.7858 & 27.69 & 0.7406 & 26.47 & 0.7980 & 30.92 & 0.9151 \\
        ESRT~\cite{lu2021efficient} & 2022 & $\times 4$ & 67.7G & 751K & 32.19 & 0.8947 & 28.69 & 0.7833 & 27.69 & 0.7379 & 26.39 & 0.7962 & 30.75 & 0.9100 \\
        ELAN-light~\cite{zhang2022efficient} & 2022 & $\times 4$ & 43.2G & 601K & 32.43 & 0.8975 & 28.78 & 0.7858 & 27.69 & 0.7406 & \textcolor{blue}{26.54} & 0.7982 & 30.92 & 0.9150 \\
        DiVANet~\cite{behjati2023single} & 2023 & $\times 4$ & 57.0G & 939K & 32.41 & 0.8973 & 28.70 & 0.7844 & 27.65 & 0.7391 & 26.42 & 0.7958 & 30.73 & 0.9119 \\
        \hhline{|=|=|=|=|=|=|=|=|=|=|=|=|=|=|=|}
        SwinIR-NG$\downarrow$ (ours) & \multirow{3}{*}{{\bf 2023}} & \multirow{3}{*}{$\times 4$} & {\bf 42.5G} & {\bf 770K} & {\bf \textcolor{blue}{32.44}} & {\bf 0.8978} & {\bf \textcolor{blue}{28.80}} & {\bf 0.7863} & {\bf 27.70} & {\bf 0.7407} & {\bf 26.47} & {\bf 0.7977} & {\bf 30.97} & {\bf 0.9147} \\
        SwinIR-NG$\downarrow^\mathcal{x}$ (ours) & & & {\bf 42.5G} & {\bf 770K} & {\bf \textcolor{red}{32.48}} & {\bf \textcolor{blue}{0.8979}} & {\bf \textcolor{red}{28.83}} & {\bf \textcolor{blue}{0.7868}} & {\bf \textcolor{blue}{27.71}} & {\bf \textcolor{blue}{0.7411}} & {\bf \textcolor{blue}{26.54}} & {\bf \textcolor{blue}{0.7998}} & {\bf \textcolor{red}{31.12}} & {\bf \textcolor{blue}{0.9158}} \\
        {\bf SwinIR-NG (ours)} & & & {\bf 63.0G} & {\bf 1,201K} & {\bf \textcolor{blue}{32.44}} & {\bf \textcolor{red}{0.8980}} & {\bf \textcolor{red}{28.83}} & {\bf \textcolor{red}{0.7870}} & {\bf \textcolor{red}{27.73}} & {\bf \textcolor{red}{0.7418}} & {\bf \textcolor{red}{26.61}} & {\bf \textcolor{red}{0.8010}} & {\bf \textcolor{blue}{31.09}} & {\bf \textcolor{red}{0.9161}} \\
        \hline
    \end{tabular}}
\end{table*}

\noindent
{\bf Training.}
We used 800 HR-LR (high- and low-resolution) image pairs from DIV2K~\cite{agustsson2017ntire} dataset.
LR images were randomly cropped into 64$\times$64 size patches augmented by random horizontal flip and rotation ($90^{\circ}$, $180^{\circ}$, $270^{\circ}$), as in the recent works~\cite{liang2021swinir,fang2022hybrid,zhang2022efficient,behjati2023single}.
We minimized $L_1$ pixel-loss between $I_{SR}$ and the ground truth $I_{HR}$: $\mathcal{L} = \lVert I_{HR}-I_{SR} \rVert_1$, with Adam~\cite{kingma2014adam} or AdamW~\cite{loshchilov2016sgdr} optimizer.
NGswin, SwinIR-NG, and HNCT-NG (improved with N-Gram) were trained by the same strategies except warm-start~\cite{lin2022revisiting}.
While we trained NGswin and SwinIR-NG from scratch on $\times 2$ and by warm-start (using pre-trained $\times 2$ weights) on $\times 3$ and $\times 4$, HNCT-NG was trained from scratch on all tasks.
Other details are in the supplementary Sec\onedot\textcolor{red}{B}.

\noindent
{\bf Evaluation.}
We evaluated the performances of the different models on the five benchmark datasets, composed of Set5~\cite{bevilacqua2012low}, Set14~\cite{zeyde2010single}, BSD100~\cite{martin2001database}, Urban100~\cite{huang2015single}, and Manga109~\cite{matsui2017sketch}.
We used PSNR (dB) and SSIM~\cite{wang2004image} scores on the Y channel of the YCbCr space as the metrics.
LR images were acquired by the MATLAB bicubic kernel from corresponding HR images matching each SR task.
\vspace{-5pt}

\subsection{Comparisons of Super-Resolution Results}
\label{quant}

\begin{figure*}[t]
    \centering
    \includegraphics[width=0.83\linewidth]{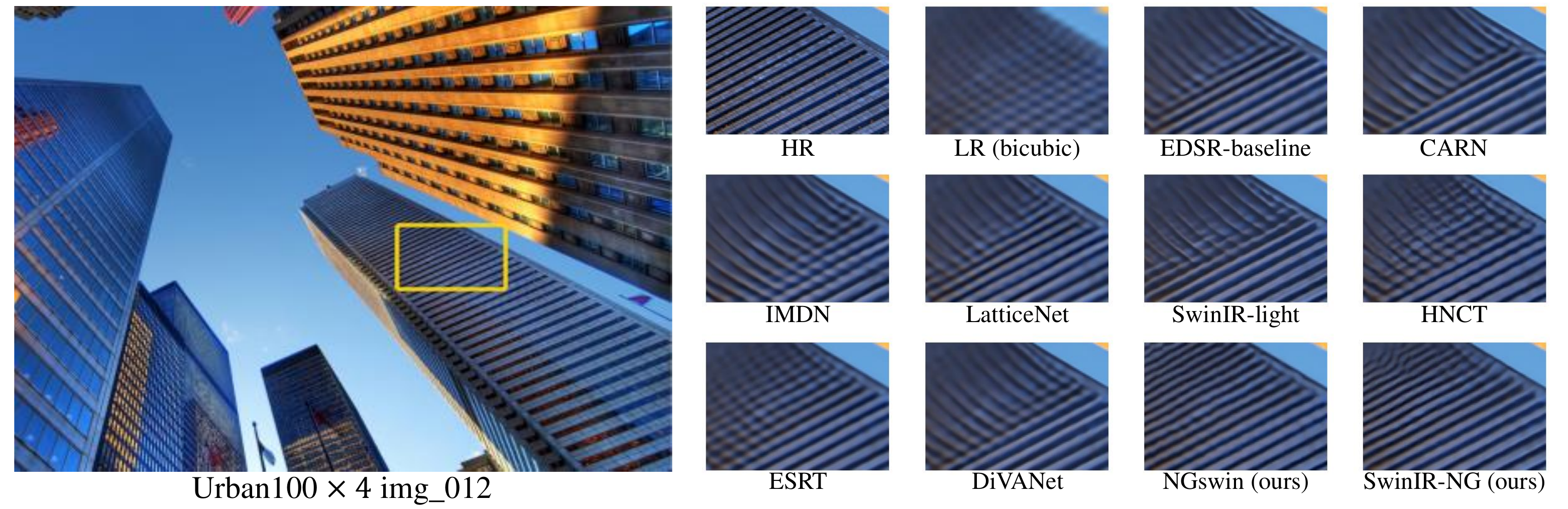} \\

    \caption{Visual comparisons with other models. `LR' is an input image. More results are illustrated in the supplementary Sec\onedot\textcolor{red}{D}.}
    \label{fig_vis_comp_main}
\end{figure*}

\begin{table}[t]
    \caption{Ablation study on the N-Gram context. The top and bottom tables are PSNR / SSIM of NGswin and HNCT, respectively.}
    \label{tab_ngram_context}
    \centering
    \resizebox{\linewidth}{!}
    {\begin{tabular}{c||c|c|c|c|c|c}
        \multicolumn{7}{c}{{\large NGswin without \vs with N-Gram}} \\
        \Xhline{2pt}
        N-Gram & Scale & Mult-Adds & \#Params & Set14 & Urban100 & Manga109 \\
        \hline
        \textit{w/o} & \multirow{2}{*}{$\times2$} & 138.20G & 750K & 33.70 / 0.9194 & 32.39 / 0.9304 & 38.86 / 0.9775 \\
        {\bf \textit{w/}} & & {\bf 140.41G} & {\bf 998K} & {\bf 33.79} / {\bf 0.9199} & {\bf 32.53} / {\bf 0.9324} & {\bf 38.97} / {\bf 0.9777} \\
        \hline
        \textit{w/o} & \multirow{2}{*}{$\times3$} & 65.53G & 759K & 30.48 / 0.8451 & 28.37 / 0.8573 & 33.81 / 0.9464 \\
        {\bf \textit{w/}} & & {\bf 66.56G} & {\bf 1,007K} & {\bf 30.53} / {\bf 0.8456} & {\bf 28.52} / {\bf 0.8603} & {\bf 33.89} / {\bf 0.9470} \\
        \hline
        \textit{w/o} & \multirow{4}{*}{$\times4$} & 35.89G & 771K & 28.70 / 0.7844 & 26.25 / 0.7918 & 30.70 / 0.9123 \\
        \textit{w/o} (channel up) & & 53.71G & 1,189K & 28.75 / 0.7854 & 26.28 / 0.7927 & 30.73 / 0.9129 \\
        \textit{w/o} (depth up) & & 47.88G & 1,061K & 28.75 / 0.7853 & 26.37 / 0.7946 & 30.78 / {\bf 0.9133} \\
        {\bf \textit{w/}} & & {\bf 36.44G} & {\bf 1,019K} & {\bf 28.78} / {\bf 0.7859} & {\bf 26.45} / {\bf 0.7963} & {\bf 30.80} / 0.9128 \\
        \Xhline{2pt}
        \multicolumn{7}{c}{} \\
        \multicolumn{7}{c}{{\large HNCT~\cite{fang2022hybrid} \vs HNCT-NG (ours)}} \\
        \Xhline{2pt}
        N-Gram & Scale & Mult-Adds & \#Params & Set14 & Urban100 & Manga109 \\
        \hline
        \textit{w/o} & \multirow{2}{*}{$\times2$} & 82.39G & 357K & {\bf 33.65} / 0.9182 & 32.22 / 0.9294 & 38.87 / {\bf 0.9774} \\
        {\bf \textit{w/}} & & {\bf 83.19G} & {\bf 424K} & 33.64 / {\bf 0.9195} & {\bf 32.35} / {\bf 0.9306} & {\bf 38.94} / {\bf 0.9774} \\
        \hline
        \textit{w/o} & \multirow{2}{*}{$\times3$} & 37.78G & 363K & 30.44 / 0.8439 & 28.28 / 0.8557 & {\bf 33.81} / 0.9459 \\
        {\bf \textit{w/}} & & {\bf 38.14G} & {\bf 431K} & {\bf 30.48} / {\bf 0.8450} & {\bf 28.38} / {\bf 0.8573} & {\bf 33.81} / {\bf 0.9464} \\
        \hline
        \textit{w/o} & \multirow{2}{*}{$\times4$} & 22.01G & 373K & 28.71 / 0.7834 & 26.20 / 0.7896 & 30.70 / 0.9112 \\
        {\bf \textit{w/}} & & {\bf 22.21G} & {\bf 440K} & {\bf 28.72} / {\bf 0.7846} & {\bf 26.23} / {\bf 0.7912} & {\bf 30.71} / {\bf 0.9114} \\
        \Xhline{2pt}
    \end{tabular}}
\end{table}

In~\cref{tab_light_comp}, we compare NGswin with other efficient SR models, including 
EDSR-baseline (CVPRW17)~\cite{lim2017enhanced}, 
MemNet (ICCV17)~\cite{tai2017memnet}, 
CARN (ECCV18)~\cite{ahn2018fast}, 
IMDN (ACMMM19)~\cite{hui2019lightweight}, 
LatticeNet (ECCV20)~\cite{luo2020latticenet}, 
RFDN-L (ECCV20)~\cite{liu2020residual}, 
SRPN-Lite (ICLR22)~\cite{zhang2021learning}, 
HNCT (CVPRW22)~\cite{fang2022hybrid}, 
and FMEN (CVPRW22)~\cite{du2022fast}.
PSNR and SSIM were evaluated on the three SR tasks.
We reported the training dataset, Mult-Adds, and the number of parameters for comparing efficiency.
The result shows that NGswin outperformed previous leading models on all benchmarks with a relatively efficient structure.
Compared to SRPN-Lite and FMEN, NGswin was data-efficient with PSNR margins up to 0.3dB and 0.19dB, respectively.
Note that the results of LatticeNet were referred from~\cite{luo2022lattice}.

As shown in~\cref{tab_sota_comp}, we improved SwinIR-light~\cite{liang2021swinir} with the N-Gram context (named as SwinIR-NG).
It was compared with the current best lightweight SR methods, including SwinIR-light (ICCVW21), ESRT (CVPRW22)~\cite{lu2021efficient}, ELAN-light (ECCV22)~\cite{zhang2022efficient}, and DiVANet (PR23)~\cite{behjati2023single}, all of which were trained on DIV2K.
SwinIR-NG outperformed them and established state-of-the-art lightweight SR on all benchmarks.
Since the restoration of highly distorted regions needs much neighbor information, the impact of the N-Gram context was especially strong for $\times 3$ or $\times 4$ tasks and Urban100 or Manga109 datasets by a PSNR margin up to 0.24dB, compared to \textit{w/o} N-Gram.
For fair comparison with respect to \#parameters, we also reduced the channels of SwinIR-NG from 60 to 48, which is denoted as $\downarrow$.
In addition, SwinIR-NG$\downarrow$ were also trained on DF2K (denoted as $^\mathcal{x}$) for further comparison.
The performance of the reduced model on $\times 4$ were reported and still better than the others with the fewest computations.
We corrected the parameters of~\cite{liang2021swinir} that omitted the relative position bias tables.

The visual comparisons of each model are in~\cref{fig_vis_comp_main}.

\begin{table}[t]
    \caption{Ablation study on N-Gram interaction. \enquote{Direction}: how many directions the network see N-Gram neighbors from. \enquote{Type}: the method for N-Gram interaction. The bottom row is a default setting. PSNR / SSIM are evaluated on $\times2$ task with NGswin.}
    \label{tab_ablation_direction_interaction}
    \centering
    \resizebox{\linewidth}{!}
    {\begin{tabular}{c||c||c|c|c|c}
    \Xhline{2pt}
        Direction & Type & Mult-Adds & \#Params & Urban100 &  Manga109 \\
        \hline
        1 & WSA & 152.41G & 1,238,056 & {\bf 32.54} / 0.9322 & 38.90 / {\bf 0.9777}  \\
        4 & WSA & 139.56G & 935,272 & 32.52 / 0.9317 & 38.92 / 0.9776 \\
        1 & CNN & 139.80G & 1,327,528 & 32.45 / 0.9316 & 38.86 / 0.9775 \\
        2 & CNN & 139.38G & 998,568 & {\bf 32.54} / 0.9321 & 38.90 / 0.9776 \\
        4 & CNN & 139.17G & 936,488 & 32.52 / 0.9320 & 38.93 / {\bf 0.9777} \\
        \hhline{===|=|=|=}
        {\bf 2} & {\bf WSA} & {\bf 140.41G} & {\bf 998,384} & 32.53 / {\bf 0.9324} & {\bf 38.97} / {\bf 0.9777} \\
    \Xhline{2pt}
    \end{tabular}}
\end{table}

\begin{figure*}[t]
    \centering
    \includegraphics[width=0.79\linewidth]{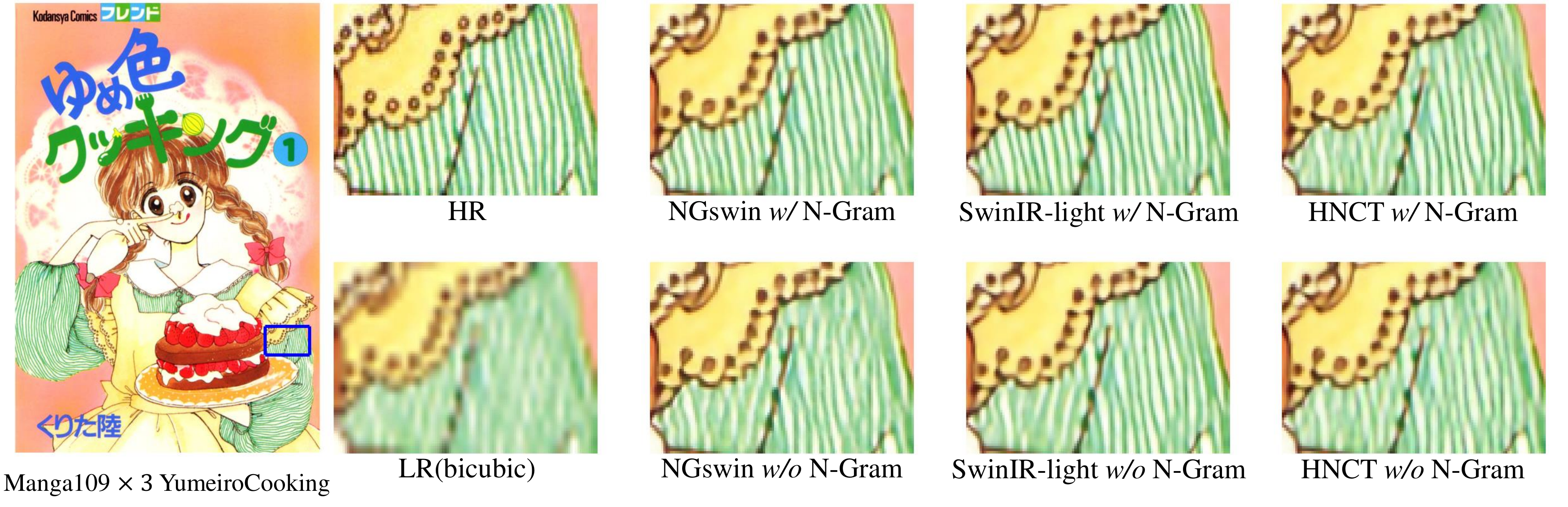}
    \caption{Visual comparisons of \textit{w/} \vs \textit{w/o} N-Gram context for NGswin, SwinIR-light, and HNCT. At the top and bottom row, the 2nd to 4th columns show the models with and without N-Gram, respectively. More visual comparisons are in the supplementary Sec\onedot\textcolor{red}{D}.}
    \label{fig_ngram_use}
\end{figure*}

\subsection{Ablation Studies}
\label{ablation}

\cref{tab_ngram_context} demonstrates that the N-Gram context enhanced Swin Transformer-based SR models with a reasonable level of sacrificed efficiency.
We denote HNCT~\cite{fang2022hybrid} improved with the N-Gram context as HNCT-NG.
The results of SwinIR-light~\cite{liang2021swinir} and a corresponding SwinIR-NG can be referred to in~\cref{tab_sota_comp}.
Interestingly, the N-Gram increased SSIM in general.
That is, our method tended to produce perceptually more similar images to the ground-truth.
Moreover, our method was robust to Urban100 and Manga109 datasets, which are hard to recover with DIV2K training dataset~\cite{magid2022texture}.
\cref{fig_ngram_use} visualizes the examples.
One can doubt the marginal gain of HNCT-NG.
This was because HNCT contains 8 Swin Transformer layers (Swins), while NGswin had 20.
The application of the N-Gram method on fewer Swins of HNCT-NG yielded less effective results.
Afterwards, we increased the depth or channel of NGswin without N-Gram for more compelling comparisons regarding to the number of parameters.
The variations were applied to $\times 4$ task, which still fell behind our proposed NGswin despite similar parameters but much more computations.

In~\cref{tab_ablation_direction_interaction}, we examined the N-Gram interaction.
\enquote{Direction} counts how many directions the network sees N-Gram neighbors from.
\enquote{Type} indicates the method for N-Gram interaction.
The two directions and WSA type were the same as explained in~\cref{NSTB}.
At one direction, the network saw the neighbors from a lower-right only.
The uni-Gram embedding never reduced channels, which was harmful for the efficiency.
In the case of four directions, neighbors from lower-right, lower-left, upper-right, and upper-left were viewed as N-Gram.
In implementation, uni-Gram embedding layer produced the four N-Gram features that have $\frac{D}{4}$ dimension.
They were merged into $D$ dimension of the N-Gram context.
For CNN interaction, we operated a conventional convolution after uni-Gram embedding  rather than \textit{sliding-WSA}.
As a result, the bi-directional N-Gram context with \textit{sliding-WSA} model was the best optimized model in terms of trade-off between performance and efficiency.
In addition, \textit{sliding-WSA} had an advantage over sliding-winodw-convolution in that SA could compute the correlations within each N-Gram.

\begin{table}[t]
    \caption{Ablation study on extra stages and SCDP bottleneck.}
    \vspace{-2.5pt}
    \label{tab_ablation_scdp}
    \centering
    \begin{subtable}[h]{\linewidth}
    \caption{The specifications of models with different stages. dep.: \# of NSTBs / res.: training input resolution. The total number of NSTBs is kept as 20.}
    \vspace{-2.5pt}
    \resizebox{\linewidth}{!}
    {
    \begin{tabular}{c|c|c|c|c|c|c}
    \Xhline{2pt}
        \multirow{2}{*}{Stages} & encoder1 & encoder2 & encoder3 & encoder4 & decoder1 & decoder2 \\ 
        \cline{2-7}
        & dep. / res. & dep. / res. & dep. / res. & dep. / res. & dep. / res. & dep. / res. \\
        \hline
        extra & 4 / 64$\times$64 & 4 / 32$\times$32 & 4 / 16$\times$16 & 4 / 8$\times$8 & 2 / 32$\times$32 & 2 / 64$\times$64 \\
        {\bf default }& {\bf 6 / 64$\times$64} & {\bf 4 / 32$\times$32} & {\bf 4 / 16$\times$16} & {\bf - / -} & {\bf 6 / 64$\times$64} & {\bf - / -} \\
    \Xhline{2pt}
    \end{tabular}
    }
    \vspace{3pt}
    \end{subtable}

    \begin{subtable}[h]{\linewidth}
    \caption{Impacts of extra stages and SCDP bottleneck. PSNR / SSIM.}
    \vspace{-2.5pt}
    \centering
    \resizebox{\linewidth}{!}
    {
    \begin{tabular}{c||c||c|c|c|c|c}
    \Xhline{2pt}
        Stages & SCDP & Scale & Mult-Adds & \#Params & Urban100 & Manga109 \\
        \hline
        extra & \textit{w/o} & \multirow{3}{*}{$\times2$} & 87.98G & 997K & 32.28 / 0.9298 & 38.72 / 0.9773 \\
        default & \textit{w/o} & & 138.88G & 992K & 32.48 / 0.9321 & 38.92 / 0.9776 \\
        {\bf default} & {\bf \textit{w/}} & & {\bf 140.41G} & {\bf 998K} & {\bf 32.53} / {\bf 0.9324} & {\bf 38.97} / {\bf 0.9777} \\
        \hline
        extra & \textit{w/o} & \multirow{3}{*}{$\times3$} & 42.10G & 1,006K & 28.33 / 0.8562 & 33.67 / 0.9453 \\
        default & \textit{w/o} & & 65.85G & 1,001K  & 28.47 / 0.8596 & 33.81 / 0.9464 \\
        {\bf default} & {\bf \textit{w/}} & & {\bf 66.56G} & {\bf 1,007K} & {\bf 28.52} / {\bf 0.8603} & {\bf 33.89} / {\bf 0.9470} \\
        \hline
        extra & \textit{w/o} & \multirow{3}{*}{$\times4$} & 23.33G & 1,018K & 26.22 / 0.7900 & 30.46 / 0.9090 \\
        default & \textit{w/o} & & 36.06G & 1,013K & 26.38 / 0.7954 & 30.71 / 0.9121 \\
        {\bf default} & {\bf \textit{w/}} & & {\bf 36.44G} & {\bf 1,019K} & {\bf 26.45} / {\bf 0.7963} & {\bf 30.80} / {\bf 0.9128} \\
    \Xhline{2pt}
    \end{tabular}
    }
    \end{subtable}
\end{table}

As shown in~\cref{tab_ablation_scdp}, while both default and extra stages had the same number of NSTBs, the extra stages handled lower resolutions.
But the extra stages appended to the encoder and decoder dropped the performance.
This is because reconstruction of high-frequency information from the space preserving HR information richly is easier than from the space preserving HR information insufficiently.
Since our SCDP bottleneck took all outputs of the encoder stages (\emph{i.e}\onedot, various resolutions), we prevented the performance drop despite hierarchical encoders.
In implementation, the bottleneck without SCDP had the depth-wise and point-wise convolutions only, and took the output of the last NSTB in the third encoder stage.

Each ablation study on the other benchmarks not reported due to page limit are in the appendix Sec\onedot\textcolor{red}{C}.
\vspace{-7pt}

\section{Conclusion}
\label{conclusion}
This paper successfully introduced the N-Gram context from the text to the vision domain for the first time in history.
Our N-Gram interaction by \textit{sliding-WSA} made NGswin, SwinIR, and HNCT overcome the limitation of Swin Transformer, where the broad regions are ignored.
SCDP bottleneck prevented the hierarchical encoder from dropping performance.
The hierarchical encoder, small decoder, and uni-Gram embedding decreased the operations significantly.
With the components above, NGswin showed competitive results compared with the previous leading SR methods.
Moreover, SwinIR-NG established state-of-the-art results.
For future works, we hope our N-Gram context can succeed on other low-level vision tasks, such as denoising and deblurring.
In closing, if the N-Gram context is extended to the universal Transformer architectures, more developments for computer vision could be expected.
\newline
\noindent
{\bf Acknowledgements.} This paper was supported by Institute of Information \& Communications Technology Planning \& Evaluation (IITP) grant (No.2022-0-00956) and Korea Health Industry Development
Institute (KHIDI) grant (No. H122C1983) funded by the Korea government (MSIT). Special thanks to Rescale Inc\onedot for HPC cloud computing resources and Somang Kim for valuable comments.

\setcounter{section}{0}
\setcounter{table}{0}
\setcounter{figure}{0}
\renewcommand\thesection{\Alph{section}}
\renewcommand\thetable{\Alph{table}}
\renewcommand\thefigure{\Alph{figure}}

\section{Details of Other Components in NGswin}
\label{supp_components}

\subsection{Sequentially Reflected Window Padding}
\label{seq_refl_win_pad}
\begin{figure}[h]
    \centering
    \includegraphics[width=\linewidth]{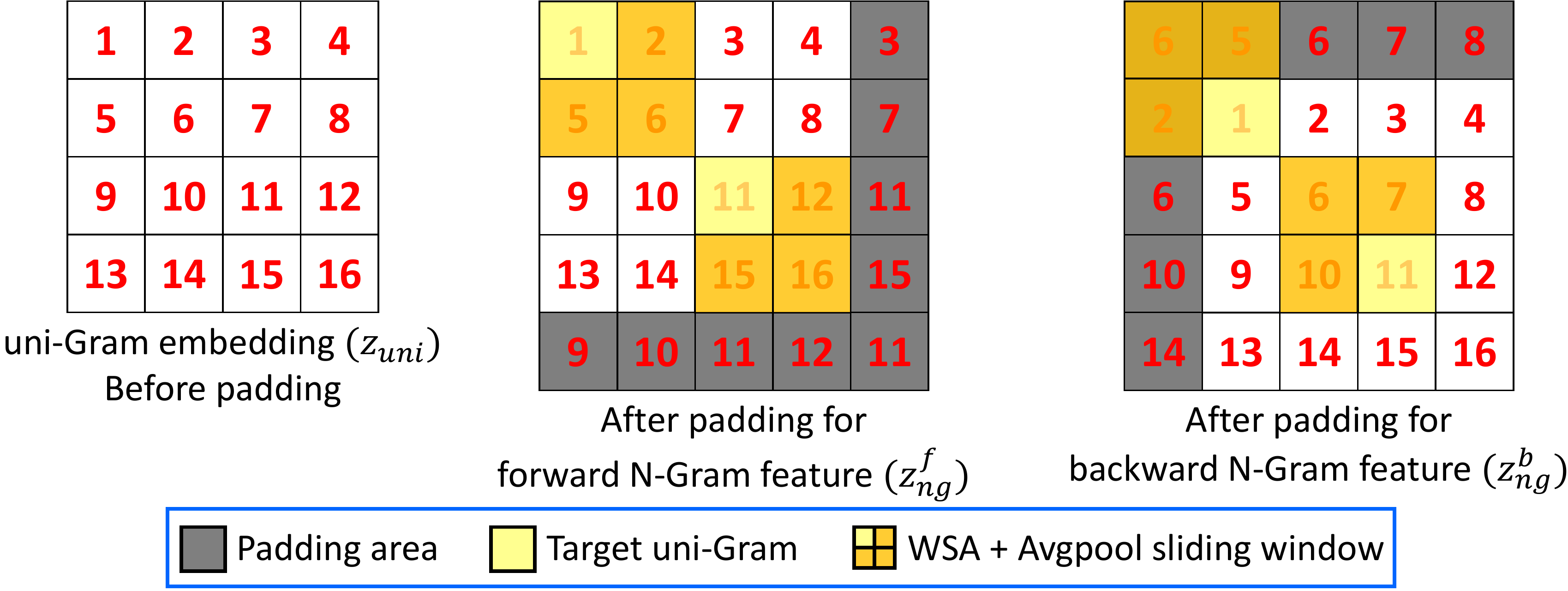}
    \caption{Sequentially reflected window padding. N-Gram size $N$ is 2. {\bf (Left)} The uni-Gram embedding before padding. {\bf (Middle)} Padding for the forward N-Gram feature $z^f_{ng}$. {\bf (Right)} Padding for the backward N-Gram feature $z^b_{ng}$. As stated in~\cref{NSTB}, \textit{sliding-WSA} weights for bi-directional N-Gram features are shared.}
    \label{fig_seq_refl}
\end{figure}

As illustrated in the middle of~\cref{fig_seq_refl}, ($N-1$) size of paddings are applied at the lower-right side of uni-Gram embedding $z_{uni}$ by sequentially reflected window padding (\textit{seq-refl-win-pad}).
Based on the outermost low/right windows, we use the upper/left ($N-1$) rows/columns of windows as padding values.
Consequently, \textit{Sliding-WSA} produces the forward N-Gram feature $z^f_{ng}$.
In turn, we can get the backward N-Gram feature $z^b_{ng}$ by simply applying the same size of paddings on the upper-left side, as in the right of the figure.
This allows some uni-Grams to interact with their padded neighbors, instead of trivial \enquote{zero} padding values.
Our \textit{seq-refl-win-pad} does not require extra Mult-Adds operations, because both zero padding and our padding method additionally give the same number of 32-bit float data to the input feature maps.
We emphasize the advantage of \textit{seq-refl-win-pad} in~\cref{other_ablation}.

\subsection{Within-Stage Residual Connections}
\label{within_stage}
\begin{figure}[t]
    \centering
    \includegraphics[width=0.93\linewidth]{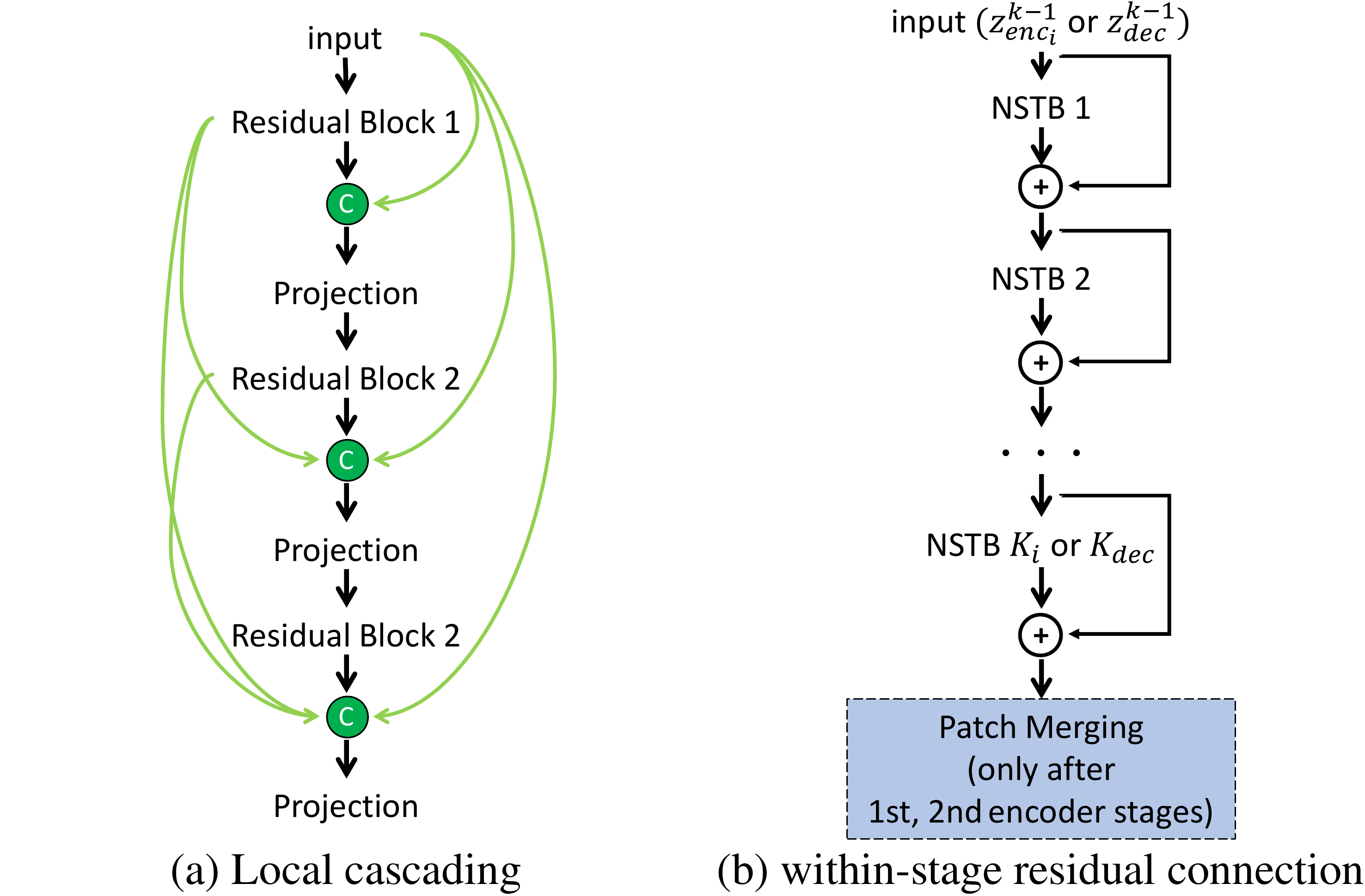}
    \caption{Comparison of local cascading and within-stage residual connections. {\bf (a)} In local cascading of CARN, the resiudal blocks are densely connected. {\bf (b)} In our within-stage residual connection, NSTBs and \textit{patch-merging} are residually connected.}
    \label{fig_within}
\end{figure}

While our across-stage pooling cascading follows the global cascading of CARN~\cite{ahn2018fast}, the elements within a stage are residually connected~\cite{he2016deep}.
Each NSTB and a \textit{patch-merging} layer (except the third encoder stage and the decoder stage) in the encoder and decoder stages are residually connected.
As described in~\cref{fig_within}, this differs from the local cascading of CARN that employed dense connections~\cite{huang2015single}.
Note that we did not specifically state the input to the decoder in the main content.
$z^{k-1}_{dec}$ in the figure is the input to $k$-$th$ ($1$$\leq$$k$$\leq$$\mathcal{K}_{dec}$) NSTB in the decoder stage.
The corresponding mapping function $\mathcal{F}^k_{dec}$ is formulated as:
\begin{equation*}
\label{eq_decoder}
    z^k_{dec} = \mathcal{F}^k_{dec}(z^{k-1}_{dec}),\ z^{k}_{dec} \in \mathbb{R}^{HW \times D},
\end{equation*}
where $z^0_{dec}$ equals $z_{scdp} + z^{\mathcal{K}_1}_{enc_1}$ from SCDP bottleneck and the last NSTB in the first encoder stage (\cref{SCDP}).
Also, $z_{dec} = \mathrm{LayerNorm}(z^{\mathcal{K}_{dec}}_{dec}) \in \mathbb{R}^{HW \times D}$.

\subsection{Reconstruction Module}
\label{reconstruction}
\begin{figure}[t]
    \centering
    \includegraphics[width=0.67\linewidth]{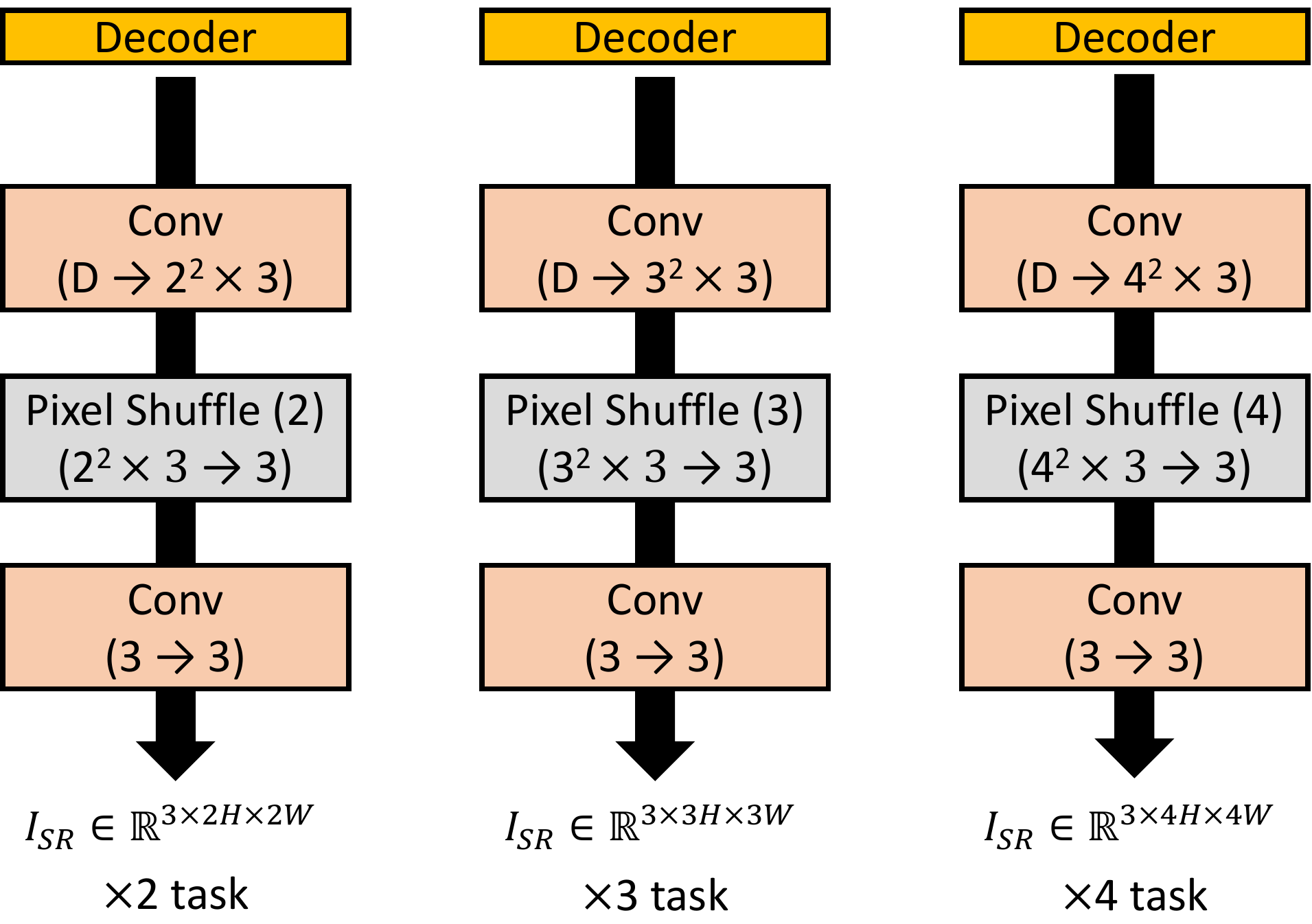}
    \caption{Comparison of the reconstruction modules for different SR tasks. A parenthesis ($a \rightarrow b$) indicates change of channels (network dimension) from $a$ to $b$. The other parenthesis ($r$) in Pixel Shuffle block indicates a scale factor (\eg, $\times4$).}
    \label{fig_reconstruction}
\end{figure}

The only difference between the $\times2$, $\times3$, and $\times 4$ models is the reconstruction module.
As depicted in~\cref{fig_reconstruction}, we vary the output channels of the first convolution and the scale factor of the pixel-shuffler.
The last convolutional layer is the difference from other methods~\cite{kim2016accurate,lim2017enhanced,ahn2018fast,liang2021swinir}, as previously mentioned in~\cref{overall_architecture}.
The input to this module is $z_s + z_{dec}$ with global skip-connection as stated in the main content ($z_s$ results from the shallow module).

\section{Experimental Setup Details and Findings}
\label{supp_experiment}
In this section, we explain experimental settings and our findings from the results of various learning strategies.
Since there are not an abundance of studies primarily focusing on training strategy itself for SR, we hope future researchers are able to gain insight from our findings.
We summarize our findings in~\cref{tab_findings}.
Although our findings in this section are not absolute truths, they can be helpful considerations for future research.

\noindent
{\bf Model Architecture.}
The number of NSTBs in the encoder and decoder, $\{\mathcal{K}_1,\mathcal{K}_2,\mathcal{K}_3,\mathcal{K}_{dec}\}$, is set to $\{6,4,4,6\}$.
The number of WSA heads (for \textit{sliding-WSA} and Swin Transformer's WSA) in each stage equals $\{\mathcal{K}_1,\mathcal{K}_2,\mathcal{K}_3,\mathcal{K}_{dec}\}$.
We set the network dimension (channel) $D$, hidden dimension of FFN (feed-forward network) after Swin Transformer's WSA, window size $M$, and N-Gram size $N$ to 64, 128, 8, and 2, respectively.
The shift size is 4, \emph{i.e}\onedot $\lfloor{\frac{M}{2}}\rfloor$ same as in Swin V1 and V2~\cite{liu2021swin,liu2022swin}.
The activation functions in FFN and after depth-wise convolution of SCDP bottleneck are GELU non-linearity.
Also, we use LeakyReLU non-linearity after the iterative max-poolings of the pixel-shuffle~\cite{shi2016real} step in the bottleneck.
For the other components not mentioned, there are no activation functions.

\noindent
{\bf Training Details.}
We implemented the model configurations, training pipeline, and evaluation procedure by PyTorch~\cite{paszke2019pytorch} on 4 NVIDIA TITAN Xp GPUs.
The batch size and training epochs were 64 and 500.
We used Adam~\cite{kingma2014adam} optimizer with $\{\beta_1, \beta_2, \epsilon\}=\{0.9, 0.999, 10^{-8}\}$ for training from scratch ($\times 2$ task) and warm-start before whole fine-tuning ($\times 3$, $\times 4$ tasks).
For whole fine-tuning phase, AdamW~\cite{loshchilov2017decoupled} was utilized with the same hyper-parameters above.
The learning rate (\textit{lr}) was initialized as $0.0004$ and decayed by half (half-decay) after $\{200, 300, 400, 425, 450, 475\}$ epochs.
At the start of the training, we placed $20$ warmup epochs~\cite{goyal2017accurate} that linearly increased \textit{lr} from $0.0$ to initialized \textit{lr} ($10^{-4}$).

\noindent
{\bf Warm-Start.}
We trained NGswin and SwinIR-NG from scratch ($\times2$) and by warm-start ($\times3$, $\times4$)~\cite{lin2022revisiting}, as mentioned in~\cref{setup}.
The warm-start scheme lasts for 300 epochs.
This strategy, therefore, needed short training times.
Warm-start was processed as follows:
Loading the pre-trained weight on $\times2$, we froze all layers except the reconstruction module, and trained this module for 50 epochs (warm-start epoch).
In this phase, \textit{lr} was kept as a constant (\emph{i.e}\onedot, $0.0004$).
After that, the whole parameters of the network were fine-tuned by back-propagation (whole fine-tuning) for 250 epochs.
We placed 10 warmup epochs at the start of whole fine-tuning.
In whole fine-tuning, \textit{lr} was halved after $\{50, 100, 150, 175, 200, 225\}$ epochs.
We compared SwinIR-NG trained by warm-start scheme to the scratch one in~\cref{tab_warmstart} to show the merits of this strategy.

\noindent
{\bf Dataset.}
We never used any extra datasets other than 800 images from DIV2K~\cite{agustsson2017ntire}.
Each data point in the training dataset was repeated 80 times in an epoch to maximize the merits of random-cropping ($64$$\times$$64$), following ELAN~\cite{zhang2022efficient}.
The random horizontal flip and rotation of $90^{\circ}$, $180^{\circ}$, $270^{\circ}$ augmented the training data.
We converted all images including the test data to \textit{\enquote{.npy}} (numpy) files with the \textit{uint8} data type, for faster loading and efficient memory usage.

\noindent
{\bf Normalization.}
We normalized the training data using the means and standard-deviations (std) of 800 LR images on RGB channels matching each task.
Expressly, it was the same as standardization.
The outputs of the reconstruction module were de-normalized (inverse of normalization), and used for calculating $L_1$ pixel-loss.
Although we trained the models with different normalization strategies such as 1.0 std and not de-normalizing before computing loss, those strategies fell behind the default strategy.
On the other hand, SwinIR-NG was trained with $1.0$ std, following SwinIR paper~\cite{liang2021swinir}.

\begin{table}[t]
    \centering
    \caption{Summary of learning strategies we find performed better with NGswin. Our findings are not absolute truths but just suggestions for the future works.}
    \label{tab_findings}
    \resizebox{0.76\linewidth}{!}
    {\begin{tabular}{rll}
        \Xhline{2pt}
        Method & Better & Worse \\
        \hline
        $\times3$, $\times4$ training & warm-start~\cite{lin2022revisiting} & scratch \\
        std in normalization & from data~\cite{he2022masked} & $1.0$~\cite{lim2017enhanced,liang2021swinir} \\
        de-normalization position & before loss~\cite{lim2017enhanced} & after loss \\
        \hhline{---}
        \textit{lr} decay & half~\cite{lim2017enhanced} & cosine~\cite{loshchilov2016sgdr} \\
        \hhline{---}
        weight decay~\cite{loshchilov2017decoupled} & no & yes \\
        gradient clipping~\cite{pascanu2012understanding,pascanu2013difficulty} & no & yes \\
        layer-wise \textit{lr} decay~\cite{clark2020electra} & no & yes \\
        \hhline{---}
        dropout~\cite{hinton2012improving} & no & yes \\
        drop-path~\cite{larsson2016fractalnet} & no & yes \\
        \Xhline{2pt}
    \end{tabular}}
\end{table}

\begin{table*}[t]
    \caption{Other ablation studies.}
    \begin{subtable}[h]{\linewidth}
    \caption{Ablation study on Swin Transformer version.}
    \label{tab_swin_version}
    \centering
    \resizebox{0.63\linewidth}{!}
    {\begin{tabular}{c||c|c|c|c|c|c|c|c}
    \hline
        Swin ver. & Scale & Mult-Adds & \#Params & Set5 & Set14 & BSD100 & Urban100 & Manga109 \\
        \hline
        V1 & \multirow{2}{*}{$\times2$} & \multirow{2}{*}{140.41G} & 998,176 & 37.99 / 0.9606 & 33.71 / 0.9192 & 32.20 / 0.9000 & 32.28 / 0.9301 & 38.69 / 0.9770 \\
        V2 & & & 998,384 & {\bf 38.05} / {\bf 0.9610} & {\bf 33.79} / {\bf 0.9199} & {\bf 32.27} / {\bf 0.9008} & {\bf 32.53} / {\bf 0.9324} & {\bf 38.97} / {\bf 0.9777} \\
        \hline
        V1 & \multirow{2}{*}{$\times3$} & \multirow{2}{*}{66.56G} & 1,006,831 & 34.42 / 0.9273 & 30.44 / 0.8445 & 29.13 / 0.8066 & 28.35 / 0.8569 & 33.66 / 0.9456 \\
        V2 & & & 1,007,039 & {\bf 34.52} / {\bf 0.9282} & {\bf 30.53} / {\bf 0.8456} & {\bf 29.19} / {\bf 0.8078} & {\bf 28.52} / {\bf 0.8603} & {\bf 33.89} / {\bf 0.9470} \\
        \hline
        V1 & \multirow{2}{*}{$\times4$} & \multirow{2}{*}{36.44G} & 1,018,948 & 32.20 / 0.8946 & 28.69 / 0.7836 & 27.61 / 0.7380 & 26.26 / 0.7916 & 30.53 / 0.9090 \\
        V2 & & & 1,019,156 & {\bf 32.33} / {\bf 0.8963} & {\bf 28.78} / {\bf 0.7859} & {\bf 27.66} / {\bf 0.7396} & {\bf 26.45} / {\bf 0.7963} & {\bf 30.80} / {\bf 0.9128} \\
    \hline
    \end{tabular}}
    \end{subtable}
    \begin{subtable}[h]{\linewidth}
    \vspace{8pt}
    \caption{Ablation study on padding method. The comparative model is NGswin.}
    \label{tab_seq_refl}
    \centering
    \resizebox{0.63\linewidth}{!}
    {\begin{tabular}{c||c|c|c|c|c|c}
    \hline
        Method & Scale & Set5 & Set14 & BSD100 & Urban100 & Manga109 \\
        \hline
        zero-pad & \multirow{2}{*}{$\times2$} & 37.82 / 0.9599 & 33.38 / 0.9160 & 32.06 / 0.8983 & 31.58 / 0.9231 & 38.10 / 0.9759 \\
        \textit{seq-refl-win-pad} & & {\bf 38.05} / {\bf 0.9610} & {\bf 33.79} / {\bf 0.9199} & {\bf 32.27} / {\bf 0.9008} & {\bf 32.53} / {\bf 0.9324} & {\bf 38.97} / {\bf 0.9777} \\
        \hline
        zero-pad & \multirow{2}{*}{$\times3$} & 34.14 / 0.9249 & 30.22 / 0.8400 & 28.99 / 0.8030 & 27.74 / 0.8439 & 33.01 / 0.9410 \\
        \textit{seq-refl-win-pad} & & {\bf 34.52} / {\bf 0.9282} & {\bf 30.53} / {\bf 0.8456} & {\bf 29.19} / {\bf 0.8078} & {\bf 28.52} / {\bf 0.8603} & {\bf 33.89} / {\bf 0.9470} \\
        \hline
        zero-pad & \multirow{2}{*}{$\times4$} & 31.90 / 0.8906 & 28.45 / 0.7782 & 27.47 / 0.7332 & 25.72 / 0.7740 & 29.89 / 0.9016 \\
        \textit{seq-refl-win-pad} & & {\bf 32.33} / {\bf 0.8963} & {\bf 28.78} / {\bf 0.7859} & {\bf 27.66} / {\bf 0.7396} & {\bf 26.45} / {\bf 0.7963} & {\bf 30.80} / {\bf 0.9128} \\
    \hline
    \end{tabular}}
    \end{subtable}
    \begin{subtable}[h]{\linewidth}
    \vspace{8pt}
    \caption{Ablation study on warm-start. The comparative model is SwinIR-NG.}
    \label{tab_warmstart}
    \centering
    \resizebox{0.63\linewidth}{!}
    {\begin{tabular}{c||c|c|c|c|c|c}
    \hline
        Method & Scale & Set5 & Set14 & BSD100 & Urban100 & Manga109 \\
        \hline
        scratch & \multirow{2}{*}{$\times3$} & {\bf 34.65} / 0.9291 & {\bf 30.59} / {\bf 0.8471} & 29.23 / {\bf 0.8090} & 28.71 / 0.8636 & 34.17 / 0.9485 \\
        warm-start & & 34.64 / {\bf 0.9293} & 30.58 / {\bf 0.8471} & {\bf 29.24} / {\bf 0.8090} & {\bf 28.75} / {\bf 0.8639} & {\bf 34.22} / {\bf 0.9488} \\
        \hline
        scratch & \multirow{2}{*}{$\times4$} & {\bf 32.45} / 0.8979 & 28.80 / 0.7867 & 27.71 / 0.7413 & 26.51 / 0.7992 & 31.02 / 0.9158 \\
        warm-start & & 32.44 / {\bf 0.8980} & {\bf 28.83} / {\bf 0.7870} & {\bf 27.73} / {\bf 0.7418} & {\bf 26.61} / {\bf 0.8010} & {\bf 31.09} / {\bf 0.9161} \\
    \hline
    \end{tabular}}
    \end{subtable}
\end{table*}

\noindent
{\bf Learning Rate Decay.}
We observed out that cosine learning rate decay (cosine-decay)~\cite{loshchilov2016sgdr} did not perform well on SR tasks.
It was because the underfitting (not overfitting) is a crucial issue to SR~\cite{lin2022revisiting}.
Interestingly, it differs from the high-level vision tasks such as classification, object detection, and semantic segmentation.
We hypothesize that the cosine-decay reduces the \textit{lr} faster than the half-decay (keeping a constant for long phases) and leads to the underfitting.
However, we also observed that a decay point that was too early or too late ended up with the wrong converging point and decreased performances.

\noindent
{\bf Regularization.}
We found that the SR tasks were hampered by the regularization strategies, such as weight decay~\cite{loshchilov2017decoupled}, gradient clipping~\cite{pascanu2012understanding,pascanu2013difficulty}, and layer-wise \textit{lr} decay~\cite{clark2020electra}.
We compared a non-regularization strategy with the methods with 0.05 weight decay or 5.0 gradient clipping.
However, these strategies dropped the performance.
Similarly, while layer-wise \textit{lr} decay improved the performance of high-level vision tasks when fine-tuning Transformer models~\cite{bao2021beit,he2022masked}, our models could not learn the representations well with that regularization.
This is also because SR tasks suffer from underfitting unlike recognition tasks.

\noindent
{\bf Dropout.}
Considering that the crucial issue of SR tasks is underfitting, the dropout also had a negative effect on our work.
Although we utilized the different dropout~\cite{hinton2012improving} and drop-path~\cite{larsson2016fractalnet} rates, they were not good for NGswin.
However, a recent work~\cite{kong2022reflash} has demonstrated the appropriate dropout strategy could improve SR performance.

\vspace{-5.3pt}
\section{Other Ablation Studies}
\label{other_ablation}

\noindent
{\bf Swin Transformer Version.}
\cref{tab_swin_version} demonstrates the superiority of SwinV2 over SwinV1 for SR tasks of NGswin.
As mentioned in~\cite{liu2022swin}, it is because dot-product self-attention of SwinV1 tends to make a few pixel pairs dominate the trained attention maps.
But SR tasks need not some certain but neighbor pixels to recover degraded regions and reconstruct HR images.
Since normalization is inherent in the cosine similarity of SwinV2, scaled-cosine self-attention can hinder some certain pixels from hugely affecting reconstruction tasks.

\noindent
{\bf Padding.}
We investigated the impacts of \textit{seq-refl-win-pad} in~\cref{tab_seq_refl}.
The trivial zero-padding (zero-pad) often conveys meaningless values to the feature maps, which causes several degraded regions to interact with empty pixels.
It was a severely adverse method for uni-Gram embedding that had significantly low resolution (8$\times$8, 4$\times$4, 2$\times$2).
However, \textit{seq-refl-win-pad} could give non-zero neighbors \textemdash some neighbors from other directions, to be precise\textemdash\space to the uni-Grams that were insufficient to neighbors.
As a result, the networks could learn more meaningful representations, compared to zero-pad.
Even the zero-pad approach was worse than the model without the N-Gram context.
Mult-Adds operations and the number of parameters of the models were unchanged.

\noindent
{\bf Warm-Start.}
It is obvious that the warm-start strategy requires much shorter training time than training from scratch (scratch).
As explained in~\cref{supp_experiment}, the scratch and warm-start scheme lasted for 500 and 300 epochs.
Meanwhile, the 50 warm-start epochs that only updated the reconstruction module spent fractional times.
Therefore, the warm-start scheme was about twice faster than scratch.
However, we wondered if this scheme would also be better when it comes to SR performances as in RCAN-it~\cite{lin2022revisiting}.
\cref{tab_warmstart} shows that SwinIR-NG trained by warm-start scheme recorded higher scores than scratch.
Therefore, the warm-start training strategy is superior to scratch in terms of both time resource and performances for SR tasks.

\noindent
{\bf Other Benchmarks.}
Due to the page limit in the main content, we did not report the results of ablation studies on some SR benchmark test datasets.
In~\cref{tab_other_bench}, we posted those results including benchmarks already shown in~\cref{ablation}.
The results consistently showed the positive effect of each proposed (or employed) approach.

\begin{figure*}[t]
    \centering
    \includegraphics[width=0.95\linewidth]{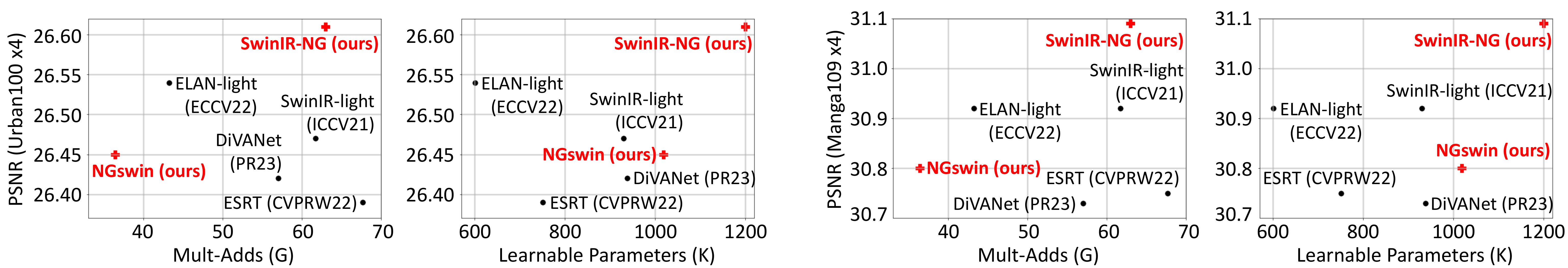}
    \caption{Trade-off between performance and efficiency. The parameter sizes of NGswin and SwinIR-NG are 4.04MB and 4.74MB.}
    \label{fig_tradeoff}
\end{figure*}

\section{More Visual Comparisons}
\label{vis_comp}
\begin{figure*}[t]
    \centering
    \begin{tabular}{c}
        \includegraphics[width=0.945\linewidth]{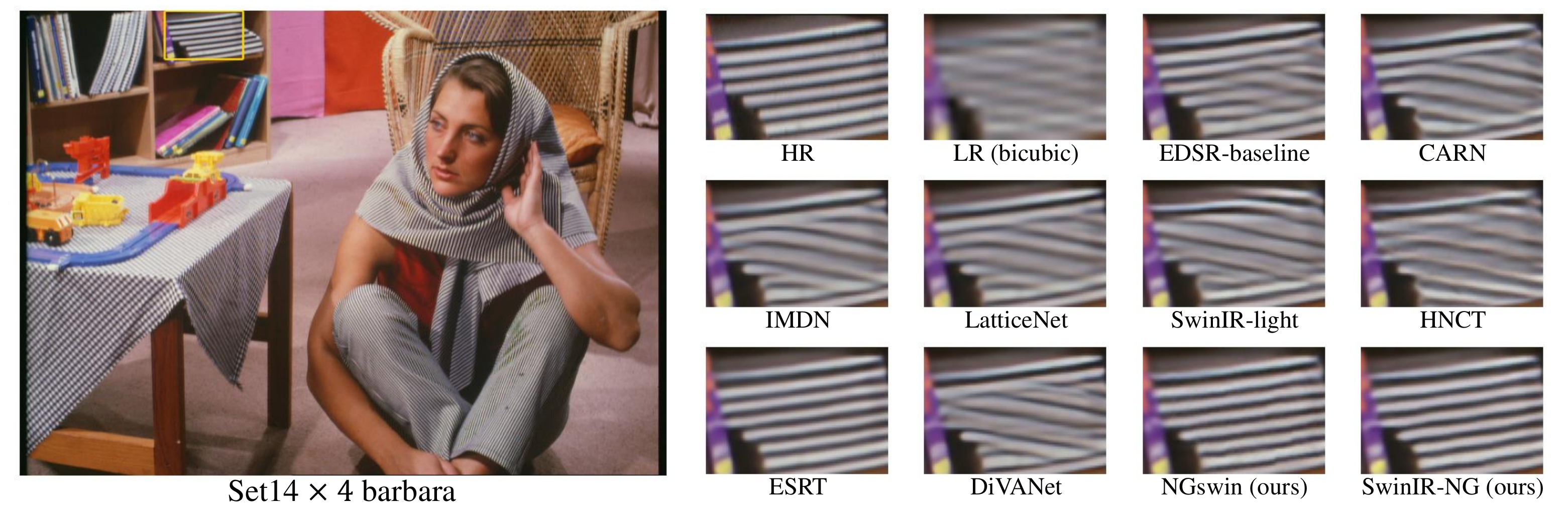} \\
        \includegraphics[width=0.945\linewidth]{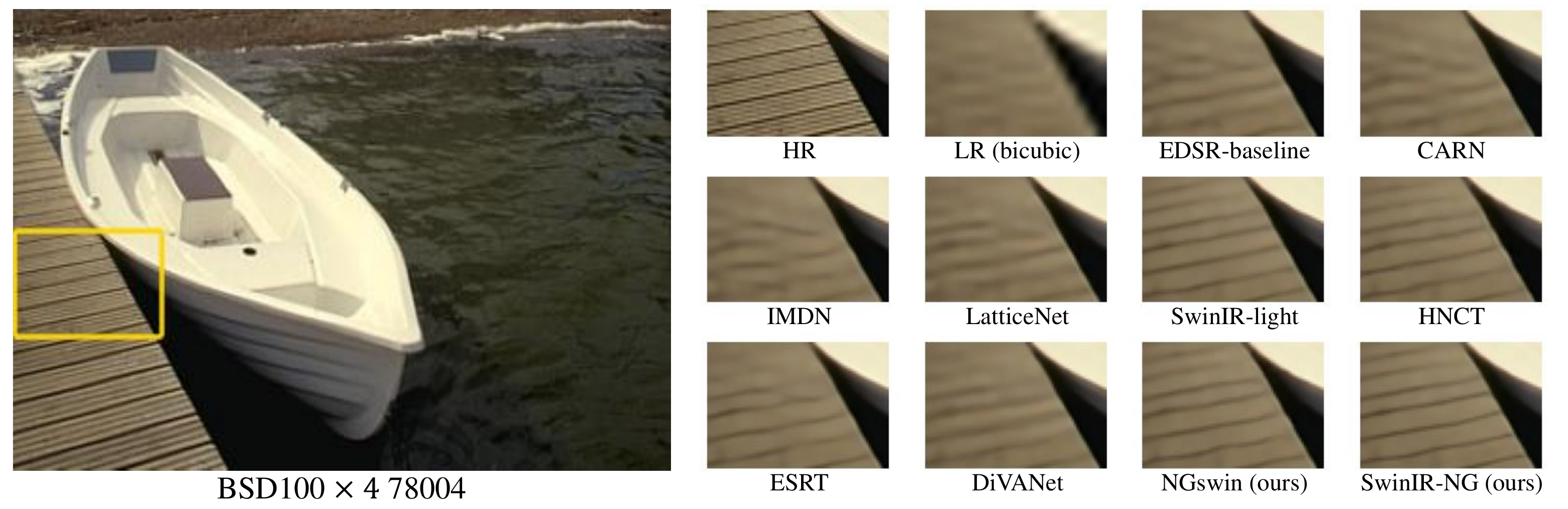} \\
        \includegraphics[width=0.945\linewidth]{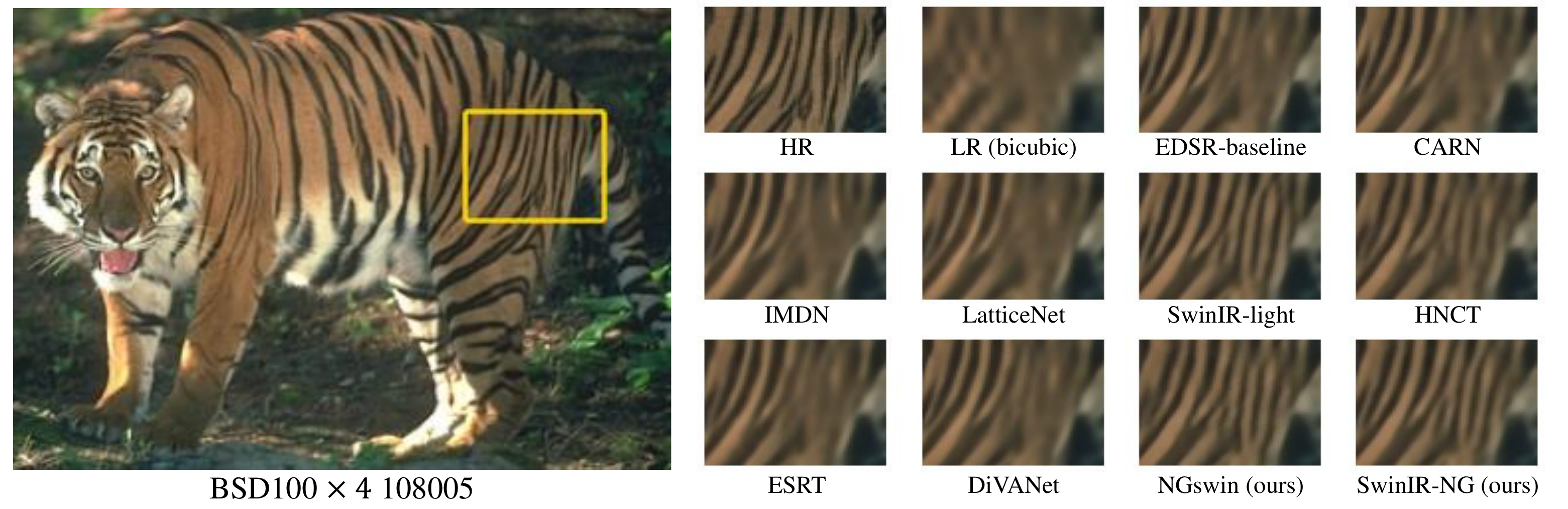} \\
        \includegraphics[width=0.945\linewidth]{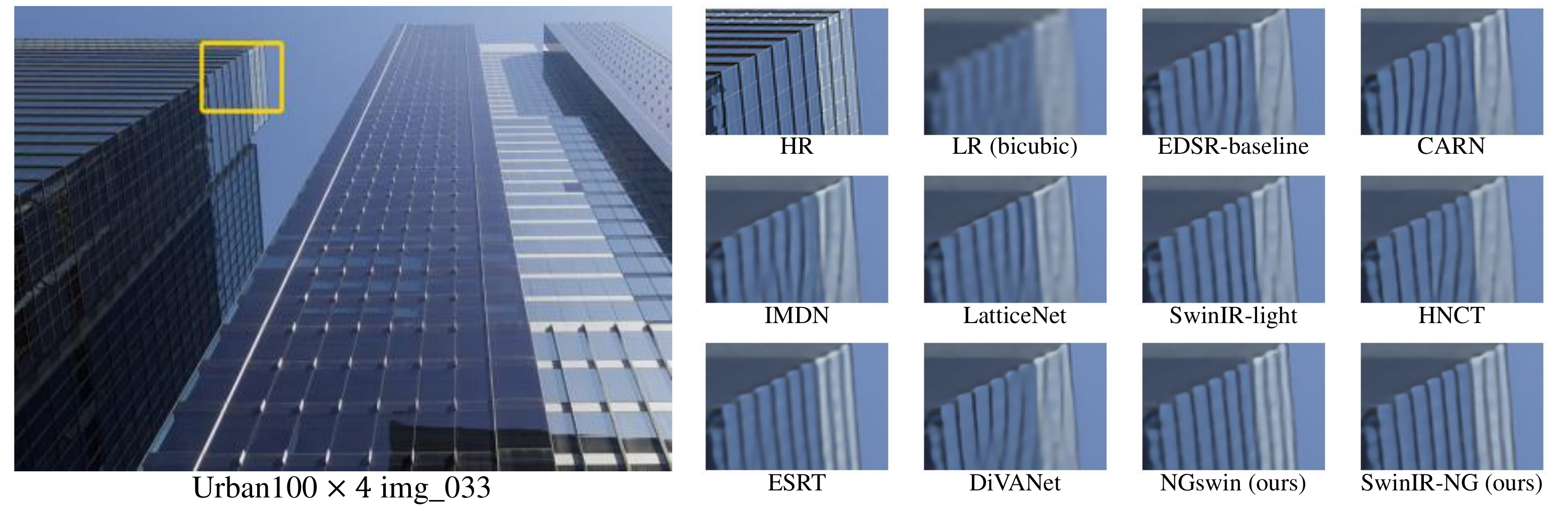} \\
    \end{tabular}

    \caption{Visual comparisons ($\times 4$). \enquote{LR (bicubic)} indicates the low-resolution input images from bicubic interpolation.}
    \label{fig_comp_others1}
\end{figure*}

\begin{figure*}[t]
    \centering
    \begin{tabular}{c}
        \includegraphics[width=0.945\linewidth]{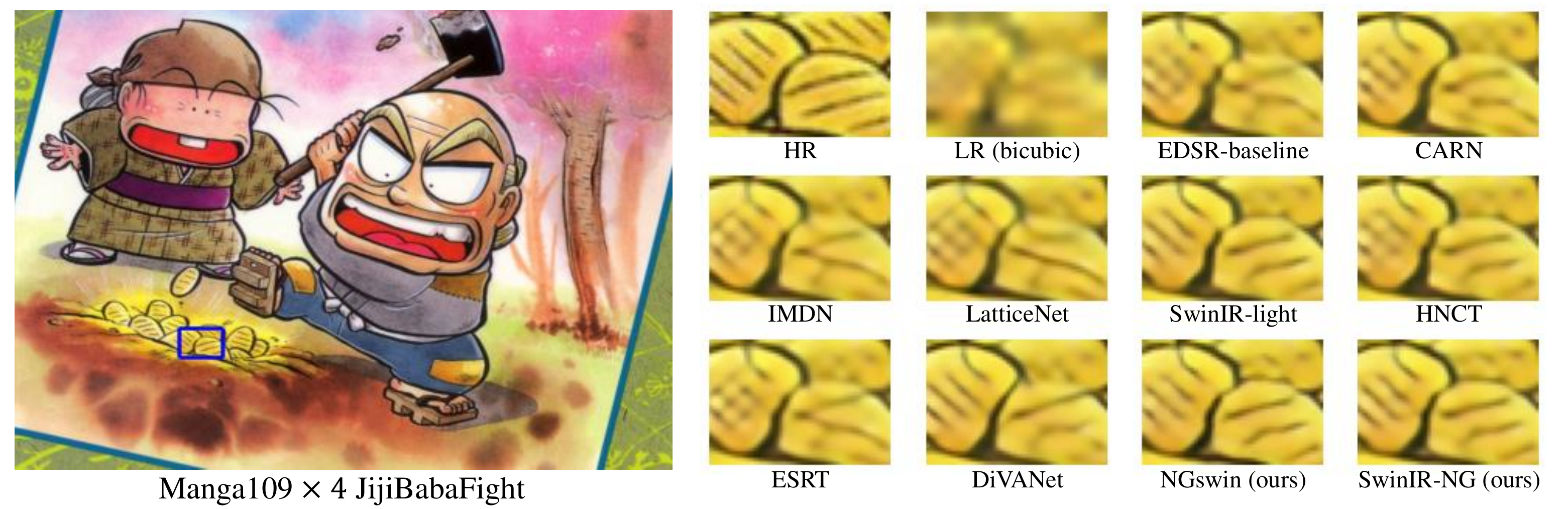} \\
        \includegraphics[width=0.945\linewidth]{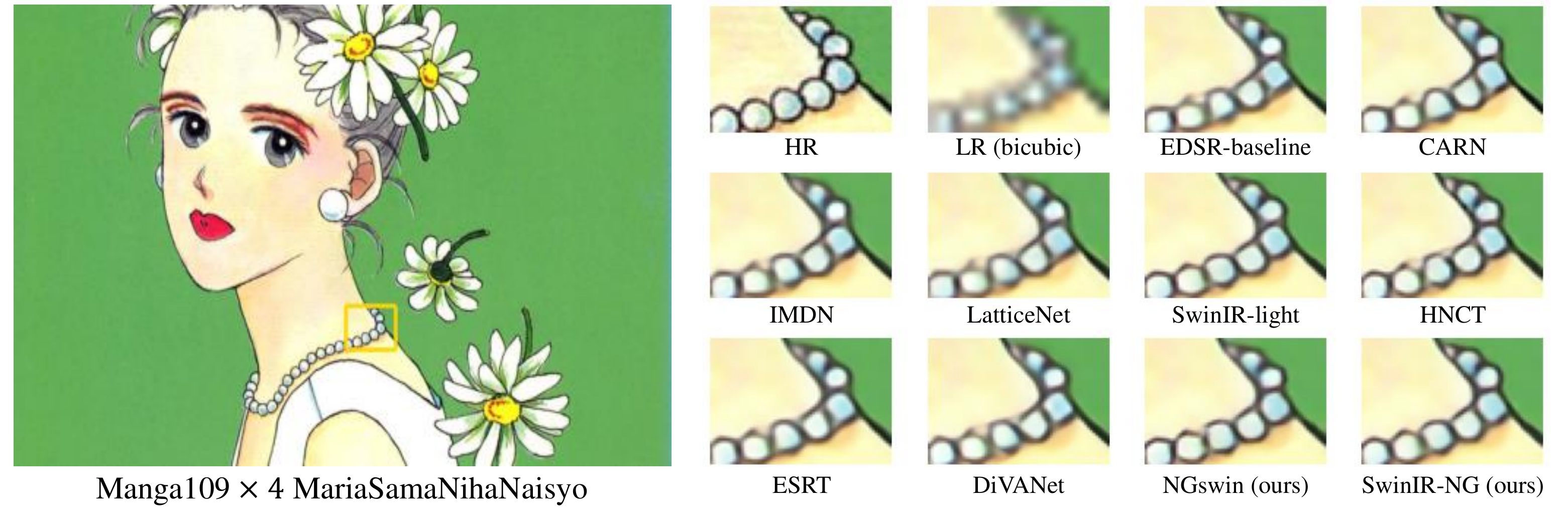} \\
        \includegraphics[width=0.945\linewidth]{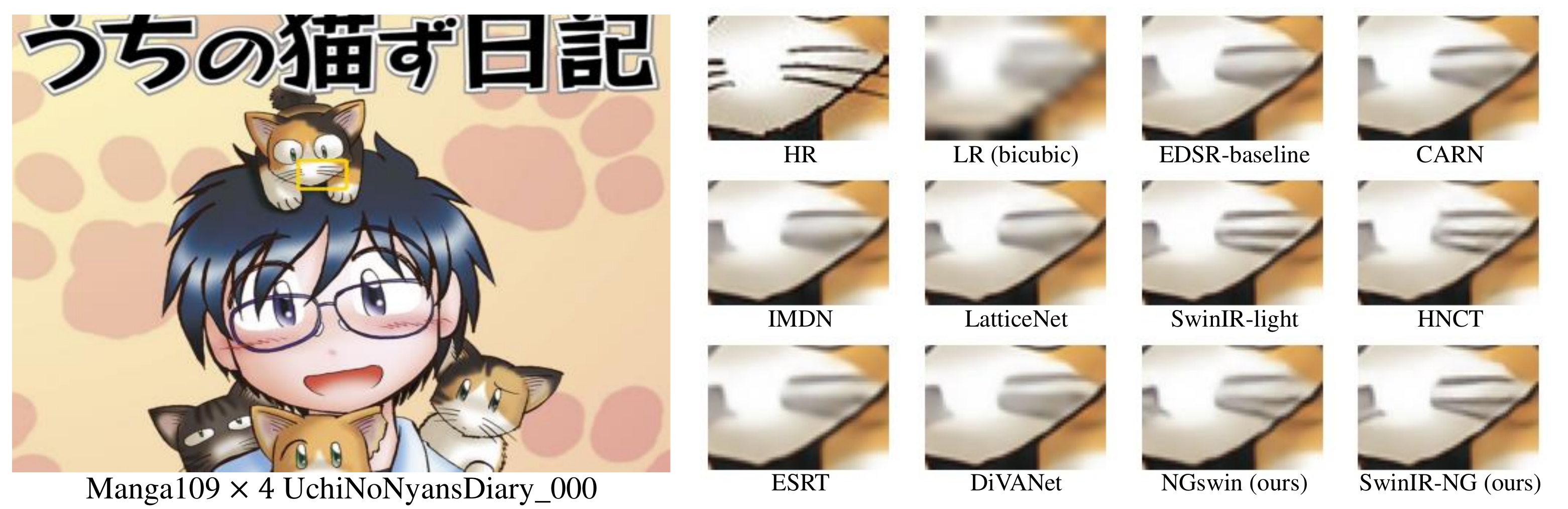} \\
        \includegraphics[width=0.945\linewidth]{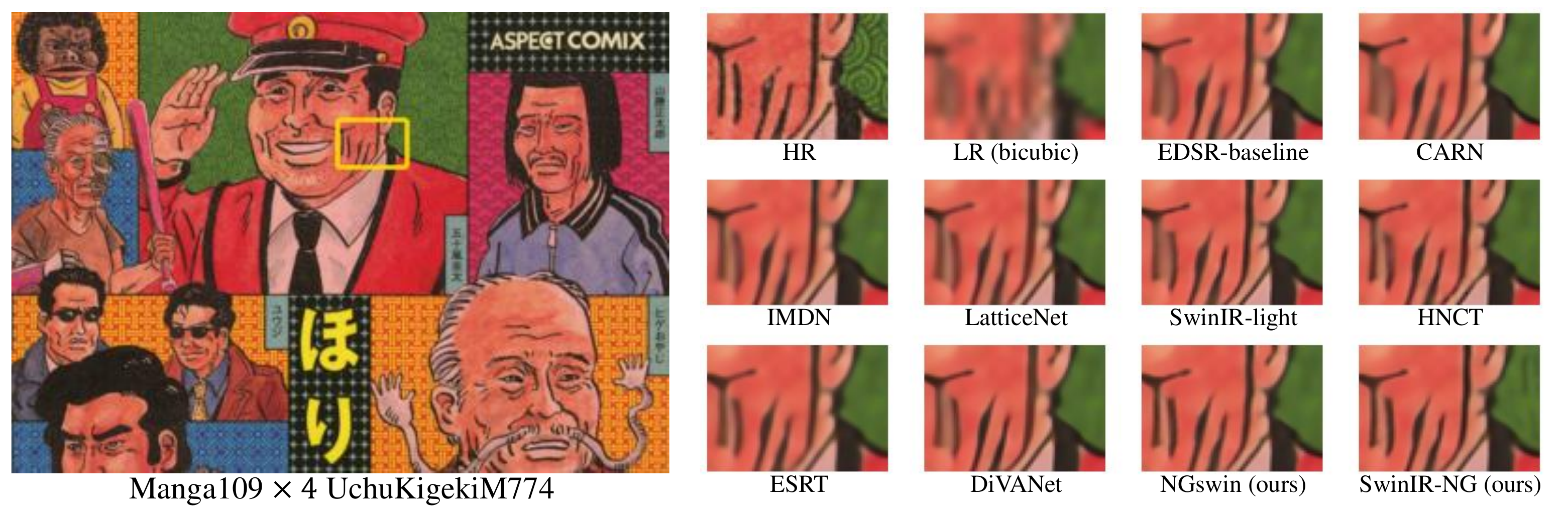} \\
    \end{tabular}

    \caption{Visual comparisons ($\times 4$). \enquote{LR (bicubic)} indicates the low-resolution input images from bicubic interpolation.}
    \label{fig_comp_others2}
\end{figure*}

\begin{figure*}[t]
    \centering
    \begin{tabular}{c}
        \includegraphics[width=0.945\linewidth]{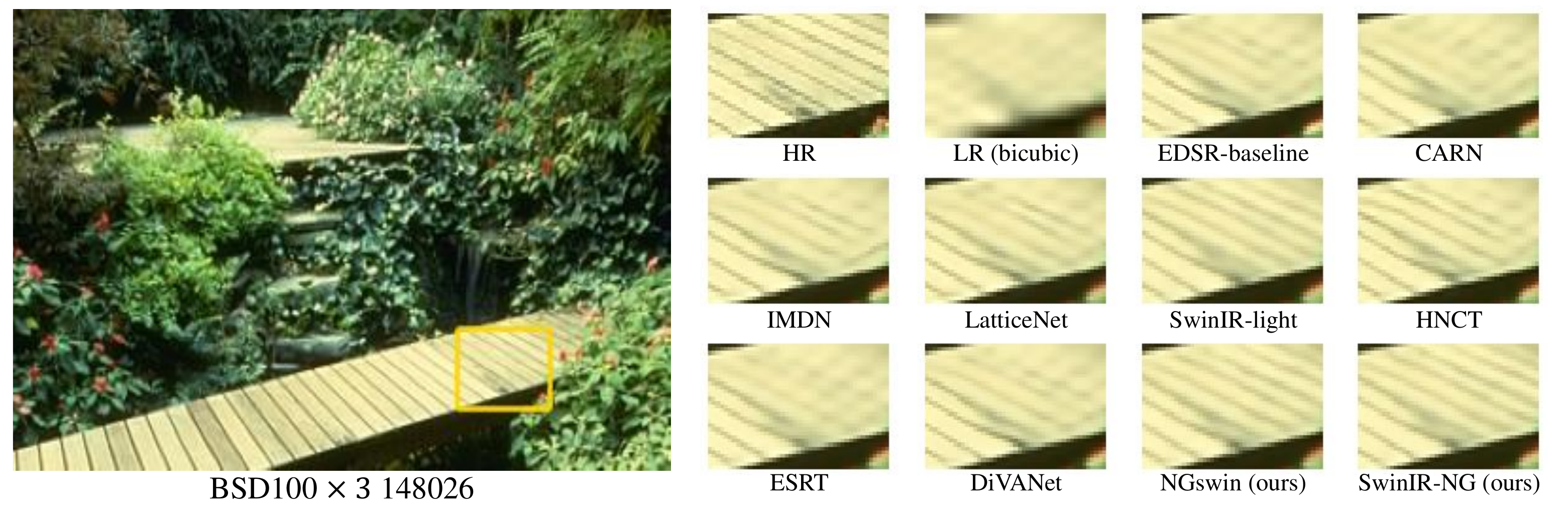} \\
        \includegraphics[width=0.945\linewidth]{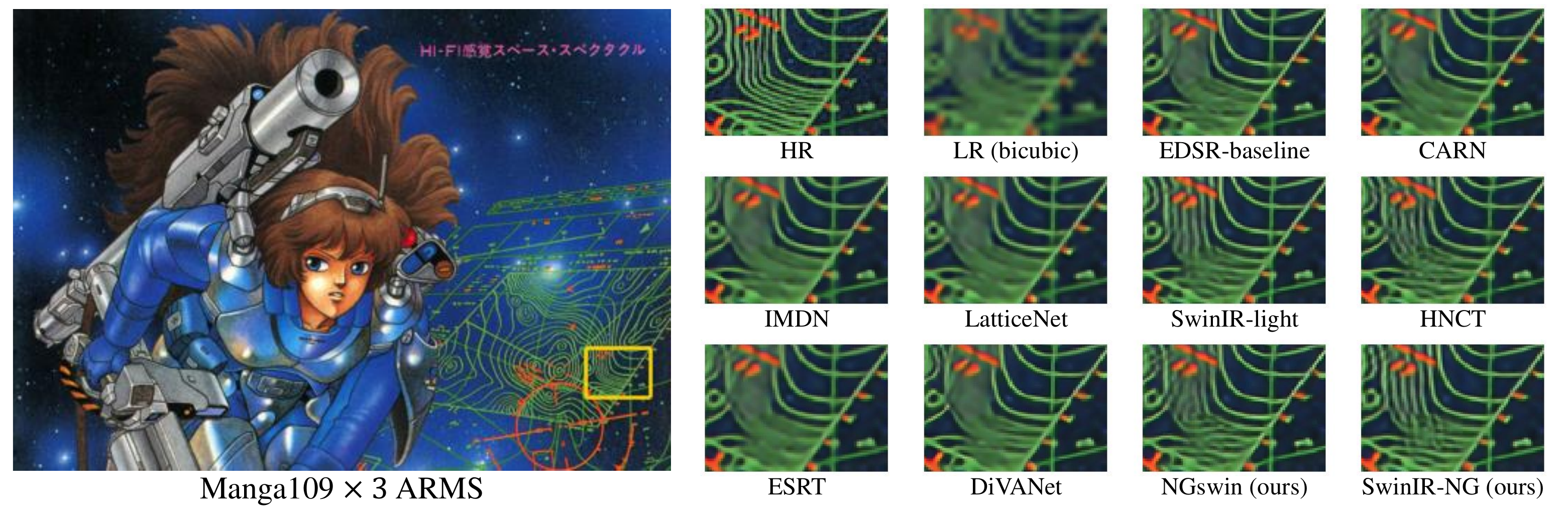} \\
        \includegraphics[width=0.945\linewidth]{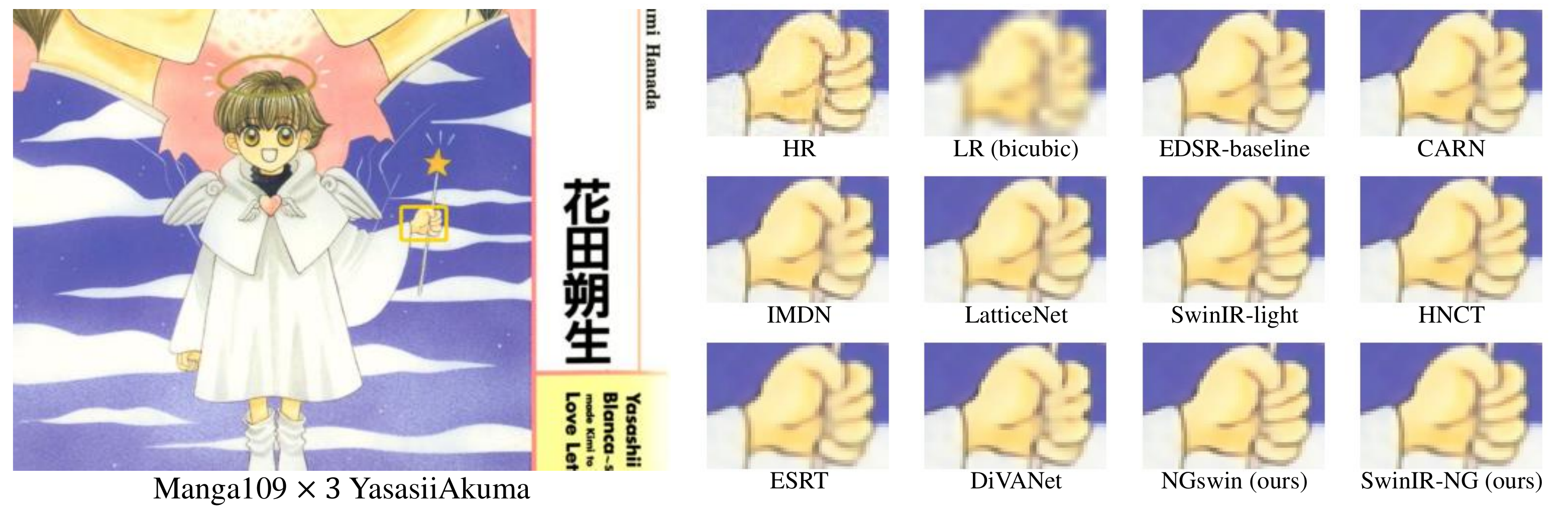} \\
        \includegraphics[width=0.945\linewidth]{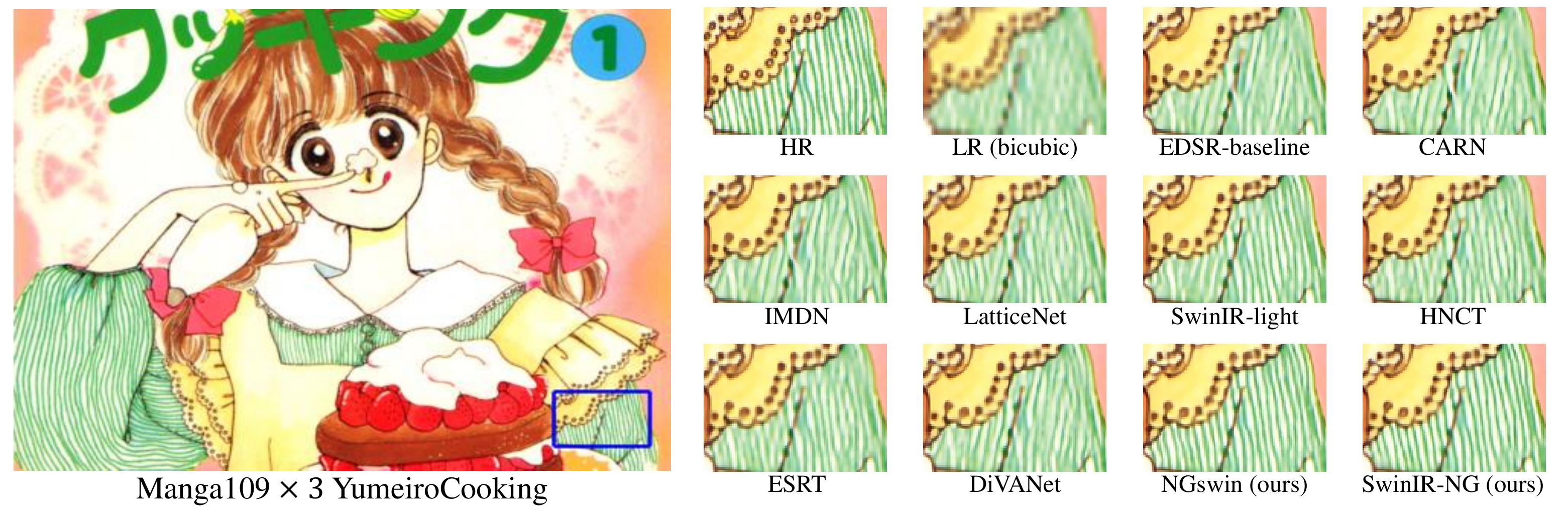} \\
    \end{tabular}

    \caption{Visual comparisons ($\times 3$). \enquote{LR (bicubic)} indicates the low-resolution input images from bicubic interpolation.}
    \label{fig_comp_others3}
\end{figure*}

\begin{figure*}[t]
    \centering
    \begin{tabular}{c}
        \includegraphics[width=0.945\linewidth]{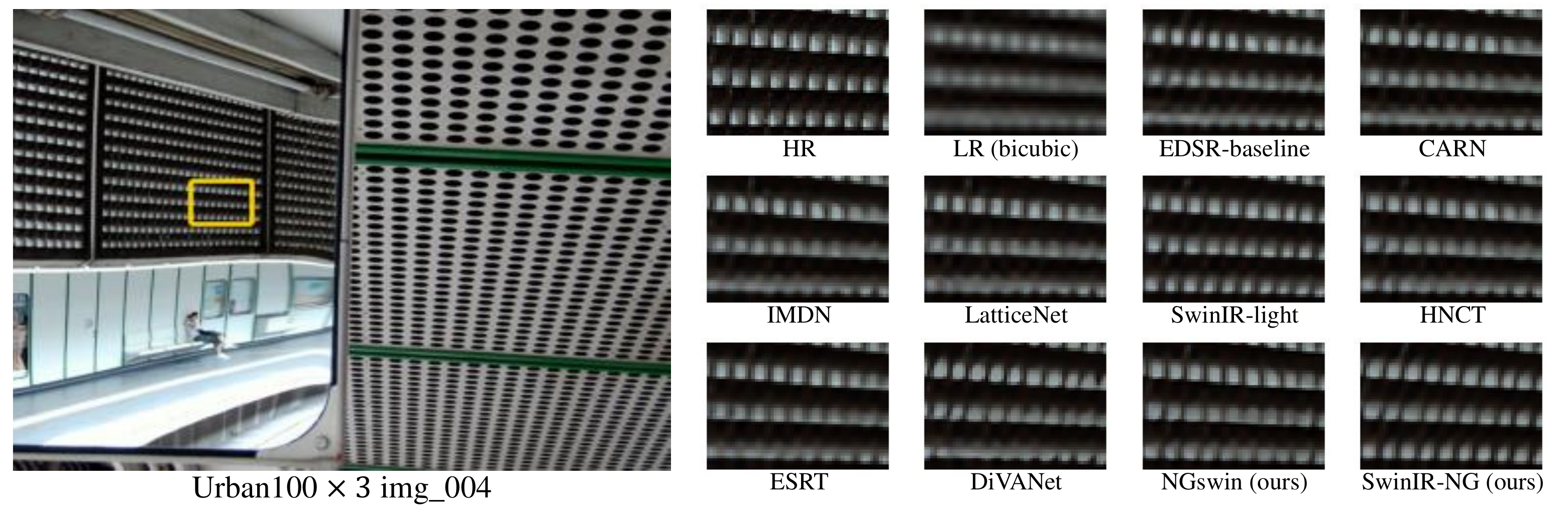} \\
        \includegraphics[width=0.945\linewidth]{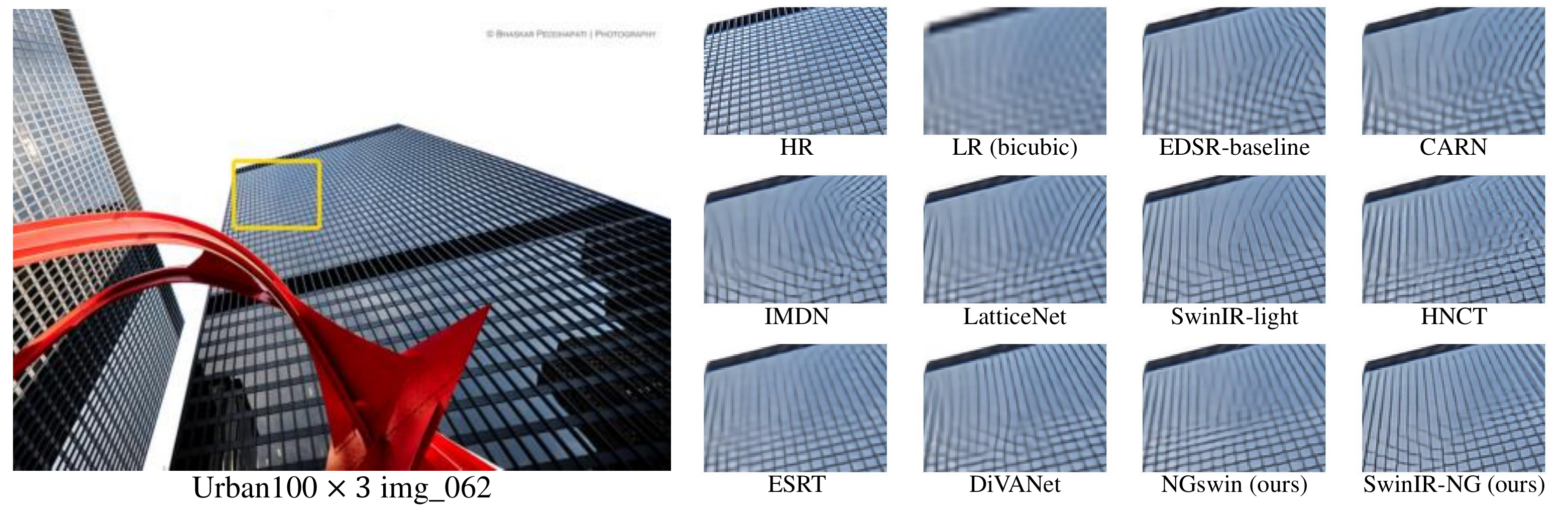} \\
        \includegraphics[width=0.945\linewidth]{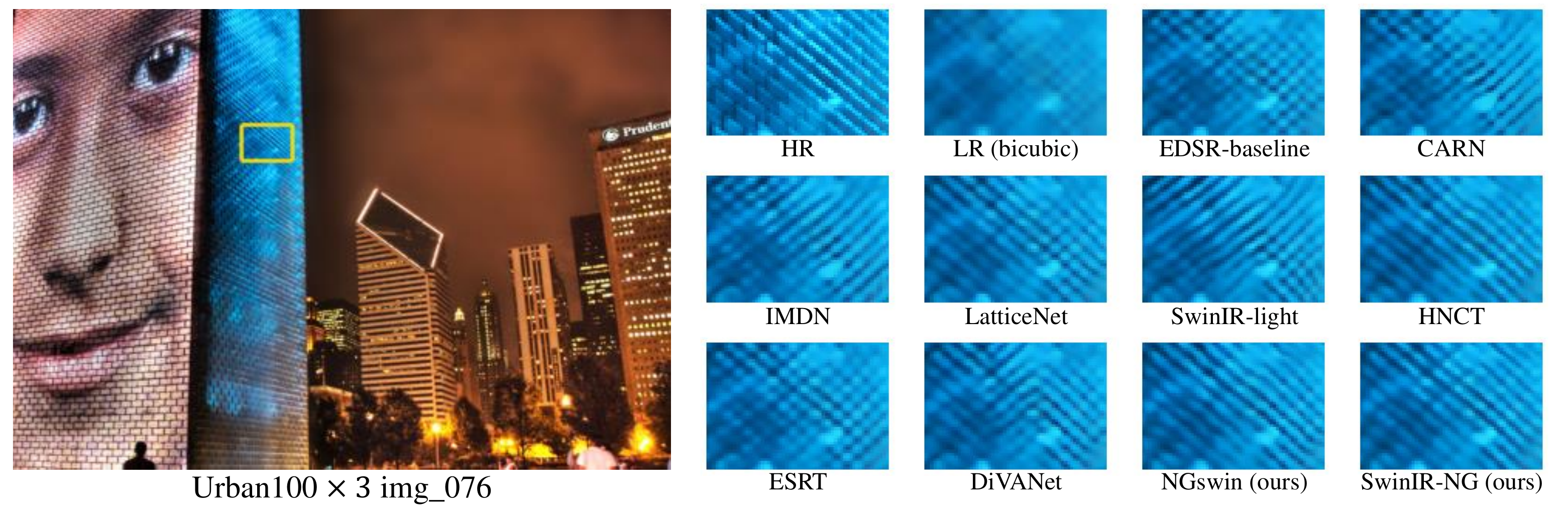} \\
        \includegraphics[width=0.945\linewidth]{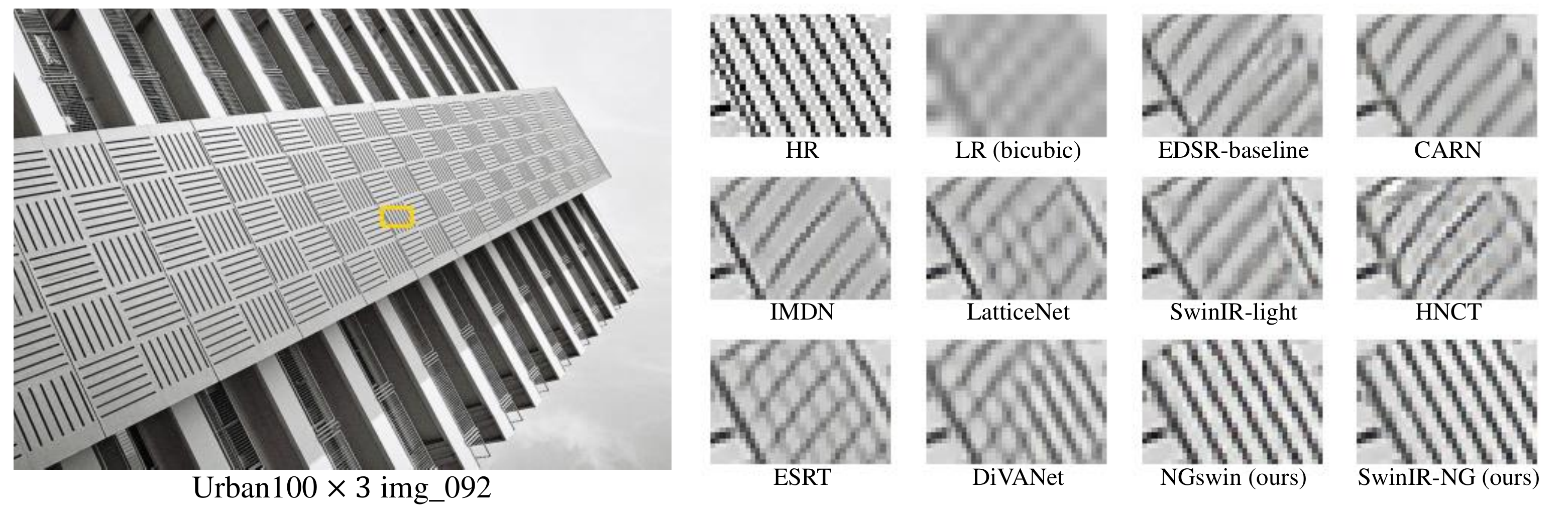} \\
    \end{tabular}

    \caption{Visual comparisons ($\times 3$). \enquote{LR (bicubic)} indicates the low-resolution input images from bicubic interpolation.}
    \label{fig_comp_others4}
\end{figure*}

\begin{figure*}[t]
    \centering
    \begin{tabular}{c}
        \includegraphics[width=\linewidth]{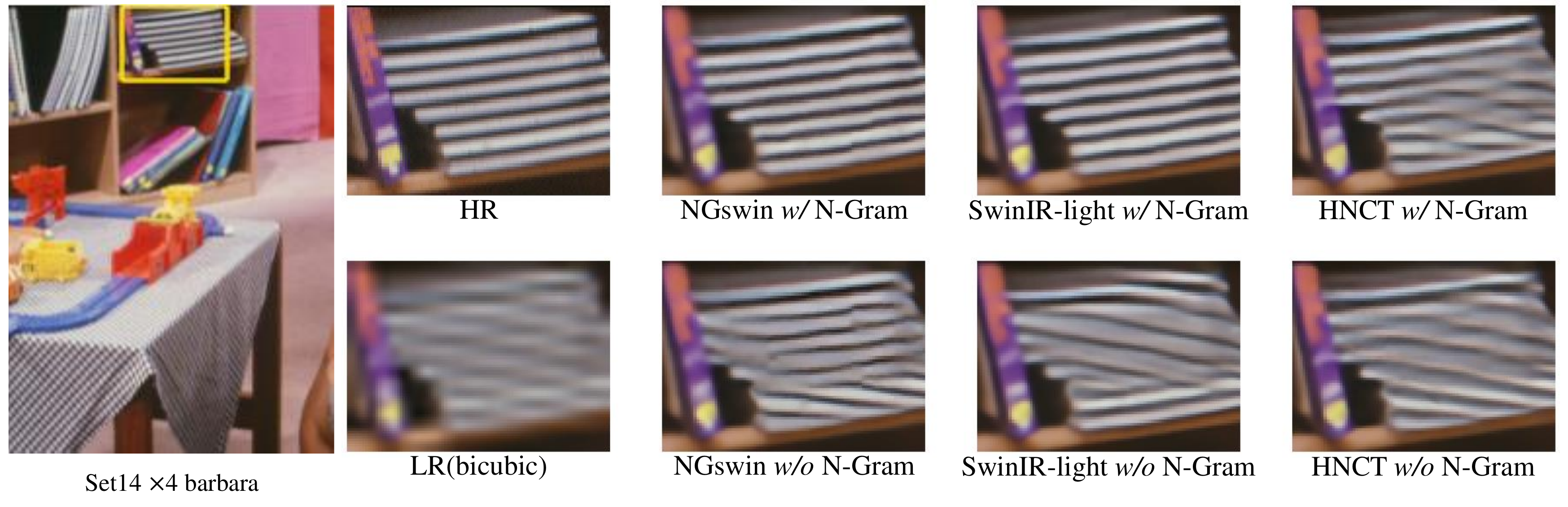} \\
        \includegraphics[width=\linewidth]{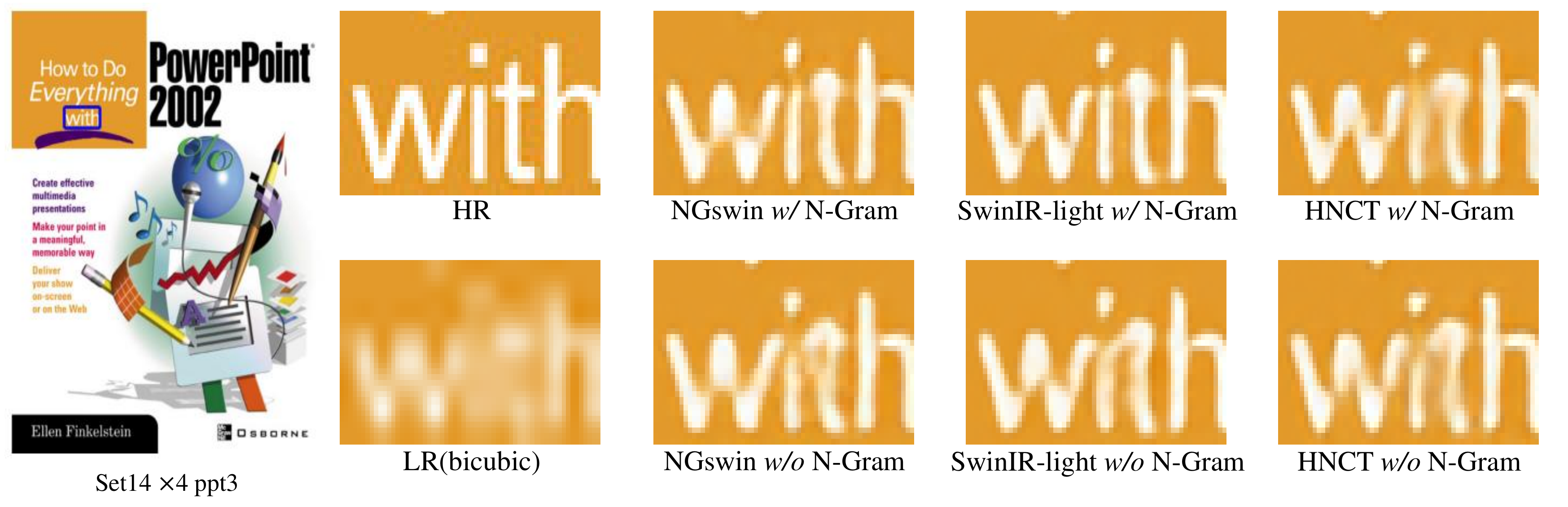} \\
        \includegraphics[width=\linewidth]{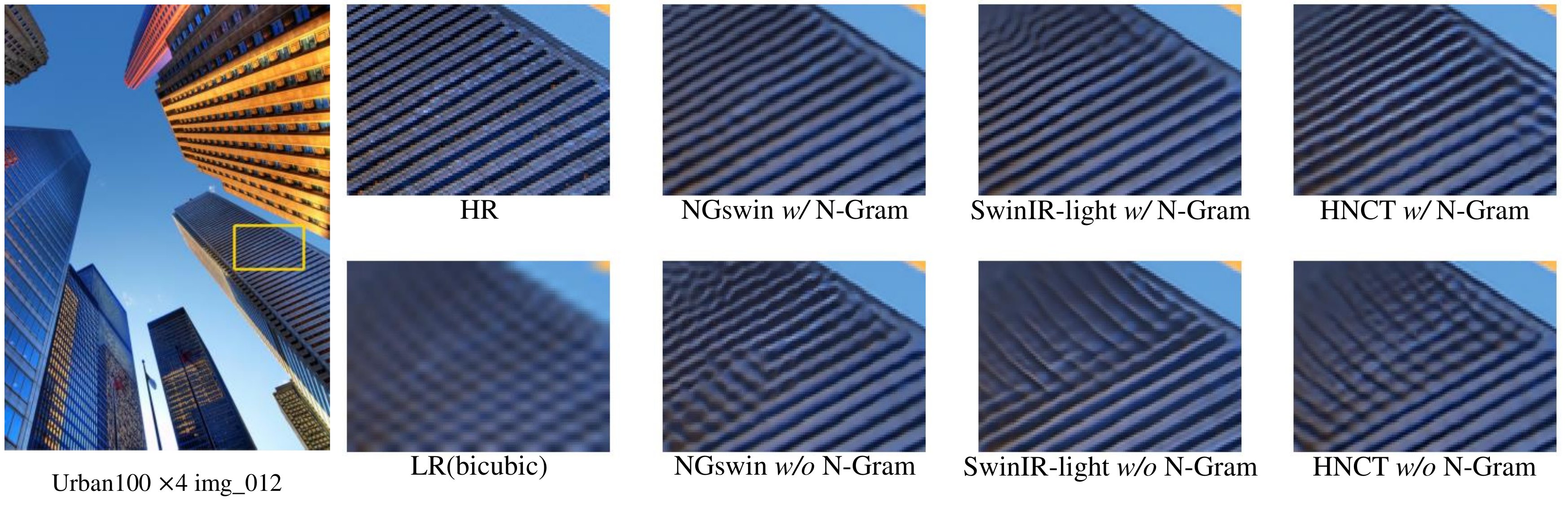} \\
    \end{tabular}

    \caption{Visual comparisons of the models with \vs without the N-Gram context ($\times 4$). \enquote{LR (bicubic)} indicates the low-resolution input images from bicubic interpolation.}
    \label{fig_comp_ngram1}
\end{figure*}

\begin{figure*}[t]
    \centering
    \begin{tabular}{c}
        \includegraphics[width=\linewidth]{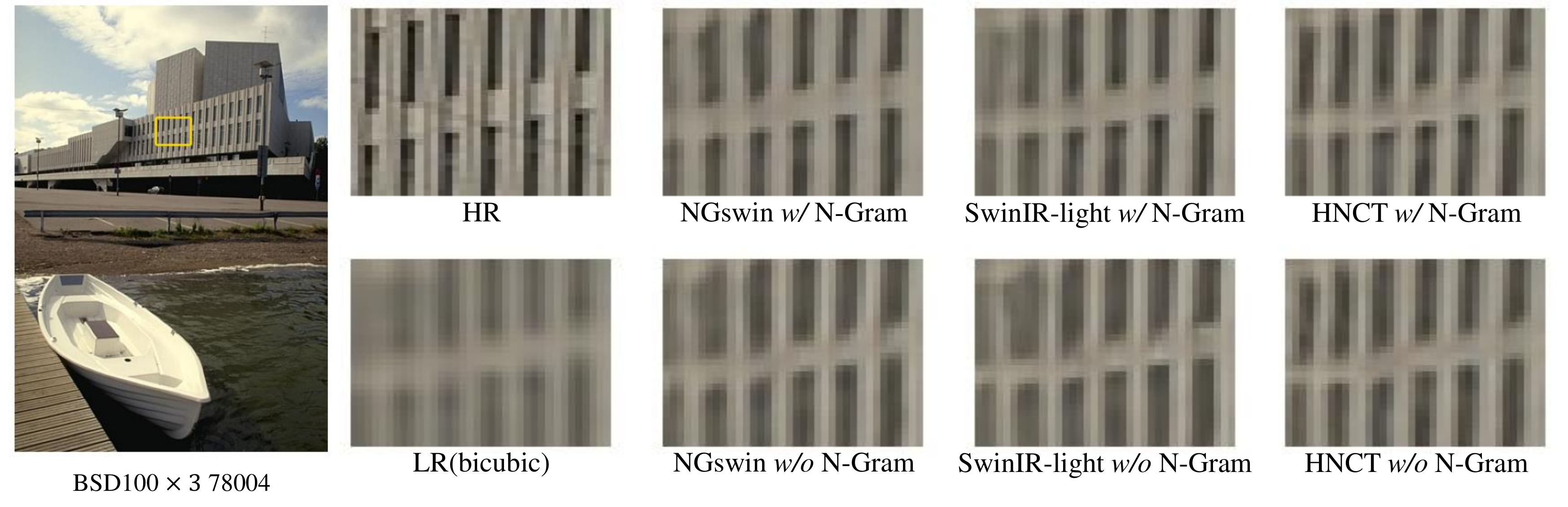} \\
        \includegraphics[width=\linewidth]{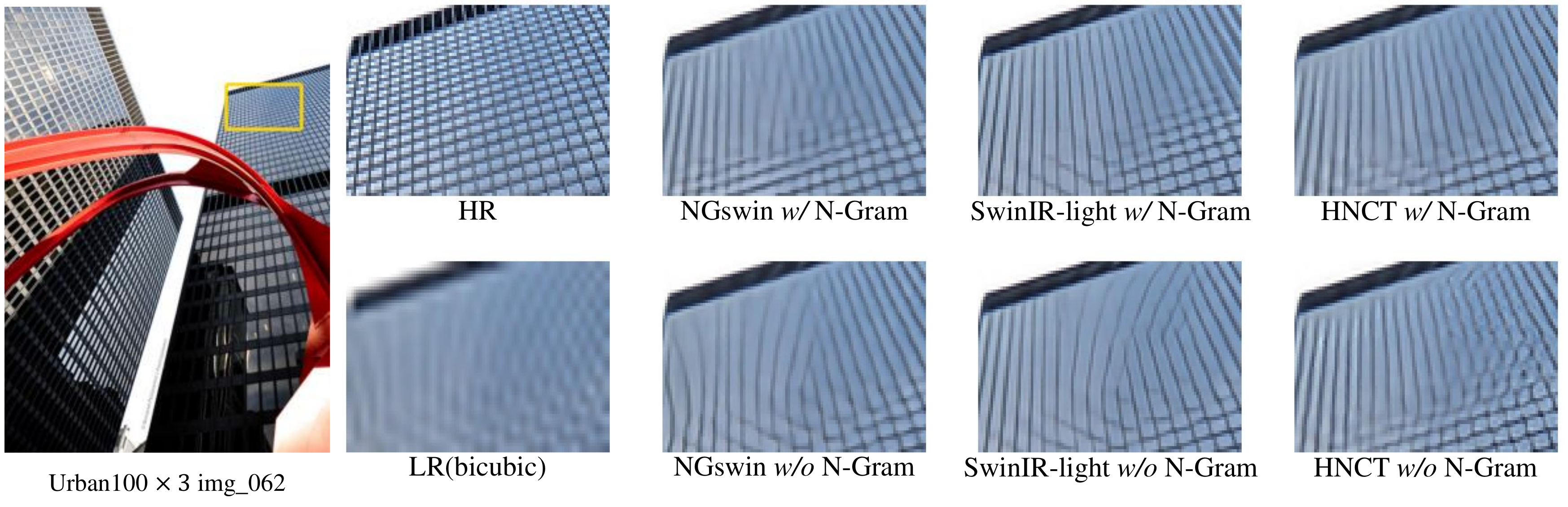} \\
        \includegraphics[width=\linewidth]{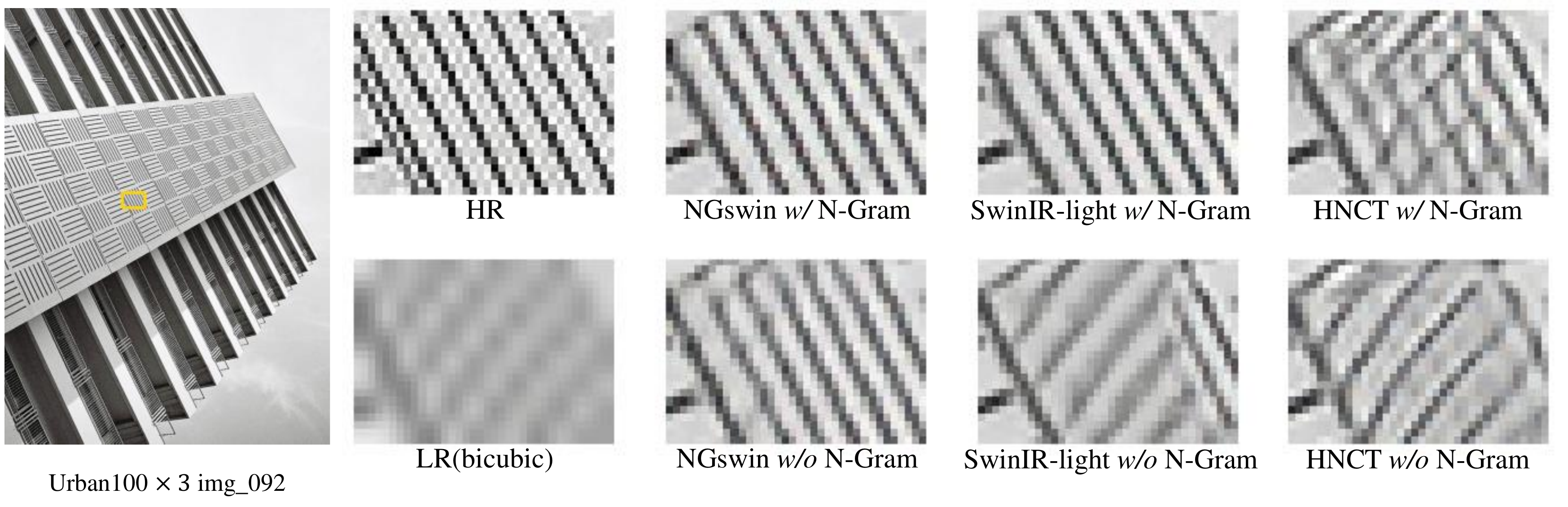} \\
    \end{tabular}

    \caption{Visual comparisons of the models with \vs without the N-Gram context ($\times 3$). \enquote{LR (bicubic)} indicates the low-resolution input images from bicubic interpolation.}
    \label{fig_comp_ngram2}
\end{figure*}

We supply more visual comparisons with other models in~\cref{fig_comp_others1,,fig_comp_others2,,fig_comp_others3,,fig_comp_others4}.
Also, more comparisons of the models with \vs without the N-Gram context are visualized in~\cref{fig_comp_ngram1,,fig_comp_ngram2}.

\section{Discussions and Limitations}
\label{discussion}
In this section, we discuss the characteristics of our novel methods including the novelty and limitations.
Moreover, we reflect on other tasks that are not addressed in this paper but can be developed by our N-Gram context.
Finally, it is considered how this work can be extended to further improve our methods and cover broader tasks.

\noindent
{\bf Methods.}
\cref{fig_tradeoff} illustrates the trade-off between performance and efficiency of our models and the best lightweight SR methods.
NGswin has the fewest Mult-Adds operations and SwinIR-NG presents the best performance.

\noindent
{\bf [NGswin]}
As shown in~\cref{SCDP} and~\cref{tab_ablation_scdp}, our proposed SCDP bottleneck compensated the performance loss of the hierarchical encoder.
However, one may doubt whether NGswin can be improved by abandoning the hierarchical architecture, as in ELAN-light and SwinIR-light.
Of course, the non-hierarchical structure would significantly lead to performance gains.
However, the following approximation shows the influence of training input resolution $hw$ of a single NSTB on Mult-Adds of $\times r$ task:
\vspace{-7pt}
\begin{equation*}
\vspace{-7pt}
    \mathrm{MultAdds}(\mathrm{NSTB}) \approx (10 \times hw / 2^{12} / (\frac{r}{2})^2)\mathrm{G}.
\end{equation*}
\noindent
Therefore, if there were no \textit{patch-merging} layers, the eight NSTBs in the 2nd and 3rd encoder stages would have increased the operations by about 17G for the $\times 4$ task.

\noindent
{\bf [SwinIR-NG]}
Compared with SwinIR-light (\textit{w/o} N-Gram), SwinIR-NG (\textit{w/} N-Gram) needs a small number of extra operations to establish state-of-the-art lightweight SR.
The parameters were also kept as a tolerable size (4.74MB) for semiconductor system, as stated in~\cref{sec:intro}.
However, it is a limitation that SwinIR-NG requires more parameters and operations than ELAN-light~\cite{zhang2022efficient}.
These results are due to our intention of focusing on improved performance.

\noindent
{\bf [N-Gram Context]}
Our N-Gram context differs from recent attention mechanisms proposed for the efficient self-attention (SA).
First, while our \textit{sliding-WSA} produces the average correlations in the spatial space, channel attention (CA)~\cite{zhang2018image} computed SA in the channel space, which was employed in Restormer~\cite{Zamir2021Restormer} and NAFNet~\cite{chen2022simple}.
Second, group-wise multi-scale self-attention (GMSA) proposed by ELAN-light~\cite{zhang2022efficient} divided the feature maps into $K$ groups to avoid intensive operations.
In contrast, the dimensionality reduction of our channel-reducing group convolution (uni-Gram embedding) decreases the time complexity of \textit{sliding-WSA}.
Third, cross-shaped window (CSWin)~\cite{dong2022cswin} enlarged the receptive field of SA by splitting multi-heads horizontally and vertically, then computed SA in each multi-head group.
Recently, Cross Aggregation Transformer (CAT)~\cite{zheng2022cross} adopted the similar SA strategy for multiple image restoration tasks with a large model size.
Whereas, we calculate SA within $N^2$ uni-Gram embeddings.
The weight sharing for the bi-directional N-Gram features also broadens the receptive field.
As a result, the receptive field of \textit{sliding-WSA} is expanded $2N^2$ times.
Lastly, the parameters of the N-Gram context can be further reduced by properly adopting and varying other methods.

\noindent
{\bf [SwinV2]}
Most recently, Swin2SR~\cite{conde2022swin2sr} adapted SwinV2 like NGswin.
Unlike our model, Swin2SR employed a continuous relative position bias~\cite{liu2022swin}.
It is compelling that Swin2SR demonstrated a potential that SwinIR-NG could be improved by SwinV2, as in our \cref{tab_swin_version}.
However, Swin2SR was trained with 3,450 images from a merged dataset of DIV2K and Flickr2K~\cite{timofte2017ntire}, which were even more than our 800 training images.
Likewise, the Swin2SR paper only reported the results of the $\times2$ lightweight SR task.
Therefore, despite its remarkable performance, we excluded Swin2SR from~\cref{tab_sota_comp} for fair comparison.

\begin{figure*}[t]
    \centering
    \begin{tabular}{c}
        \includegraphics[width=0.81\linewidth]{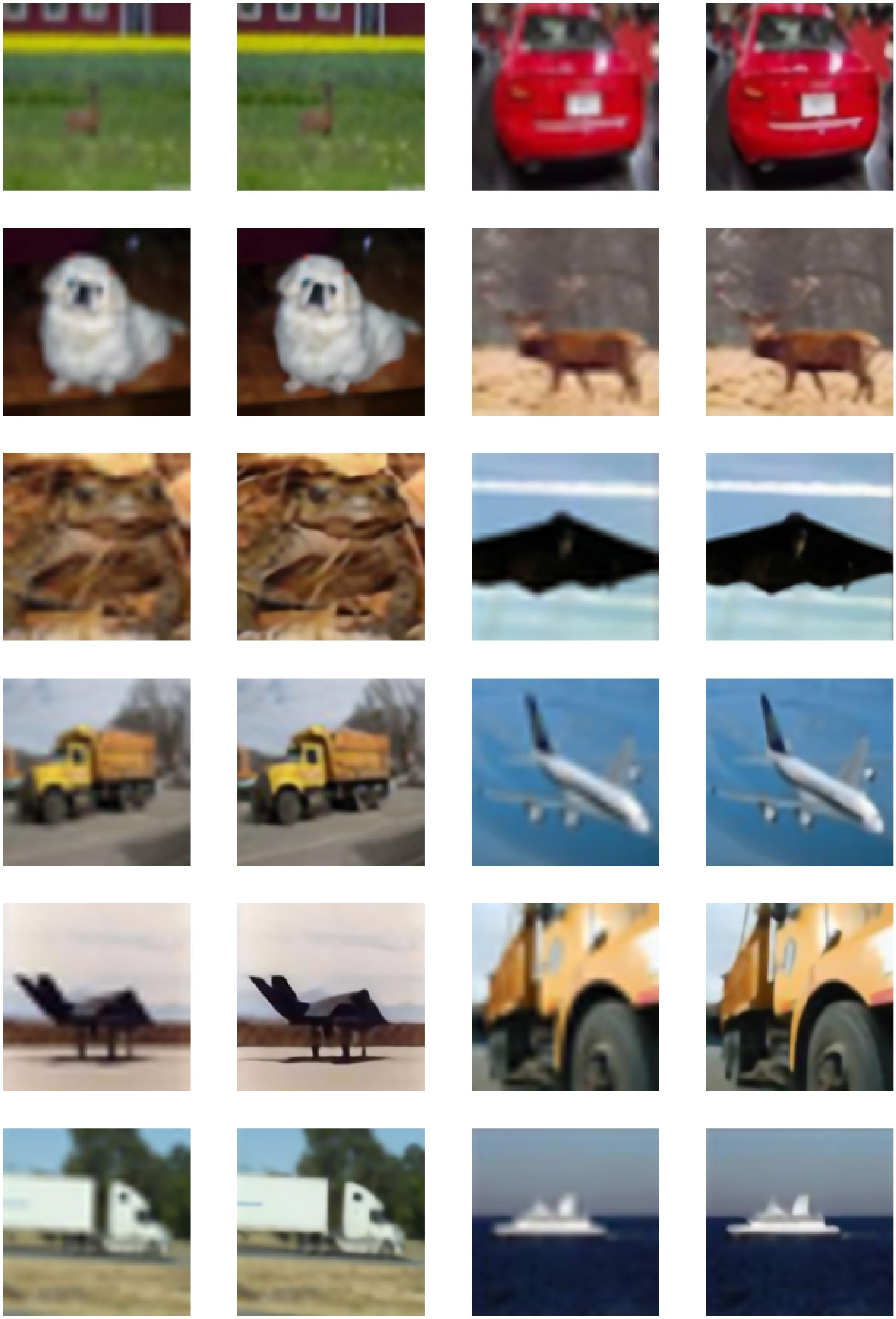} \\
    \end{tabular}

    \caption{Visual results with SwinIR-NG on CIFAR10~\cite{krizhevsky2009learning} ($\times 4$). Our technique may boost classification tasks with the sharper edges of super-resolution results. The 1st and 3rd columns are LR (bicubic) images. The 2nd and 4th columns are from SwinIR-NG. As this figure is secondary provision, we do not compare ours with other models.}
    \label{fig_cifar1}
\end{figure*}

\begin{figure*}[t]
    \centering
    \begin{tabular}{c}
        \includegraphics[width=0.81\linewidth]{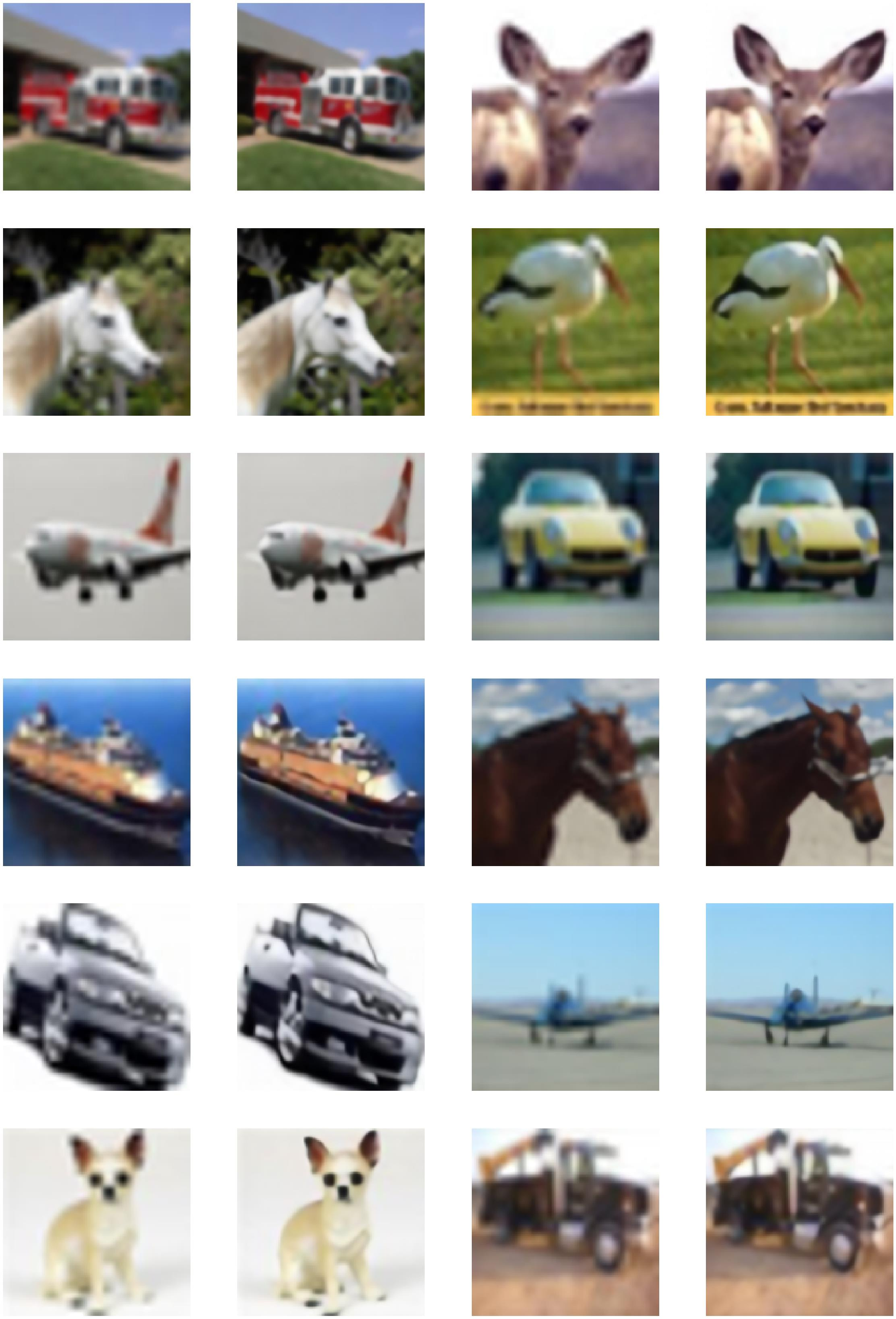} \\
    \end{tabular}

    \caption{Visual results with SwinIR-NG on CIFAR10~\cite{krizhevsky2009learning} ($\times 4$). The explanations are in~\cref{fig_cifar1}.}
    \label{fig_cifar2}
\end{figure*}

\begin{figure*}[t]
    \centering
    \begin{tabular}{c}
        \includegraphics[width=0.9\linewidth]{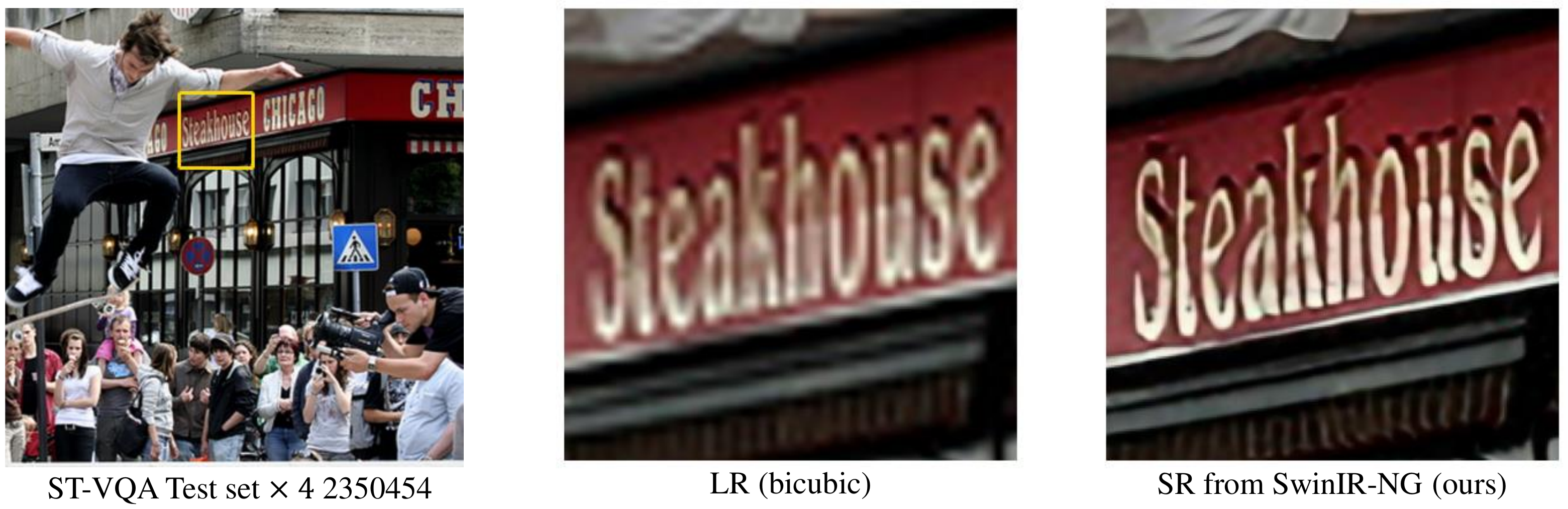} \\
        \includegraphics[width=0.9\linewidth]{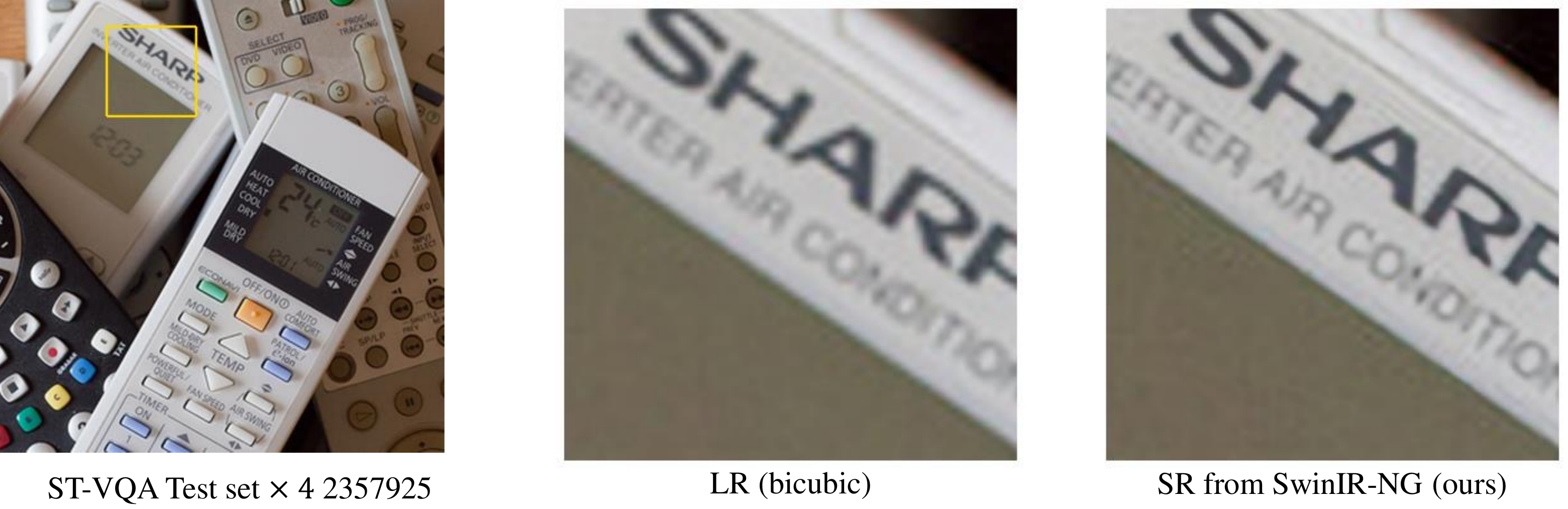} \\
        \includegraphics[width=0.9\linewidth]{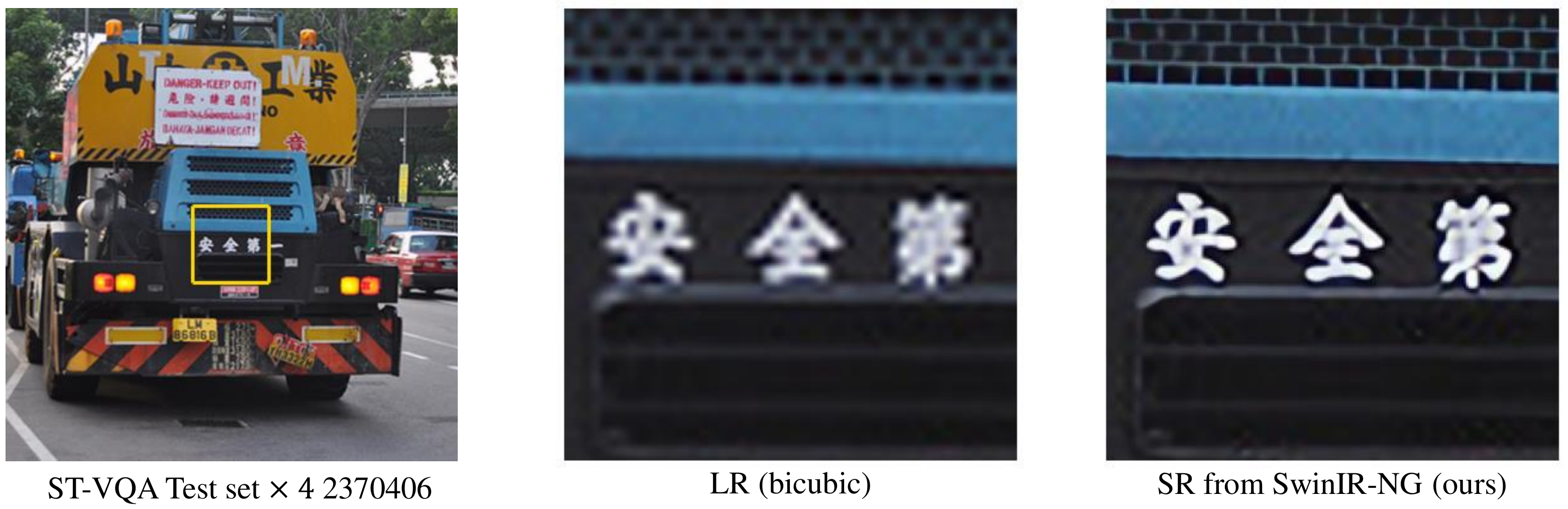} \\
        \includegraphics[width=0.9\linewidth]{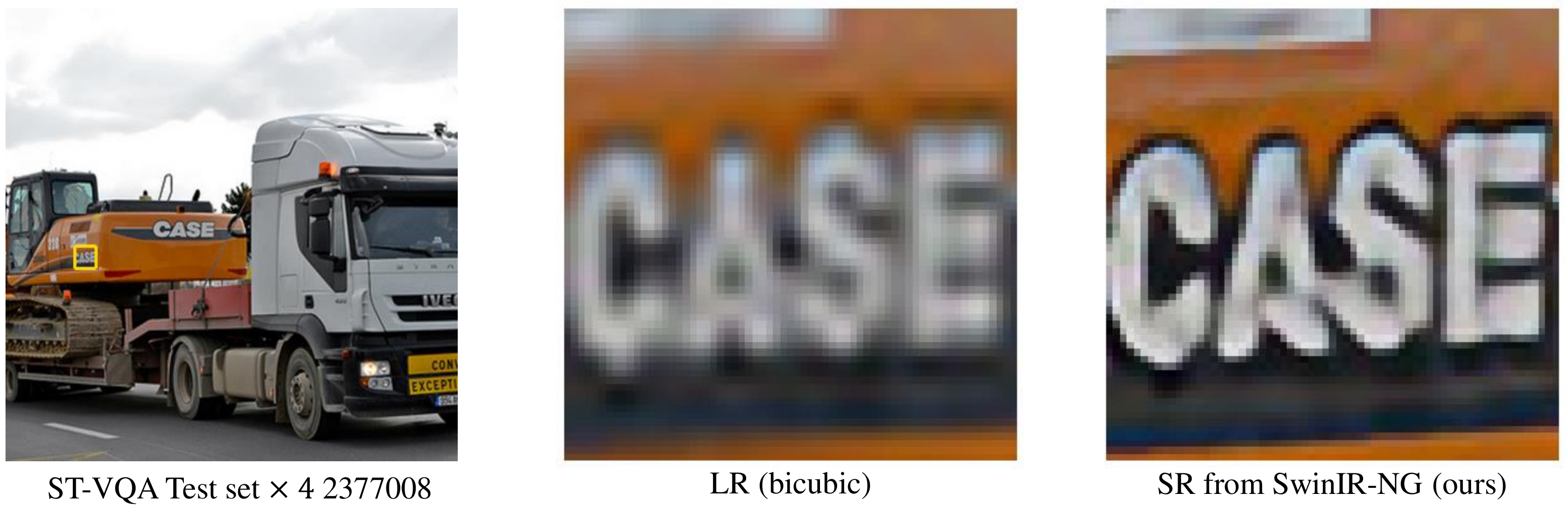} \\
    \end{tabular}

    \caption{SR results with SwinIR-NG on ST-VQA Test set~\cite{biten2019scene} ($\times 4$). Our SwinIR-NG makes the scene text more accurate to be detected than the original LR images. We expect our work to boost detection task as well. Like~\cref{fig_cifar1}, as this figure is secondary provision, we do not compare ours with other models.}
    \label{fig_stvqa}
\end{figure*}

\noindent
{\bf Other Tasks.}
In this paper, we worked on the super-resolution of the bicubic LR images.
Recently, some researchers studied the blind SR~\cite{liu2022blind}, where the input LR images are from unknown degradation.
In addition, other low-level vision tasks, such as deblurring, denoising, deraining, and JPEG artifact reduction, were developed by attention mechanisms~\cite{zhang2022accurate,Zamir2021Restormer,wang2022uformer,chen2021pre}.
Since these image restoration tasks also need the contextual information of distorted regions like the bicubic SR, our model introducing N-Gram to image would be helpful.
Secondarily, we visualized how our work can boost high-level vision tasks, such as classification (CIFAR10~\cite{krizhevsky2009learning}) and ST-VQA (Scene Text Visual Question Answering)~\cite{biten2019scene} in~\cref{fig_cifar1,,fig_cifar2,,fig_stvqa}.

\clearpage
\begin{table*}[t]
    \caption{The results of ablation studies on entire benchmarks. The tables in the parenthesis are the corresponding ones of~\cref{ablation}. The results in {\bf bold} are the best of each comparative content. PSNR / SSIM are reported.}
    \label{tab_other_bench}
    \centering
    \begin{subtable}[h]{0.85\linewidth}
    \caption{N-Gram context (\cref{tab_ngram_context}).}
    \resizebox{\linewidth}{!}
    {\begin{tabular}{c|c|c|c|c|c|c|c|c}
        \multicolumn{9}{c}{{\small NGswin without \vs with N-Gram}} \\
        \hline
        N-Gram & Scale & Mult-Adds & \#Params & Set5 & Set14 & BSD100 & Urban100 & Manga109 \\
        \hline
        \textit{w/o} & \multirow{2}{*}{$\times2$} & 138.20G & 750K & {\bf 38.05} / 0.9609 & 33.70 / 0.9194 & 32.25 / 0.9006 & 32.39 / 0.9304 & 38.86 / 0.9775 \\
        {\bf \textit{w/}} & & {\bf 140.41G} & {\bf 998K} & {\bf 38.05} / {\bf 0.9610} & {\bf 33.79} / {\bf 0.9199} & {\bf 32.27} / {\bf 0.9008} & {\bf 32.53} / {\bf 0.9324} & {\bf 38.97} / {\bf 0.9777} \\
        \hline
        \textit{w/o} & \multirow{2}{*}{$\times3$} & 65.53G & 759K & {\bf 34.53} / 0.9281 & 30.48 / 0.8451 & 29.15 / 0.8073 & 28.37 / 0.8573 & 33.81 / 0.9464 \\
        {\bf \textit{w/}} & & {\bf 66.56G} & {\bf 1,007K} & 34.52 / {\bf 0.9282} & {\bf 30.53} / {\bf 0.8456} & {\bf 29.19} / {\bf 0.8078} & {\bf 28.52} / {\bf 0.8603} & {\bf 33.89} / {\bf 0.9470} \\
        \hline
        \textit{w/o} & \multirow{4}{*}{$\times4$} & 35.89G & 771K & 32.34 / 0.8963 & 28.70 / 0.7844 & 27.63 / 0.7390 & 26.25 / 0.7918 & 30.70 / 0.9123 \\
        \textit{w/o} (channel up) & & 53.71G & 1,189K & 32.37 / {\bf 0.8973} & 28.75 / 0.7854 & 27.65 / 0.7396 & 26.28 / 0.7927 & 30.73 / 0.9129 \\
        \textit{w/o} (depth up) & & 47.88G & 1,061K & {\bf 32.40} / 0.8967 & 28.75 / 0.7853 & {\bf 27.66} / {\bf 0.7398} & 26.37 / 0.7946 & 30.78 / {\bf 0.9133} \\
        {\bf \textit{w/}} & & {\bf 36.44G} & {\bf 1,019K} & 32.33 / 0.8963 & {\bf 28.78} / {\bf 0.7859} & {\bf 27.66} / 0.7396 & {\bf 26.45} / {\bf 0.7963} & {\bf 30.80} / 0.9128 \\
        \hline
        \multicolumn{9}{c}{{}} \\
        \multicolumn{9}{c}{{\small HNCT \vs HNCT-NG}} \\
        \hline
        N-Gram & Scale & Mult-Adds & \#Params & Set5 & Set14 & BSD100 & Urban100 & Manga109 \\
        \hline
        \textit{w/o} & \multirow{2}{*}{$\times2$} & 82.39G & 357K & 38.08 / 0.9608 & {\bf 33.65} / 0.9182 & 32.22 / 0.9001 & 32.22 / 0.9294 & 38.87 / {\bf 0.9774} \\
        {\bf \textit{w/}} & & {\bf 83.19G} & {\bf 424K} & {\bf 38.10} / {\bf 0.9610} & 33.64 / {\bf 0.9195} & {\bf 32.25} / {\bf 0.9006} & {\bf 32.35} / {\bf 0.9306} & {\bf 38.94} / {\bf 0.9774} \\
        \hline
        \textit{w/o} & \multirow{2}{*}{$\times3$} & 37.78G & 363K & 34.47 / 0.9275 & 30.44 / 0.8439 & 29.15 / 0.8067 & 28.28 / 0.8557 & {\bf 33.81} / 0.9459 \\
        {\bf \textit{w/}} & & {\bf 38.14G} & {\bf 431K} & {\bf 34.48} / {\bf 0.9280} & {\bf 30.48} / {\bf 0.8450} & {\bf 29.16} / {\bf 0.8074} & {\bf 28.38} / {\bf 0.8573} & {\bf 33.81} / {\bf 0.9464} \\
        \hline
        \textit{w/o} & \multirow{2}{*}{$\times4$} & 22.01G & 373K & 32.31 / 0.8957 & 28.71 / 0.7834 & 27.63 / 0.7381 & 26.20 / 0.7896 & 30.70 / 0.9112 \\
        {\bf \textit{w/}} & & {\bf 22.21G} & {\bf 440K} & {\bf 32.32} / {\bf 0.8960} & {\bf 28.72} / {\bf 0.7846} & {\bf 27.65} / {\bf 0.7391} & {\bf 26.23} / {\bf 0.7912} & {\bf 30.71} / {\bf 0.9114} \\
        \hline
    \end{tabular}}
    \end{subtable}
    \begin{subtable}[h]{0.85\linewidth}
    \vspace{20pt}
    \caption{N-Gram directions and interaction (\cref{tab_ablation_direction_interaction}). The second best results are in \underline{underline}.}
    \resizebox{\linewidth}{!}
    {\begin{tabular}{c||c||c|c|c|c|c|c|c}
    \hline
        Direction & Type & Mult-Adds & \#Params & Set5 & Set14 & BSD100 & Urban100 &  Manga109 \\
        \hline
        1 & WSA & 152.41G & 1,238,056 & \underline{38.05} / {\bf 0.9610} & 33.78 / 0.9198 & \underline{32.26} / 0.9006 & {\bf 32.54} / 0.9322 & 38.90 / {\bf 0.9777}  \\
        4 & WSA & 139.56G & 935,272 & {\bf 38.07} / 0.9609 & 33.76 / 0.9197 & 32.25 / \underline{0.9007} & 32.52 / 0.9317 & 38.92 / 0.9776 \\
        1 & CNN & 139.80G & 1,327,528 & 38.04 / {\bf 0.9610} & 33.77 / 0.9197 & 32.25 / 0.9005 & 32.45 / 0.9316 & 38.86 / 0.9775 \\
        2 & CNN & 139.38G & 998,568 & 38.04 / {\bf 0.9610} & {\bf 33.83} / {\bf 0.9203} & \underline{32.26} / \underline{0.9007} & {\bf 32.54} / 0.9321 & 38.90 / 0.9776 \\
        4 & CNN & 139.17G & 936,488 & 38.02 / 0.9609 & 33.77 / 0.9178 & \underline{32.26} / 0.9006 & 32.52 / 0.9320 & \underline{38.93} / {\bf 0.9777} \\
        \hline
        {\bf 2} & {\bf WSA} & {\bf 140.41G} & {\bf 998,384} & \underline{38.05} / {\bf 0.9610} & \underline{33.79} / \underline{0.9199} & {\bf 32.27} / {\bf 0.9008} & 32.53 / {\bf 0.9324} & {\bf 38.97} / {\bf 0.9777} \\
    \hline
    \end{tabular}}
    \end{subtable}
    \begin{subtable}[h]{0.85\linewidth}
    \vspace{20pt}
    \caption{Extra stages and SCDP bottleneck (\cref{tab_ablation_scdp}).}
    \centering
    \resizebox{\linewidth}{!}
    {
    \begin{tabular}{c||c||c|c|c|c|c|c|c|c}
    \hline
        Stages & SCDP & Scale & Mult-Adds & \#Params & Set5 & Set14 & BSD100 & Urban100 & Manga109 \\
        \hline
        extra & \textit{w/o} & \multirow{3}{*}{$\times2$} & 87.98G & 997K & 38.02 / 0.9607 & 33.71 / 0.9193 & 32.20 / 0.8999 & 32.28 / 0.9298 & 38.72 / 0.9773 \\
        default & \textit{w/o} & & 138.88G & 992K & {\bf 38.08} / 0.9609 & {\bf 33.81} / {\bf 0.9199} & 32.24 / 0.9005 & 32.48 / 0.9321 & 38.92 / 0.9776 \\
        {\bf default} & {\bf \textit{w/}} & & {\bf 140.41G} & {\bf 998K} & 38.05 / {\bf 0.9610} & 33.79 / {\bf 0.9199} & {\bf 32.27} / {\bf 0.9008} & {\bf 32.53} / {\bf 0.9324} & {\bf 38.97} / {\bf 0.9777} \\
        \hline
        extra & \textit{w/o} & \multirow{3}{*}{$\times3$} & 42.10G & 1,006K & 34.38 / 0.9272 & 30.43 / 0.8437 & 29.11 / 0.8060 & 28.33 / 0.8562 & 33.67 / 0.9453 \\
        default & \textit{w/o} & & 65.85G & 1,001K & 34.47 / 0.9277 & 30.49 / 0.8454 & 29.17 / 0.8073 & 28.47 / 0.8596 & 33.81 / 0.9464 \\
        {\bf default} & {\bf \textit{w/}} & & {\bf 66.56G} & {\bf 1,007K} & {\bf 34.52} / {\bf 0.9282} & {\bf 30.53} / {\bf 0.8456} & {\bf 29.19} / {\bf 0.8078} & {\bf 28.52} / {\bf 0.8603} & {\bf 33.89} / {\bf 0.9470} \\
        \hline
        extra & \textit{w/o} & \multirow{3}{*}{$\times4$} & 23.33G & 1,018K & 32.17 / 0.8943 & 28.65 / 0.7827 & 27.59 / 0.7369 & 26.22 / 0.7900 & 30.46 / 0.9090 \\
        default & \textit{w/o} & & 36.06G & 1,013K & 32.29 / 0.8957 & 28.73 / 0.7849 & 27.64 / 0.7391 & 26.38 / 0.7954 & 30.71 / 0.9121 \\
        {\bf default} & {\bf \textit{w/}} & & {\bf 36.44G} & {\bf 1,019K} & {\bf 32.33} / {\bf 0.8963} & {\bf 28.78} / {\bf 0.7859} & {\bf 27.66} / {\bf 0.7396} & {\bf 26.45} / {\bf 0.7963} & {\bf 30.80} / {\bf 0.9128} \\
    \hline
    \end{tabular}
    }
    \end{subtable}
\end{table*}

\clearpage

\clearpage

{\small
\bibliographystyle{ieee_fullname}
\bibliography{bib}

\begin{thebibliography}{10}\itemsep=-1pt

\bibitem{agustsson2017ntire}
Eirikur Agustsson and Radu Timofte.
\newblock Ntire 2017 challenge on single image super-resolution: Dataset and
  study.
\newblock In {\em Proceedings of the IEEE conference on computer vision and
  pattern recognition workshops}, pages 126--135, 2017.

\bibitem{ahn2018fast}
Namhyuk Ahn, Byungkon Kang, and Kyung-Ah Sohn.
\newblock Fast, accurate, and lightweight super-resolution with cascading
  residual network.
\newblock In {\em Proceedings of the European conference on computer vision
  (ECCV)}, pages 252--268, 2018.

\bibitem{ahn2022efficient}
Namhyuk Ahn, Byungkon Kang, and Kyung-Ah Sohn.
\newblock Efficient deep neural network for photo-realistic image
  super-resolution.
\newblock {\em Pattern Recognition}, 127:108649, 2022.

\bibitem{arbelaez2010contour}
Pablo Arbelaez, Michael Maire, Charless Fowlkes, and Jitendra Malik.
\newblock Contour detection and hierarchical image segmentation.
\newblock {\em IEEE transactions on pattern analysis and machine intelligence},
  33(5):898--916, 2010.

\bibitem{ba2016layer}
Jimmy~Lei Ba, Jamie~Ryan Kiros, and Geoffrey~E Hinton.
\newblock Layer normalization.
\newblock {\em arXiv preprint arXiv:1607.06450}, 2016.

\bibitem{bao2021beit}
Hangbo Bao, Li Dong, and Furu Wei.
\newblock Beit: Bert pre-training of image transformers.
\newblock {\em arXiv preprint arXiv:2106.08254}, 2021.

\bibitem{behjati2023single}
Parichehr Behjati, Pau Rodriguez, Carles Fern{\'a}ndez, Isabelle Hupont, Armin
  Mehri, and Jordi Gonz{\`a}lez.
\newblock Single image super-resolution based on directional variance attention
  network.
\newblock {\em Pattern Recognition}, 133:108997, 2023.

\bibitem{bevilacqua2012low}
Marco Bevilacqua, Aline Roumy, Christine Guillemot, and Marie~Line
  Alberi-Morel.
\newblock Low-complexity single-image super-resolution based on nonnegative
  neighbor embedding.
\newblock 2012.

\bibitem{biten2019scene}
Ali~Furkan Biten, Ruben Tito, Andres Mafla, Lluis Gomez, Mar{\c{c}}al Rusinol,
  Ernest Valveny, CV Jawahar, and Dimosthenis Karatzas.
\newblock Scene text visual question answering.
\newblock In {\em Proceedings of the IEEE/CVF international conference on
  computer vision}, pages 4291--4301, 2019.

\bibitem{brown1992class}
Peter~F Brown, Vincent~J Della~Pietra, Peter~V Desouza, Jennifer~C Lai, and
  Robert~L Mercer.
\newblock Class-based n-gram models of natural language.
\newblock {\em Computational linguistics}, 18(4):467--480, 1992.

\bibitem{chen2021pre}
Hanting Chen, Yunhe Wang, Tianyu Guo, Chang Xu, Yiping Deng, Zhenhua Liu, Siwei
  Ma, Chunjing Xu, Chao Xu, and Wen Gao.
\newblock Pre-trained image processing transformer.
\newblock In {\em Proceedings of the IEEE/CVF Conference on Computer Vision and
  Pattern Recognition}, pages 12299--12310, 2021.

\bibitem{chen2022simple}
Liangyu Chen, Xiaojie Chu, Xiangyu Zhang, and Jian Sun.
\newblock Simple baselines for image restoration.
\newblock In {\em Computer Vision--ECCV 2022: 17th European Conference, Tel
  Aviv, Israel, October 23--27, 2022, Proceedings, Part VII}, pages 17--33.
  Springer, 2022.

\bibitem{clark2020electra}
Kevin Clark, Minh-Thang Luong, Quoc~V Le, and Christopher~D Manning.
\newblock Electra: Pre-training text encoders as discriminators rather than
  generators.
\newblock {\em arXiv preprint arXiv:2003.10555}, 2020.

\bibitem{cohen2016group}
Taco Cohen and Max Welling.
\newblock Group equivariant convolutional networks.
\newblock In {\em International conference on machine learning}, pages
  2990--2999. PMLR, 2016.

\bibitem{conde2022swin2sr}
Marcos~V Conde, Ui-Jin Choi, Maxime Burchi, and Radu Timofte.
\newblock Swin2sr: Swinv2 transformer for compressed image super-resolution and
  restoration.
\newblock {\em arXiv preprint arXiv:2209.11345}, 2022.

\bibitem{devlin2018bert}
Jacob Devlin, Ming-Wei Chang, Kenton Lee, and Kristina Toutanova.
\newblock Bert: Pre-training of deep bidirectional transformers for language
  understanding.
\newblock {\em arXiv preprint arXiv:1810.04805}, 2018.

\bibitem{diao2019zen}
Shizhe Diao, Jiaxin Bai, Yan Song, Tong Zhang, and Yonggang Wang.
\newblock Zen: Pre-training chinese text encoder enhanced by n-gram
  representations.
\newblock {\em arXiv preprint arXiv:1911.00720}, 2019.

\bibitem{ding2022davit}
Mingyu Ding, Bin Xiao, Noel Codella, Ping Luo, Jingdong Wang, and Lu Yuan.
\newblock Davit: Dual attention vision transformers.
\newblock In {\em Computer Vision--ECCV 2022: 17th European Conference, Tel
  Aviv, Israel, October 23--27, 2022, Proceedings, Part XXIV}, pages 74--92.
  Springer, 2022.

\bibitem{dong2022cswin}
Xiaoyi Dong, Jianmin Bao, Dongdong Chen, Weiming Zhang, Nenghai Yu, Lu Yuan,
  Dong Chen, and Baining Guo.
\newblock Cswin transformer: A general vision transformer backbone with
  cross-shaped windows.
\newblock In {\em Proceedings of the IEEE/CVF Conference on Computer Vision and
  Pattern Recognition}, pages 12124--12134, 2022.

\bibitem{dosovitskiy2020image}
Alexey Dosovitskiy, Lucas Beyer, Alexander Kolesnikov, Dirk Weissenborn,
  Xiaohua Zhai, Thomas Unterthiner, Mostafa Dehghani, Matthias Minderer, Georg
  Heigold, Sylvain Gelly, et~al.
\newblock An image is worth 16x16 words: Transformers for image recognition at
  scale.
\newblock {\em arXiv preprint arXiv:2010.11929}, 2020.

\bibitem{du2022fast}
Zongcai Du, Ding Liu, Jie Liu, Jie Tang, Gangshan Wu, and Lean Fu.
\newblock Fast and memory-efficient network towards efficient image
  super-resolution.
\newblock In {\em Proceedings of the IEEE/CVF Conference on Computer Vision and
  Pattern Recognition}, pages 853--862, 2022.

\bibitem{fang2022hybrid}
Jinsheng Fang, Hanjiang Lin, Xinyu Chen, and Kun Zeng.
\newblock A hybrid network of cnn and transformer for lightweight image
  super-resolution.
\newblock In {\em Proceedings of the IEEE/CVF Conference on Computer Vision and
  Pattern Recognition}, pages 1103--1112, 2022.

\bibitem{goyal2017accurate}
Priya Goyal, Piotr Doll{\'a}r, Ross Girshick, Pieter Noordhuis, Lukasz
  Wesolowski, Aapo Kyrola, Andrew Tulloch, Yangqing Jia, and Kaiming He.
\newblock Accurate, large minibatch sgd: Training imagenet in 1 hour.
\newblock {\em arXiv preprint arXiv:1706.02677}, 2017.

\bibitem{han2021flash}
Runze Han, Yachen Xiang, Peng Huang, Yihao Shan, Xiaoyan Liu, and Jinfeng Kang.
\newblock Flash memory array for efficient implementation of deep neural
  networks.
\newblock {\em Advanced Intelligent Systems}, 3(5):2000161, 2021.

\bibitem{he2022masked}
Kaiming He, Xinlei Chen, Saining Xie, Yanghao Li, Piotr Doll{\'a}r, and Ross
  Girshick.
\newblock Masked autoencoders are scalable vision learners.
\newblock In {\em Proceedings of the IEEE/CVF Conference on Computer Vision and
  Pattern Recognition}, pages 16000--16009, 2022.

\bibitem{he2016deep}
Kaiming He, Xiangyu Zhang, Shaoqing Ren, and Jian Sun.
\newblock Deep residual learning for image recognition.
\newblock In {\em Proceedings of the IEEE conference on computer vision and
  pattern recognition}, pages 770--778, 2016.

\bibitem{he2019view}
Xinwei He, Tengteng Huang, Song Bai, and Xiang Bai.
\newblock View n-gram network for 3d object retrieval.
\newblock In {\em Proceedings of the IEEE/CVF International Conference on
  Computer Vision}, pages 7515--7524, 2019.

\bibitem{hinton2012improving}
Geoffrey~E Hinton, Nitish Srivastava, Alex Krizhevsky, Ilya Sutskever, and
  Ruslan~R Salakhutdinov.
\newblock Improving neural networks by preventing co-adaptation of feature
  detectors.
\newblock {\em arXiv preprint arXiv:1207.0580}, 2012.

\bibitem{huang2017densely}
Gao Huang, Zhuang Liu, Laurens Van Der~Maaten, and Kilian~Q Weinberger.
\newblock Densely connected convolutional networks.
\newblock In {\em Proceedings of the IEEE conference on computer vision and
  pattern recognition}, pages 4700--4708, 2017.

\bibitem{huang2015single}
Jia-Bin Huang, Abhishek Singh, and Narendra Ahuja.
\newblock Single image super-resolution from transformed self-exemplars.
\newblock In {\em Proceedings of the IEEE conference on computer vision and
  pattern recognition}, pages 5197--5206, 2015.

\bibitem{hui2019lightweight}
Zheng Hui, Xinbo Gao, Yunchu Yang, and Xiumei Wang.
\newblock Lightweight image super-resolution with information
  multi-distillation network.
\newblock In {\em Proceedings of the 27th acm international conference on
  multimedia}, pages 2024--2032, 2019.

\bibitem{kim2016accurate}
Jiwon Kim, Jung~Kwon Lee, and Kyoung~Mu Lee.
\newblock Accurate image super-resolution using very deep convolutional
  networks.
\newblock In {\em Proceedings of the IEEE conference on computer vision and
  pattern recognition}, pages 1646--1654, 2016.

\bibitem{kingma2014adam}
Diederik~P Kingma and Jimmy Ba.
\newblock Adam: A method for stochastic optimization.
\newblock {\em arXiv preprint arXiv:1412.6980}, 2014.

\bibitem{kong2022reflash}
Xiangtao Kong, Xina Liu, Jinjin Gu, Yu Qiao, and Chao Dong.
\newblock Reflash dropout in image super-resolution.
\newblock In {\em Proceedings of the IEEE/CVF Conference on Computer Vision and
  Pattern Recognition}, pages 6002--6012, 2022.

\bibitem{krizhevsky2009learning}
Alex Krizhevsky, Geoffrey Hinton, et~al.
\newblock Learning multiple layers of features from tiny images.
\newblock 2009.

\bibitem{kulkami2016texture}
Pradnya Kulkami, Andrew Stranieri, and Julien Ugon.
\newblock Texture image classification using pixel n-grams.
\newblock In {\em Proceedings of 2016 IEEE International Conference on Signal
  and Image Processing (ICSIP)}, pages 137--141. IEEE, 2016.

\bibitem{larsson2016fractalnet}
Gustav Larsson, Michael Maire, and Gregory Shakhnarovich.
\newblock Fractalnet: Ultra-deep neural networks without residuals.
\newblock {\em arXiv preprint arXiv:1605.07648}, 2016.

\bibitem{liang2021swinir}
Jingyun Liang, Jiezhang Cao, Guolei Sun, Kai Zhang, Luc Van~Gool, and Radu
  Timofte.
\newblock Swinir: Image restoration using swin transformer.
\newblock In {\em Proceedings of the IEEE/CVF International Conference on
  Computer Vision}, pages 1833--1844, 2021.

\bibitem{lim2017enhanced}
Bee Lim, Sanghyun Son, Heewon Kim, Seungjun Nah, and Kyoung Mu~Lee.
\newblock Enhanced deep residual networks for single image super-resolution.
\newblock In {\em Proceedings of the IEEE conference on computer vision and
  pattern recognition workshops}, pages 136--144, 2017.

\bibitem{lin2022revisiting}
Zudi Lin, Prateek Garg, Atmadeep Banerjee, Salma~Abdel Magid, Deqing Sun, Yulun
  Zhang, Luc Van~Gool, Donglai Wei, and Hanspeter Pfister.
\newblock Revisiting rcan: Improved training for image super-resolution.
\newblock {\em arXiv preprint arXiv:2201.11279}, 2022.

\bibitem{liu2022blind}
Anran Liu, Yihao Liu, Jinjin Gu, Yu Qiao, and Chao Dong.
\newblock Blind image super-resolution: A survey and beyond.
\newblock {\em IEEE Transactions on Pattern Analysis and Machine Intelligence},
  2022.

\bibitem{liu2020residual}
Jie Liu, Jie Tang, and Gangshan Wu.
\newblock Residual feature distillation network for lightweight image
  super-resolution.
\newblock In {\em European Conference on Computer Vision}, pages 41--55.
  Springer, 2020.

\bibitem{liu2022swin}
Ze Liu, Han Hu, Yutong Lin, Zhuliang Yao, Zhenda Xie, Yixuan Wei, Jia Ning, Yue
  Cao, Zheng Zhang, Li Dong, et~al.
\newblock Swin transformer v2: Scaling up capacity and resolution.
\newblock In {\em Proceedings of the IEEE/CVF Conference on Computer Vision and
  Pattern Recognition}, pages 12009--12019, 2022.

\bibitem{liu2021swin}
Ze Liu, Yutong Lin, Yue Cao, Han Hu, Yixuan Wei, Zheng Zhang, Stephen Lin, and
  Baining Guo.
\newblock Swin transformer: Hierarchical vision transformer using shifted
  windows.
\newblock In {\em Proceedings of the IEEE/CVF International Conference on
  Computer Vision}, pages 10012--10022, 2021.

\bibitem{lopez2019word}
Inigo Lopez-Gazpio, Montse Maritxalar, Mirella Lapata, and Eneko Agirre.
\newblock Word n-gram attention models for sentence similarity and inference.
\newblock {\em Expert Systems with Applications}, 132:1--11, 2019.

\bibitem{loshchilov2016sgdr}
Ilya Loshchilov and Frank Hutter.
\newblock Sgdr: Stochastic gradient descent with warm restarts.
\newblock {\em arXiv preprint arXiv:1608.03983}, 2016.

\bibitem{loshchilov2017decoupled}
Ilya Loshchilov and Frank Hutter.
\newblock Decoupled weight decay regularization.
\newblock {\em arXiv preprint arXiv:1711.05101}, 2017.

\bibitem{lu2021efficient}
Zhisheng Lu, Hong Liu, Juncheng Li, and Linlin Zhang.
\newblock Efficient transformer for single image super-resolution.
\newblock {\em arXiv preprint arXiv:2108.11084}, 2021.

\bibitem{luo2022lattice}
Xiaotong Luo, Yanyun Qu, Yuan Xie, Yulun Zhang, Cuihua Li, and Yun Fu.
\newblock Lattice network for lightweight image restoration.
\newblock {\em IEEE Transactions on Pattern Analysis and Machine Intelligence},
  2022.

\bibitem{luo2020latticenet}
Xiaotong Luo, Yuan Xie, Yulun Zhang, Yanyun Qu, Cuihua Li, and Yun Fu.
\newblock Latticenet: Towards lightweight image super-resolution with lattice
  block.
\newblock In {\em European Conference on Computer Vision}, pages 272--289.
  Springer, 2020.

\bibitem{magid2022texture}
Salma~Abdel Magid, Zudi Lin, Donglai Wei, Yulun Zhang, Jinjin Gu, and Hanspeter
  Pfister.
\newblock Texture-based error analysis for image super-resolution.
\newblock In {\em Proceedings of the IEEE/CVF Conference on Computer Vision and
  Pattern Recognition}, pages 2118--2127, 2022.

\bibitem{majumder2002n}
P Majumder, M Mitra, and BB Chaudhuri.
\newblock N-gram: a language independent approach to ir and nlp.
\newblock In {\em International conference on universal knowledge and
  language}, 2002.

\bibitem{martin2001database}
David Martin, Charless Fowlkes, Doron Tal, and Jitendra Malik.
\newblock A database of human segmented natural images and its application to
  evaluating segmentation algorithms and measuring ecological statistics.
\newblock In {\em Proceedings Eighth IEEE International Conference on Computer
  Vision. ICCV 2001}, volume~2, pages 416--423. IEEE, 2001.

\bibitem{matsui2017sketch}
Yusuke Matsui, Kota Ito, Yuji Aramaki, Azuma Fujimoto, Toru Ogawa, Toshihiko
  Yamasaki, and Kiyoharu Aizawa.
\newblock Sketch-based manga retrieval using manga109 dataset.
\newblock {\em Multimedia Tools and Applications}, 76(20):21811--21838, 2017.

\bibitem{niu2020single}
Ben Niu, Weilei Wen, Wenqi Ren, Xiangde Zhang, Lianping Yang, Shuzhen Wang,
  Kaihao Zhang, Xiaochun Cao, and Haifeng Shen.
\newblock Single image super-resolution via a holistic attention network.
\newblock In {\em European conference on computer vision}, pages 191--207.
  Springer, 2020.

\bibitem{pagliardini2017unsupervised}
Matteo Pagliardini, Prakhar Gupta, and Martin Jaggi.
\newblock Unsupervised learning of sentence embeddings using compositional
  n-gram features.
\newblock {\em arXiv preprint arXiv:1703.02507}, 2017.

\bibitem{pang2022masked}
Yatian Pang, Wenxiao Wang, Francis~EH Tay, Wei Liu, Yonghong Tian, and Li Yuan.
\newblock Masked autoencoders for point cloud self-supervised learning.
\newblock {\em arXiv preprint arXiv:2203.06604}, 2022.

\bibitem{pascanu2012understanding}
Razvan Pascanu, Tomas Mikolov, and Yoshua Bengio.
\newblock Understanding the exploding gradient problem.
\newblock {\em CoRR, abs/1211.5063}, 2(417):1, 2012.

\bibitem{pascanu2013difficulty}
Razvan Pascanu, Tomas Mikolov, and Yoshua Bengio.
\newblock On the difficulty of training recurrent neural networks.
\newblock In {\em International conference on machine learning}, pages
  1310--1318. PMLR, 2013.

\bibitem{paszke2019pytorch}
Adam Paszke, Sam Gross, Francisco Massa, Adam Lerer, James Bradbury, Gregory
  Chanan, Trevor Killeen, Zeming Lin, Natalia Gimelshein, Luca Antiga, et~al.
\newblock Pytorch: An imperative style, high-performance deep learning library.
\newblock {\em Advances in neural information processing systems}, 32, 2019.

\bibitem{reed1993pruning}
Russell Reed.
\newblock Pruning algorithms-a survey.
\newblock {\em IEEE transactions on Neural Networks}, 4(5):740--747, 1993.

\bibitem{ronneberger2015u}
Olaf Ronneberger, Philipp Fischer, and Thomas Brox.
\newblock U-net: Convolutional networks for biomedical image segmentation.
\newblock In {\em International Conference on Medical image computing and
  computer-assisted intervention}, pages 234--241. Springer, 2015.

\bibitem{ruiz2021dopant}
Hans-Christian Ruiz-Euler, Unai Alegre-Ibarra, Bram van~de Ven, Hajo Broersma,
  Peter~A Bobbert, and Wilfred~G van~der Wiel.
\newblock Dopant network processing units: towards efficient neural network
  emulators with high-capacity nanoelectronic nodes.
\newblock {\em Neuromorphic Computing and Engineering}, 1(2):024002, 2021.

\bibitem{shi2016real}
Wenzhe Shi, Jose Caballero, Ferenc Husz{\'a}r, Johannes Totz, Andrew~P Aitken,
  Rob Bishop, Daniel Rueckert, and Zehan Wang.
\newblock Real-time single image and video super-resolution using an efficient
  sub-pixel convolutional neural network.
\newblock In {\em Proceedings of the IEEE conference on computer vision and
  pattern recognition}, pages 1874--1883, 2016.

\bibitem{song2021zen}
Yan Song, Tong Zhang, Yonggang Wang, and Kai-Fu Lee.
\newblock Zen 2.0: Continue training and adaption for n-gram enhanced text
  encoders.
\newblock {\em arXiv preprint arXiv:2105.01279}, 2021.

\bibitem{tai2017memnet}
Ying Tai, Jian Yang, Xiaoming Liu, and Chunyan Xu.
\newblock Memnet: A persistent memory network for image restoration.
\newblock In {\em Proceedings of the IEEE international conference on computer
  vision}, pages 4539--4547, 2017.

\bibitem{timofte2017ntire}
Radu Timofte, Eirikur Agustsson, Luc Van~Gool, Ming-Hsuan Yang, and Lei Zhang.
\newblock Ntire 2017 challenge on single image super-resolution: Methods and
  results.
\newblock In {\em Proceedings of the IEEE conference on computer vision and
  pattern recognition workshops}, pages 114--125, 2017.

\bibitem{tu2022maxvit}
Zhengzhong Tu, Hossein Talebi, Han Zhang, Feng Yang, Peyman Milanfar, Alan
  Bovik, and Yinxiao Li.
\newblock Maxvit: Multi-axis vision transformer.
\newblock In {\em Computer Vision--ECCV 2022: 17th European Conference, Tel
  Aviv, Israel, October 23--27, 2022, Proceedings, Part XXIV}, pages 459--479.
  Springer, 2022.

\bibitem{vaswani2017attention}
Ashish Vaswani, Noam Shazeer, Niki Parmar, Jakob Uszkoreit, Llion Jones,
  Aidan~N Gomez, {\L}ukasz Kaiser, and Illia Polosukhin.
\newblock Attention is all you need.
\newblock {\em Advances in neural information processing systems}, 30, 2017.

\bibitem{wang2004image}
Zhou Wang, Alan~C Bovik, Hamid~R Sheikh, and Eero~P Simoncelli.
\newblock Image quality assessment: from error visibility to structural
  similarity.
\newblock {\em IEEE transactions on image processing}, 13(4):600--612, 2004.

\bibitem{wang2022uformer}
Zhendong Wang, Xiaodong Cun, Jianmin Bao, Wengang Zhou, Jianzhuang Liu, and
  Houqiang Li.
\newblock Uformer: A general u-shaped transformer for image restoration.
\newblock In {\em Proceedings of the IEEE/CVF Conference on Computer Vision and
  Pattern Recognition}, pages 17683--17693, 2022.

\bibitem{yang2021focal}
Jianwei Yang, Chunyuan Li, Pengchuan Zhang, Xiyang Dai, Bin Xiao, Lu Yuan, and
  Jianfeng Gao.
\newblock Focal self-attention for local-global interactions in vision
  transformers.
\newblock {\em arXiv preprint arXiv:2107.00641}, 2021.

\bibitem{yang2010image}
Jianchao Yang, John Wright, Thomas~S Huang, and Yi Ma.
\newblock Image super-resolution via sparse representation.
\newblock {\em IEEE transactions on image processing}, 19(11):2861--2873, 2010.

\bibitem{yu2021glance}
Qihang Yu, Yingda Xia, Yutong Bai, Yongyi Lu, Alan~L Yuille, and Wei Shen.
\newblock Glance-and-gaze vision transformer.
\newblock {\em Advances in Neural Information Processing Systems},
  34:12992--13003, 2021.

\bibitem{Zamir2021Restormer}
Syed~Waqas Zamir, Aditya Arora, Salman Khan, Munawar Hayat, Fahad~Shahbaz Khan,
  and Ming-Hsuan Yang.
\newblock Restormer: Efficient transformer for high-resolution image
  restoration.
\newblock In {\em CVPR}, 2022.

\bibitem{zeyde2010single}
Roman Zeyde, Michael Elad, and Matan Protter.
\newblock On single image scale-up using sparse-representations.
\newblock In {\em International conference on curves and surfaces}, pages
  711--730. Springer, 2010.

\bibitem{zhang2022accurate}
Jiale Zhang, Yulun Zhang, Jinjin Gu, Yongbing Zhang, Linghe Kong, and Xin Yuan.
\newblock Accurate image restoration with attention retractable transformer.
\newblock {\em arXiv preprint arXiv:2210.01427}, 2022.

\bibitem{zhang2022vsa}
Qiming Zhang, Yufei Xu, Jing Zhang, and Dacheng Tao.
\newblock Vsa: learning varied-size window attention in vision transformers.
\newblock In {\em Computer Vision--ECCV 2022: 17th European Conference, Tel
  Aviv, Israel, October 23--27, 2022, Proceedings, Part XXV}, pages 466--483.
  Springer, 2022.

\bibitem{zhang2022efficient}
Xindong Zhang, Hui Zeng, Shi Guo, and Lei Zhang.
\newblock Efficient long-range attention network for image super-resolution.
\newblock {\em arXiv preprint arXiv:2203.06697}, 2022.

\bibitem{zhang2018image}
Yulun Zhang, Kunpeng Li, Kai Li, Lichen Wang, Bineng Zhong, and Yun Fu.
\newblock Image super-resolution using very deep residual channel attention
  networks.
\newblock In {\em Proceedings of the European conference on computer vision
  (ECCV)}, pages 286--301, 2018.

\bibitem{zhang2021learning}
Yulun Zhang, Huan Wang, Can Qin, and Yun Fu.
\newblock Learning efficient image super-resolution networks via
  structure-regularized pruning.
\newblock In {\em International Conference on Learning Representations}, 2021.

\bibitem{zheng2022cross}
Chen Zheng, Yulun Zhang, Jinjin Gu, Yongbing Zhang, Linghe Kong, and Xin Yuan.
\newblock Cross aggregation transformer for image restoration.
\newblock {\em arXiv preprint arXiv:2211.13654}, 2022.

\end{thebibliography}
}

\end{document}